\documentclass[letterpaper, 10 pt, journal, twoside]{ieeetran} 
\usepackage{graphicx}
\usepackage{amsmath}
\usepackage{bbding}
\usepackage{pifont}
\usepackage{wasysym}
\usepackage{amssymb}
\usepackage{hyperref}
\usepackage{bbm}
\usepackage{threeparttable} 
\usepackage{cite} 
\usepackage{booktabs}
\usepackage{xcolor}
\usepackage{multirow}
\usepackage{blindtext}
\usepackage{bm}
\usepackage{subcaption}
\usepackage{amsmath,amssymb,mathtools}
\usepackage[linesnumbered, ruled]{algorithm2e}
\DeclareMathOperator*{\argmax}{argmax}
\DeclareMathOperator*{\argmin}{argmin}
\SetKwInput{KwInput}{Input}
\SetKwInput{KwOutput}{Output}

\usepackage{cite}

\IEEEoverridecommandlockouts                              
\usepackage{geometry}
\usepackage{afterpage}
\usepackage{mwe}
\title{RNGDet: Road Network Graph Detection by Transformer in Aerial Images}
\author{Zhenhua Xu, \IEEEmembership{Student Member, IEEE}, 
Yuxuan Liu, \IEEEmembership{Student Member, IEEE},\\ Lu Gan, \IEEEmembership{Student Member, IEEE},  Yuxiang Sun, \IEEEmembership{Member, IEEE}, Xinyu Wu, \IEEEmembership{Member, IEEE}, \\ Ming Liu, \IEEEmembership{Senior Member, IEEE}, and Lujia Wang,  \IEEEmembership{Member, IEEE} %

\thanks{This work was supported by Zhongshan Science and Technology Bureau Fund, under project 2020AG002, Foshan-HKUST Project no. FSUST20-SHCIRI06C, and Guangdong Basic and Applied Basic Research Foundation project no. 2020A0505090008, awarded to Prof. Ming Liu.}

\thanks{Zhenhua Xu, Yuxuan Liu, Lu Gan are with The Hong Kong University of Science and Technology (email: \{zxubg,yliuhb,lganaa\}@connect.ust.hk).}
\thanks{Yuxiang Sun is with the Department of Mechanical Engineering, The Hong Kong Polytechnic University, Hung Hom, Kowloon, Hong Kong (e-mail: yx.sun@polyu.edu.hk, sun.yuxiang@outlook.com).}
\thanks{Xinyu Wu is with Shenzhen Institutes of Advanced Technology, CAS, Shenzhen, China (email: xy.wu@siat.ac.cn)}
\thanks{Ming Liu is with The Hong Kong University of Science and Technology (Guangzhou), Nansha, Guangzhou, 511400, Guangdong, China, and also with The Hong Kong University of Science and Technology, Hong Kong SAR, China, and also with HKUST Shenzhen-Hong Kong Collaborative Innovation Research Institute, Futian, Shenzhen. (email: eelium@ust.hk)}
\thanks{Lujia Wang is with The Hong Kong University of Science and Technology, and also with Clear Water Bay Insitute of Autonomous Driving (Shenzhen) (email: eewang@ust.hk).}
\thanks{\textit{Corresponding author: Lujia Wang.}}
} 
\IEEEaftertitletext{\vspace{-1.0\baselineskip}}
\begin{document}

\maketitle
\begin{abstract}      
Road network graphs provide critical information for autonomous-vehicle applications, such as drivable areas that can be used for motion planning algorithms. To find road network graphs, manually annotation is usually inefficient and labor-intensive. Automatically detecting road network graphs could alleviate this issue, but existing works still have some limitations. For example, segmentation-based approaches could not ensure satisfactory topology correctness, and graph-based approaches could not present precise enough detection results. To provide a solution to these problems, we propose a novel approach based on transformer and imitation learning in this paper. In view of that high-resolution aerial images could be easily accessed all over the world nowadays, we make use of aerial images in our approach. Taken as input an aerial image, our approach iteratively generates road network graphs vertex-by-vertex. Our approach can handle complicated intersection points with various numbers of incident road segments. We evaluate our approach on a publicly available dataset. The superiority of our approach is demonstrated through the comparative experiments. Our work is accompanied with a demonstration video which is available at \url{https://tonyxuqaq.github.io/projects/RNGDet/}.

\vspace{0.25cm}

\end{abstract}

\section{Introduction}
\IEEEPARstart{I}{n} recent years, road networks have attracted considerable attention in the field of autonomous driving. The graph of road networks can provide fundamental information for autonomous-vehicle applications. 
The graph of road networks is a kind of vectorized data representation, which consists of vertices and edges \cite{chiang2009extracting}. Each road segment could be seen as a graph edge, and the intersection points of road segments are vertices. Manually annotating the road network graph is time-consuming and labor-intensive, especially when road networks cover a large area (e.g., a whole city). Therefore, how to automatically detect road network graphs using automatic algorithms in large areas is of great interest to the research community.

 \begin{figure}[t]
    \begin{subfigure}{.24\textwidth}
        \includegraphics[width=\textwidth]{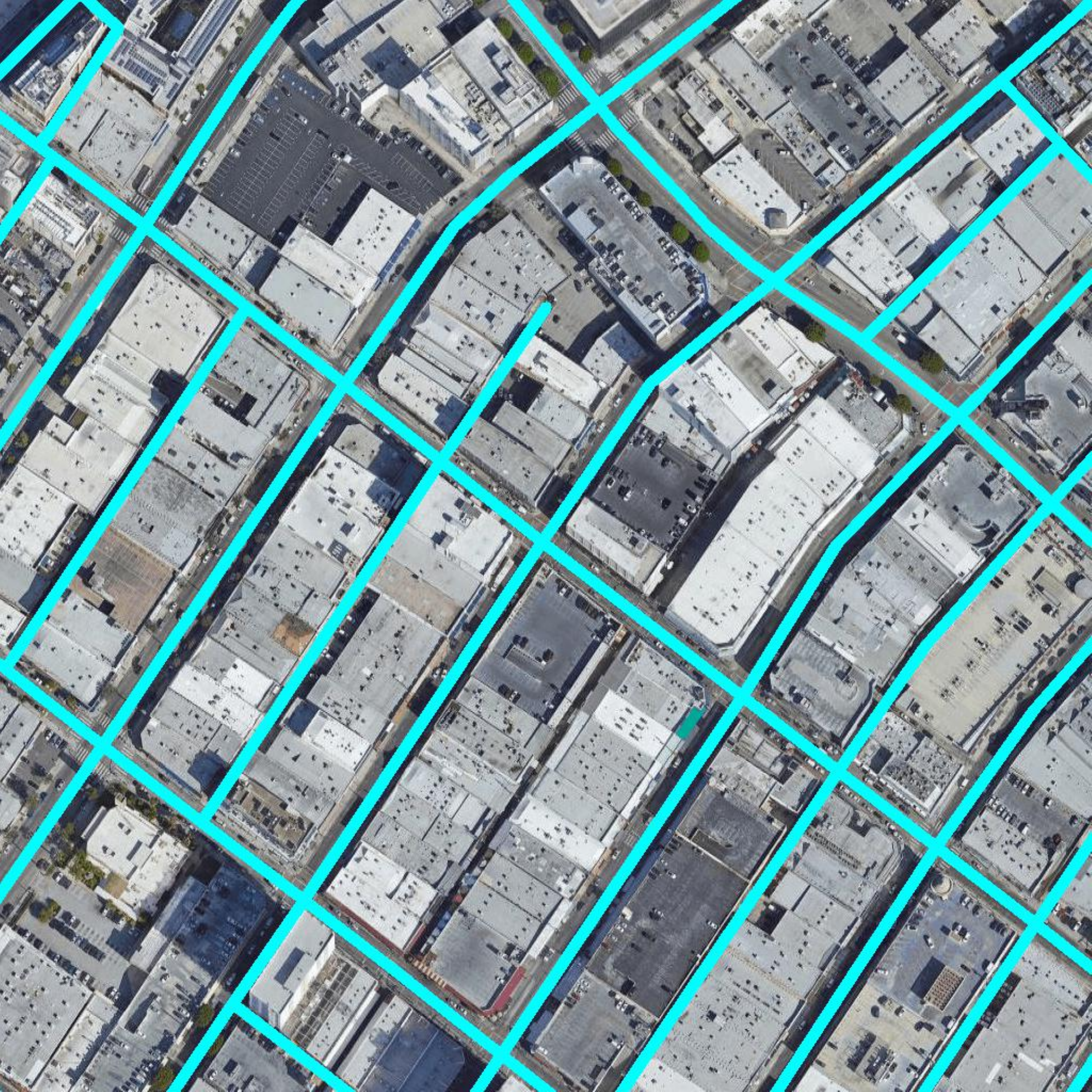}\caption{Ground-truth}
    \end{subfigure}\hfill
    \begin{subfigure}{.24\textwidth}
        \includegraphics[width=\textwidth]{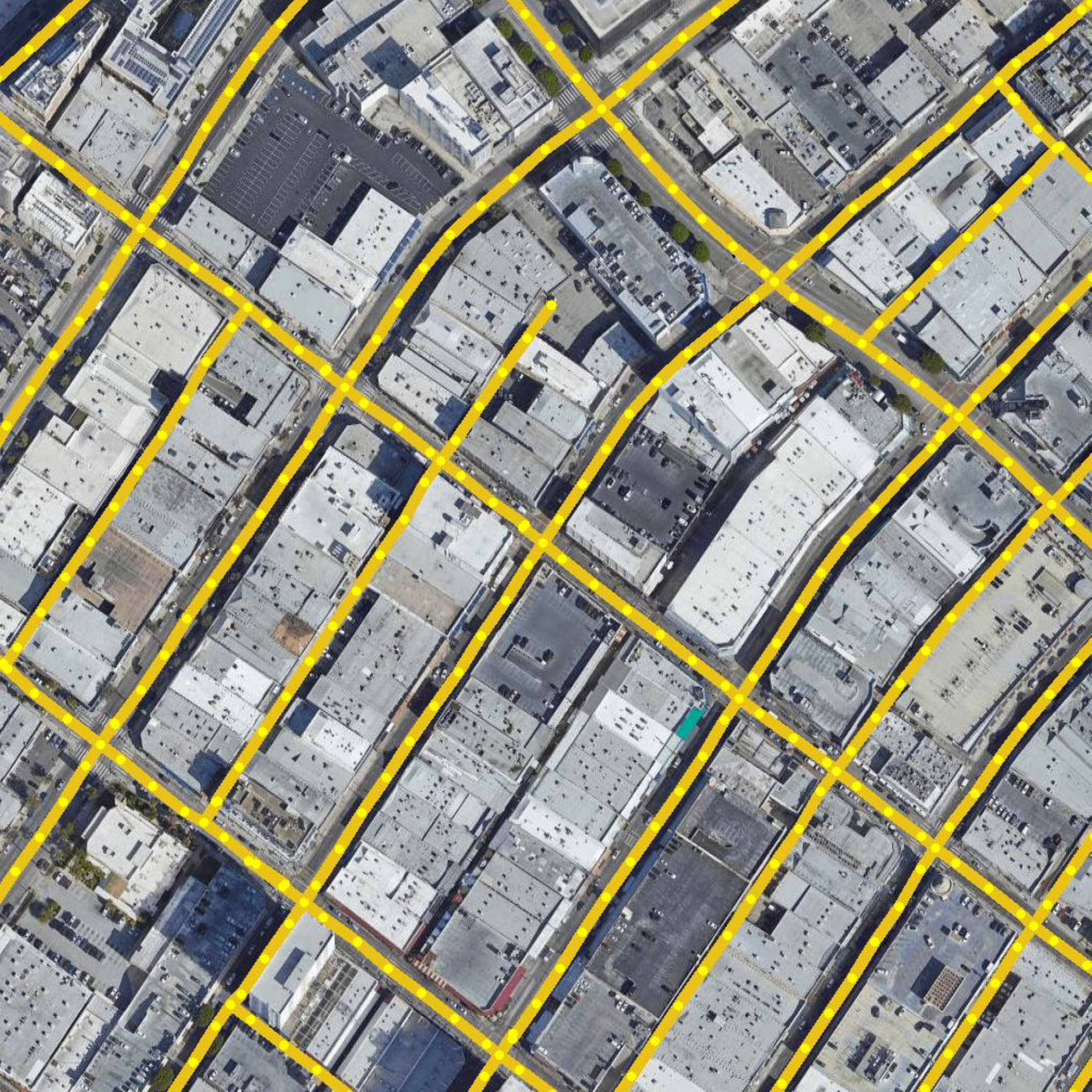}\caption{RNGDet}
    \end{subfigure}
    \caption{A sample result of RNGDet. (a) The ground-truth road network graph (cyan lines). (b) The graph predicted by RNGDet (orange lines for edges and yellow points for vertices). We can see that RNGDet effectively detects the road network graph with high quality. For better visualization, the lines are drawn with a thicker width while it is actually one-pixel width. The figure is best viewed in color. Please zoom in for details.}
    \label{sample_demo_result}
\end{figure}

To address this issue, past approaches on road network graph detection usually use aerial images obtained from unmanned aerial vehicles (UAVs) or satellites \cite{mnih2010learning}. As aerial imaging technology evolves, high-resolution and high-quality aerial images can be easily accessed world-widely nowadays. Moreover, some aerial imaging datasets also provide extra channels besides Red-Green-Blue (RGB), such as the infrared channel \cite{nyc_dataset}, making them more informative for detection purpose. So, this work also uses aerial images for road network graph detection. 

Existing works on road network graph detection can be generally classified into two categories: segmentation-based approaches \cite{hu2014road,shi2013spectral,unsalan2012road,cheng2016road,batra2019improved,mattyus2017deeproadmapper,mnih2010learning,etten2020city,gedara2021spin,cheng2017automatic,zhou2021split} and graph-based approaches \cite{bastani2018roadtracer,li2018polymapper,he2020sat2graph,tan2020vecroad,belli2019image}. The segmentation-based approaches first predict the probabilistic segmentation map of the road network graph, and then conduct a series of processing to obtain the graph structure of the road network, such as skeletonization and filtering. Most of the early works on road network graph detection in this field fall into this category. The segmentation-based approaches could present good results in the pixel-level evaluation (e.g., by F1 score) due to the use of existing powerful semantic segmentation networks, but they usually suffer from unsatisfactory topology correctness such as incorrect crossroad connectivity and false disconnection on the road. 
To address this issue, recent graph-based approaches resort to detecting the graph of road networks directly \cite{bastani2018roadtracer,li2018polymapper,tan2020vecroad,belli2019image}. They usually first predict candidate initial vertices, then, starting from each candidate initial vertex, train a decision-making agent to predict adjacent vertices of the current vertex. In this way, road network graphs can be generated vertex-by-vertex in an iterative manner. Although these graph-based approaches could enhance the topology correctness, they are usually composed of two separate stages, making them hard to be optimized in an end-to-end way. The separate stages might accumulate errors and hence degrade their effectiveness and efficiency.

To provide a solution to these issues, in this paper, we propose a graph-based end-to-end approach named \underline{R}oad \underline{N}etwork \underline{G}raph \underline{Det}ection by Transformer (RNGDet). Similar to previous graph-based approaches, RNGDet starts from predicted candidate initial vertices to extract local visual features using a convolutional neural network (CNN) backbone, and then sends the features to a transformer network inspired by the DETR structure \cite{carion2020end}. 
Due to the use of deep vertex queries, RNGDet can directly predict any number of adjacent vertices of the current vertex at one time, so that it can handle any road networks, even those with complicated topology (e.g., road intersections of arbitrary numbers of road segments). Different from previous graph-based approaches, RNGDet can be optimized as a whole and trained end-to-end. We train RNGDet through imitation learning to enable it to take the most appropriate action under different circumstances. To generate the training data (i.e., expert demonstration from the imitation learning perspective), we propose a sampling algorithm to supervise the agent to explore the whole road network. 
The proposed RNGDet is trained and evaluated on a publicly available dataset released by RoadTracer \cite{bastani2018roadtracer}. With this dataset, we compare RNGDet with state-of-the-art works based on multiple evaluation metric scores. An example of RNGDet is visualized in Fig. \ref{sample_demo_result}. The contributions of our work are summarized below:
\begin{itemize}
    \item We propose an end-to-end trainable approach named RNGDet based on transformer and imitation learning to automatically detect the road network graph.
    \item We propose an algorithm to automatically generate training samples for RNGDet.
    \item We evaluate RNGDet and compare it with state-of-the-art works on a publicly available dataset to demonstrate the superiority of RNGDet.
\end{itemize}

The remainder of this article is organized as follows. Section II introduces related works. Section III describes the structure and working pipeline of RNGDet. Section IV presents experimental results, discussions, and limitations. Conclusions and future work are drawn in the last section.

\section{Related works}
\subsection{Segmentation-based approaches}
Segmentation-based road network graph detection approaches \cite{hu2014road,shi2013spectral,unsalan2012road,cheng2016road,batra2019improved,mattyus2017deeproadmapper,mnih2010learning,etten2020city,gedara2021spin,cheng2017automatic,zhou2021split,ganganguli2019geogan,ganhe2020generative,gankang2019transferring} mainly have two stages: (1) predict the segmentation map (i.e., probabilistic map of the road network) and (2) process the segmentation map and obtain graph structures by post-processing, such as skeletonization and heuristic-based algorithms \cite{newwenzel2019simultaneous,newbulatov2017chain}. It is believed that \cite{mnih2010learning} proposed by Mnih \textit{et al.} is the first work that implemented neural networks to detect the road network in aerial images. They first split the large aerial image into small patches, then predicted the road network within each patch and finally merged patches into the final predicted road network segmentation map. Most later segmentation-based works followed a similar pipeline, but with more powerful segmentation networks, such as U-Net \cite{ranzato2015sequence}, DeepLab V3+ \cite{chen2018encoder}, FPN \cite{lin2017feature} and HRNet \cite{wang2020deep}. Batra \textit{et al.} \cite{batra2019improved} extended the aforementioned works by adding another refinement network to fix incorrect pixels in the predicted segmentation map, which effectively improved the final performance. In \cite{gedara2021spin}, Gedara \textit{et al.} proposed a Spatial and Interaction Space Graph Reasoning
(SPIN) module, which performed reasoning over graphs constructed on spatial and interaction spaces projected from the feature maps. In this way, spatial and topology information are better utilized to improve road detection performance.
Generative adversarial network (GAN) is also a popular tool for map generation tasks \cite{ganganguli2019geogan,ganhe2020generative,gankang2019transferring}, but usually, these works output maps in raster format, such as styled map tiles. Thus, post-processing is still needed to vectorize maps.

After obtaining the probabilistic map of a road network, post-processing should be conducted to extract the road centerline, filter out outlier road segments and fix incorrect disconnections of the road network. To better extract the road-network graph, in \cite{newbulatov2017chain}, Bulatov \textit{et al.} proposed a similarity criteria to fuse raw road segments into chains and designed a two-stage algorithm to optimize the obtained road chains. This work can handle curve roads and sharp circle arcs well. Wenzel \textit{et al.} \cite{newwenzel2019simultaneous} further enhanced \cite{newbulatov2017chain} to generate longer and more accurate road network chains by proposing an iterative greedy optimization procedure.

Since semantic segmentation only works on pixel-level predictions, the topology information is not effectively considered. Thus, the road network graph obtained by segmentation-based approaches tends to have poor topology correctness. Moreover, handcrafted or heuristic post-processing algorithms cannot effectively correct errors in the obtained road network graph.

\subsection{Graph-based approaches}
Different from segmentation-based approaches, graph-based approaches for road network graph detection can directly output the graph structure. Most of the graph-based works detect the road network graph by iterative graph generation \cite{bastani2018roadtracer,li2018polymapper,tan2020vecroad,he2020sat2graph,belli2019image}. RoadTracer \cite{bastani2018roadtracer} is believed to be the first work that iteratively generates the road network graph. In this work, ground-truth initial vertices were used to initialize the iteration. From each initial vertex, RoadTracer predicted the direction of the adjacent vertices of the current vertex as a multi-class classification problem, then moved the agent in the predicted direction by a fixed length. RoadTracer presented much better topology performance than past segmentation-based approaches, but it failed to detect road intersections with high quality due to the fixed step length. Tan \textit{et al.} \cite{tan2020vecroad} solved this problem by replacing the direction prediction with heatmap prediction, where heatmap demonstrated the probabilistic distribution of adjacent vertices of the current vertex. After obtaining the heatmap, the authors extracted local peaks as the predicted adjacent vertices. Although this approach had dynamic step length and presented superior performance to past works, it can not be optimized in an end-to-end way due to the post-processing of the heatmap, which degrades the final performance. In addition, this work cannot distinguish vertices that are close to each other.

Different from aforementioned iterative graph-based approaches, Song \textit{et al.} \cite{he2020sat2graph} proposed Sat2Graph to directly predict the road network graph. Taken as input an aerial image, Sat2Graph predicted a $19$ dimension tensor. This high dimension tensor contained all the information of the road network graph so that the graph could be decoded from the feature tensor by algorithms proposed in Sat2Graph \cite{he2020sat2graph}. However, Sat2Graph had the isomorphic encoding issue analyzed in \cite{he2020sat2graph}, which made it difficult to supervise during training. Moreover, it cannot distinguish road segments whose intersection angle is small.

\subsection{Graph detection of objects similar to road networks}
There are some tasks that are similar to road network detection, such as the detection of road boundaries \cite{liang2019convolutional,xu2021topo,xu2021csboundary}, road lane lines \cite{homayounfar2018hierarchical,homayounfar2019dagmapper,li2021hdmapnet,yang2019road}, road lane \cite{he2022lane,zhou2021automatic} and road curbs \cite{zhxu2021icurb,xu2021cp}. Even though these works do not work on road network detection, their tasks are similar to ours and some ideas or techniques are also inspiring to us. Xu \textit{et al.} \cite{zhxu2021icurb} first proposed to analyze the graph detection problem using imitation learning, and designed a DAgger-based system for road curb detection following the DAgger algorithm \cite{ross2011reduction}. Homayounfar \textit{et al.} proposed DagMapper \cite{homayounfar2019dagmapper} to detect road lane line graph in the point cloud map on the highway. DagMapper can predict the direction of adjacent vertices of the current vertex, and whether the agent should create a new lane line branch when lane line intersections were encountered. Although these works could be inspiring for our task, they cannot handle the road network graph detection task since road networks have much more complicated topology (e.g., road split, road merge and crossroads).

\subsection{Transformer-based detection}
In recent years, transformer \cite{vaswani2017attention} has been receiving more and more attention since its powerful parallelization capacity and great ability to handle sequential tasks. Considering these properties, Carion \textit{et al.} \cite{carion2020end} proposed Detection by Transformer (DETR) for one-shot 2D object detection, which is anchor free and can be trained in an end-to-end way. After extracting image features by a CNN backbone, DETR sent the obtained features as well as multiple object queries to a transformer, and then obtained the bounding box coordinates of objects. Each bounding box was encoded by a 4D embedding. Therefore, taken as input an image, DETR can directly output the coordinates of object bounding boxes. Xu \textit{et al.} \cite{xu2021line} adapted DETR to line segment detection task and named the new model as Line Segment Detection by Transformer (LETR). In this paper, the authors encoded each line segment as an embedding and predicted the encoding embedding by a DETR-based network. Similarly, Can \textit{et al.} \cite{can2021structured} modified the DETR network and pursued to detect lane centerlines. Each lane centerline was fitted by a B-spline and each B-spline was encoded by an embedding. In this way, the authors could directly detect all lane centerlines at one time by predicting the embedding that encoded B-spline information. Our work RNGDet is also inspired by DETR, while the output embedding encodes the information of adjacent vertices of the current vertex.

\subsection{Imitation learning}
Imitation learning aims to train a decision-making agent network to mimic an expert. The most commonly used imitation learning algorithm is behavior cloning \cite{osa2018algorithmic}. In behavior cloning, an expert generates a set of demonstrations and the agent tries to learn the policy of the expert. This algorithm is efficient but it suffers from the drifting problem of imitation learning \cite{ross2010efficient}. Ross \textit{et al.} proposed a meta-algorithm called Dataset Aggregation (DAgger) \cite{ross2011reduction} which could cover a much larger state space to relieve the drifting problem. However, the DAgger algorithm presents quite a low sampling efficiency. Based on DAgger, Xu \textit{et al.} designed approaches to detect road elements by imitation learning in \cite{zhxu2021icurb,xu2021topo}. In this paper, since behavior cloning can already achieve satisfactory performance, considering the low-efficiency performance of DAgger, we adopt behavior cloning in our experiments.

\section{The Proposed Approach}
\begin{figure*}[t]
\centering
    \includegraphics[width=\linewidth]{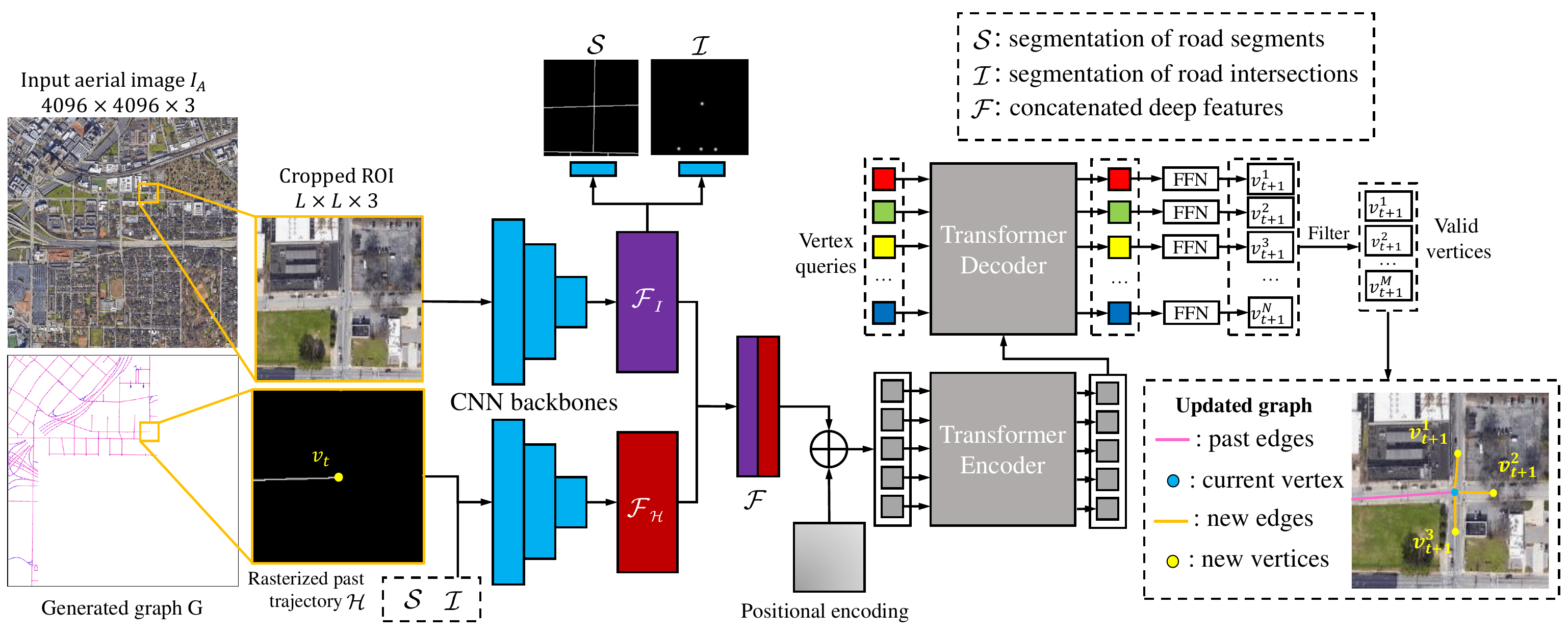}
  \caption{The system overview of RNGDet. The figure shows a single step in the graph detection process of RNGDet. In this example, RNGDet processes a road intersection. Suppose the agent is at $v_t$ at the current time step (denoted by a yellow node in $\mathcal{H}$), we crop an ROI on the input aerial image $I_A$ and rasterize the graph generated so far $G$ within the ROI as $\mathcal{H}$. A CNN backbone network predicts the feature $\mathcal{F}_I$ of the ROI. Then, $\mathcal{F}_I$ is sent to the segmentation heads to predict $\mathcal{S}$ and $\mathcal{I}$. Taken as input $\mathcal{S}$, $\mathcal{I}$ and $\mathcal{H}$, another CNN network predicts the deep feature tensor $\mathcal{F}_{\mathcal{H}}$ and concatenates it with $\mathcal{F}_I$ as the final feature tensor $\mathcal{F}$. The transformer predicts the adjacent vertices $\mathcal{V}=\{v_{t+1}^i\}_{i=1}^M$ of $v_t$. In the updated graph of this example, there are three predicted adjacent vertices $\{v_{t+1}^i\}_{i=1}^3$ (yellow nodes), and they are connected with the current vertex $v_t$ (blue node) by three new edges (orange lines). The graph $G$ is generated in this way iteratively. This figure is best viewed in color. Please zoom in for details.}
  \label{system diagram}
\end{figure*}
\subsection{Approach overview}
This work aims to detect the road network graph from aerial images, and the road network graph can be used for real-world applications (e.g., autonomous vehicle navigation). Suppose the input is a large aerial image $I_A$, then the final output should be a graph $G=(V,E)$. $E=\{e_i\}$ is a set of graph edges and each edge $e_i$ represents a road segment. $V=\{v_j\}$ is a set of graph vertices, and each vertex $v_j$ is either one intersection point of some road segments or the endpoint of a broken road. 

Based on the DETR structure, our proposed RNGDet detects the road network graph by iterations. Starting from a pre-predicted candidate initial vertex in $C=\{c_k\}_{k=1}^{K}$, RNGDet iteratively generates the road network graph by controlling an agent exploring the road network. During iterations, the history trajectory of the agent is recorded by the graph $G$. At each step, centering at the current position of the agent $v_t$, RNGDet crops an ROI $\in \mathbb{R}^{3\times L\times L}$ on $I_A$ and rasterizes $G$ within the ROI as $\mathcal{H} \in \mathbb{R}^{1\times L\times L}$. A CNN backbone network is utilized to extract the deep visual feature of the ROI as $\mathcal{F}_I$ which is sent to the segmentation heads to predict the road segment segmentation map $\mathcal{S}$ and the road intersection segmentation map $\mathcal{I}$. The candidate initial vertices in $C$ can be obtained by finding local peaks of $\mathcal{I}$. After concatenating $\mathcal{H}$ with $\mathcal{S}$ and $\mathcal{I}$, another CNN backbone network is used to extract the deep feature $\mathcal{F}_{\mathcal{H}}$. Then, $\mathcal{F}_{\mathcal{H}}$ and $\mathcal{F}_I$ are fused as the input feature tensor of the transformer. 

Taking $N$ vertex queries $Q=\{q_i\}_{i=1}^N$ as input, the transformer decoder directly predicts $N$ vertex embeddings encoding the valid probability and coordinates of vertices. Each vertex query is a learned embedding, which could be treated as a slot used by RNGDet to make the prediction of one vertex. Thus the number of vertex queries $N$ must be larger than the largest number of vertices needed to be predicted. After filtering out vertices with low probability, we obtain $M (M\leq N)$ valid vertices that are adjacent to $v_t$ as a set $\mathcal{V}=\{v_{t+1}^i\}_{i=1}^M$. If $M=0$, RNGDet pops a new candidate initial vertex from $C$ and repeats the above process; if $M=1$, the agent moves to the predicted coordinate and repeats the above process; if $M>1$, the agent pushes all the predicted vertices into set $C$, pops one vertex from set $C$ and then repeats the above process. When $C$ is empty, RNGDet stops and outputs the final road network graph. The overall pseudocode of our system is shown in Alg. \ref{alg1}. The system diagram of RNGDet is displayed in Fig. \ref{system diagram}.

\begin{algorithm}[!h]
\KwInput{An aerial image $I_A$}
\KwOutput{The road network graph $G=(V,E)$}
\Begin{
$C \gets find\_initial\_vertex(I_A)$ \\
Initialize $G$ as an empty graph\\
\While{$C$ not empty}{
        $t\gets 0$\\
        $v_t\gets C.pop()$\\
        \While{true}{
        $\mathcal{F}\gets CNN(I_A,G,v_t)$\\
        $\{\hat{v}_{t+1}^i\}_{i=1}^N \gets Transformer(\mathcal{F}|Q)$\\
        $\mathcal{V}\gets filter(\{\hat{v}_{t+1}^i\}_{i=1}^N)$\\
        \uIf{$|\mathcal{V}|=0$}{break}
        \uElseIf{$|\mathcal{V}|=1$}{
            $v_t\gets \hat{v}_{t+1}^1$\\
            $t\gets t+1$\\
            update $G$\\
        }
        \uElseIf{$|\mathcal{V}|>1$}{
            $C\gets C\cup \mathcal{V}$\\
            update $G$\\
            break\\
            }
        }
   }

return $G$}
\caption{RNGDet}
\label{alg1}
\end{algorithm}

\subsection{CNN backbone and segmentation}
The input of RNGDet is a large RGB aerial image $I_A\in \mathbb{R}^{3\times 4096\times 4096}$. Since RNGDet requires the history information for iterative graph generation, we maintain the graph $G$ recording the past trajectory of the agent. Due to the size of input images, RNGDet crops a ROI$\in \mathbb{R}^{3\times L\times L}$ ($L$ is 256 in our experiment) on $I_A$. A multi-layer convolutional neural network (CNN) is utilized to extract the deep feature of the ROI as $\mathcal{F}_I$, and the CNN backbone in this paper is ResNet \cite{he2016deep}. Based on the extracted deep feature $\mathcal{F}_I$, two feature paradigm network (FPN) \cite{lin2017feature} segmentation heads predict the segmentation map $\mathcal{S}$ and $\mathcal{I}$, where $\mathcal{S}$ demonstrates the distribution of road segments and $\mathcal{I}$ shows the distribution of road intersection points as well as endpoints of broken roads. Both segmentation tasks are binary segmentation.

To obtain the information of past trajectories, we crop $G$ in the same way as cropping the ROI and rasterize the cropped $G$ as $\mathcal{H}\in \mathbb{R}^{1\times L\times L}$ for afterward concatenation. With $\mathcal{S}$, $\mathcal{I}$ and $\mathcal{H}$ as input, another CNN network outputs the feature tensor $\mathcal{F}_{\mathcal{H}}$. Two feature tensors are concatenated together as the final feature tensor $\mathcal{F}$, containing all the information required by the transformer.

\subsection{Transformer architecture}
After obtaining the deep feature tensor $\mathcal{F}$, the transformer predicts the adjacent vertices $\{v_{t+1}^i\}_{i=1}^M$ of the current vertex $v_t$. The feature tensor $\mathcal{F}$ is reduced to a sequence of smaller tensors, and then fixed positional encoding \cite{bello2019attention,parmar2018image} is fused due to the permutation-invariant characteristics of the transformer architecture \cite{carion2020end}. The input and output of the transformer encoder have the same length. 

\begin{figure}[!h]
  \centering
    \includegraphics[width=\linewidth]{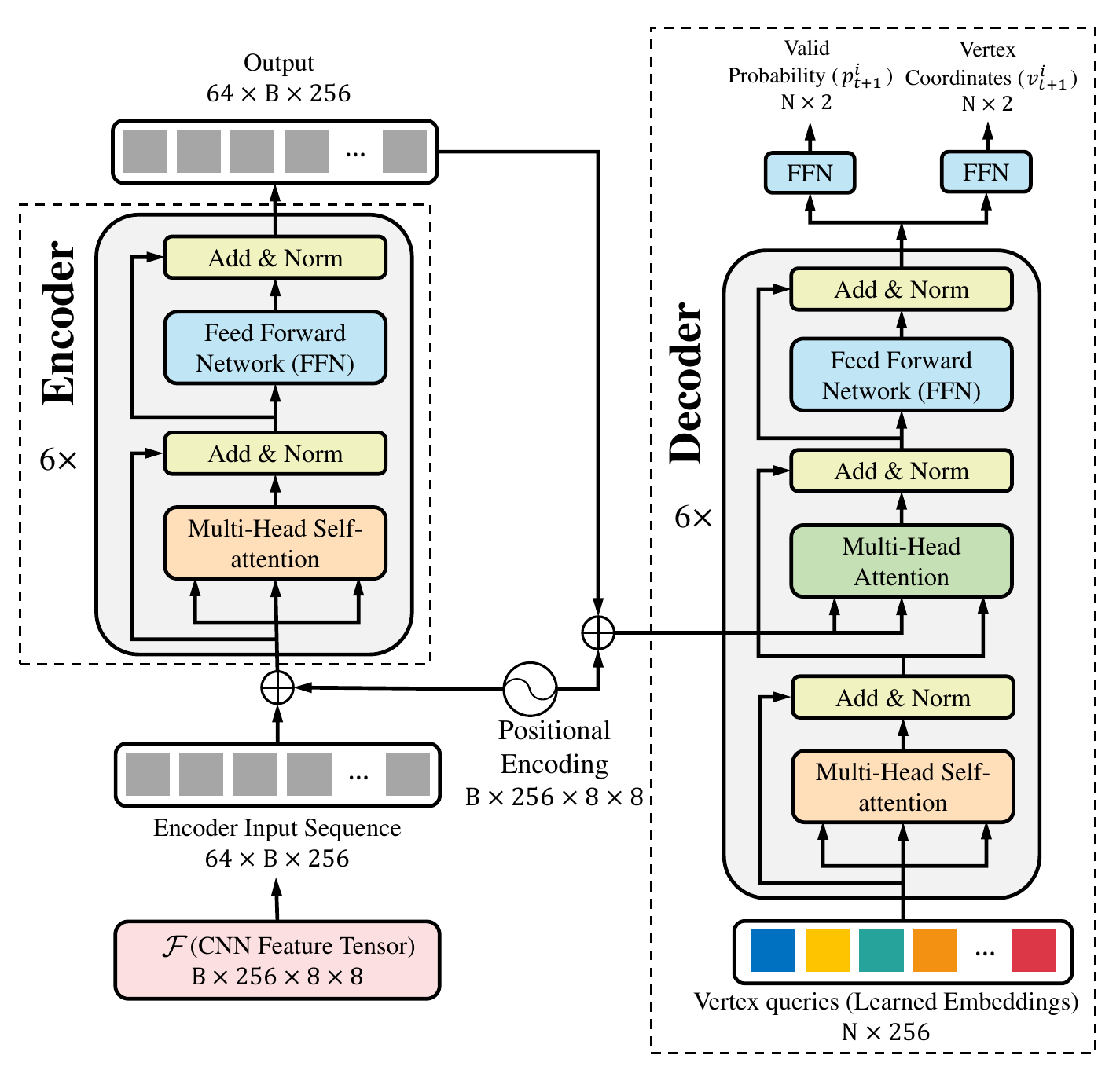}
  \caption{The architecture of the transformer in RNGDet. In this figure, $B$ represents batch size, and $N$ represents the number of vertex queries. Taken as input the predicted CNN feature tensor $\mathcal{F}$, the transformer first splits $\mathcal{F}$ into a sequence of smaller tensors. In our experiment, the $B\times256\times8\times8$-sized feature tensor is split into a 64-length sequence, which contains $B\times256$-sized tensors. Then, the transformer encoder processes the input tensors and outputs a sequence of feature tensors whose shape is the same as that of the input. Finally, based on the encoder output and vertex queries, the transformer decoder predicts $N$ vertices $\{v_{t+1}^i\}_{i=1}^N$, and the valid probability $p_{t+1}^i$ of each vertex $v_{t+1}^i$. Vertices with high enough $p_{t+1}^i$ are regarded as valid vertices and used for graph updates.}
  \label{detail}
\end{figure}

The decoder of the transformer takes in the output sequence of the encoder as well as a set of vertex queries $Q=\{q_i\}_{i=1}^N$, and predicts $N$ embeddings of the adjacent vertices. Each vertex query $q_i$ is a learned embedding and produces one predicted adjacent vertex. In fact, the transformer outputs the Maximum-a-Posterior (MAP) estimation of vertices at the next time step
\begin{equation}
    \{\argmax_{v_{t+1}^i} {p[v_{t+1}^i|\mathcal{F},v_t,q_i]}\}_{i=1}^{N},
\end{equation}
Each output embedding can be decoded into a valid probability $p^i_{t+1}$ and a 2D vertex coordinate $v_{t+1}^i$ by feed-forward networks (FFN). $p^i_{t+1}$ demonstrates the probability that $v_{t+1}^i$ is valid and should be added into the road network graph $G$. Suppose we have $M (M\leq N)$ valid predicted vertices as a set $\mathcal{V}=\{v_{t+1}^i\}_{i=1}^M$, then for each valid predicted vertex $v_{t+1}^i$, we update the road network graph $G$ by adding $v_{t+1}^i$ into $V$ and a new edge connecting $v_{t+1}^i$ with $v_t$ into $E$. The transformer architecture is visualized in Fig. \ref{detail}. 
 
\subsection{Policy for graph generation during inference period}
\begin{figure}[t]
  \centering
    \includegraphics[width=\linewidth]{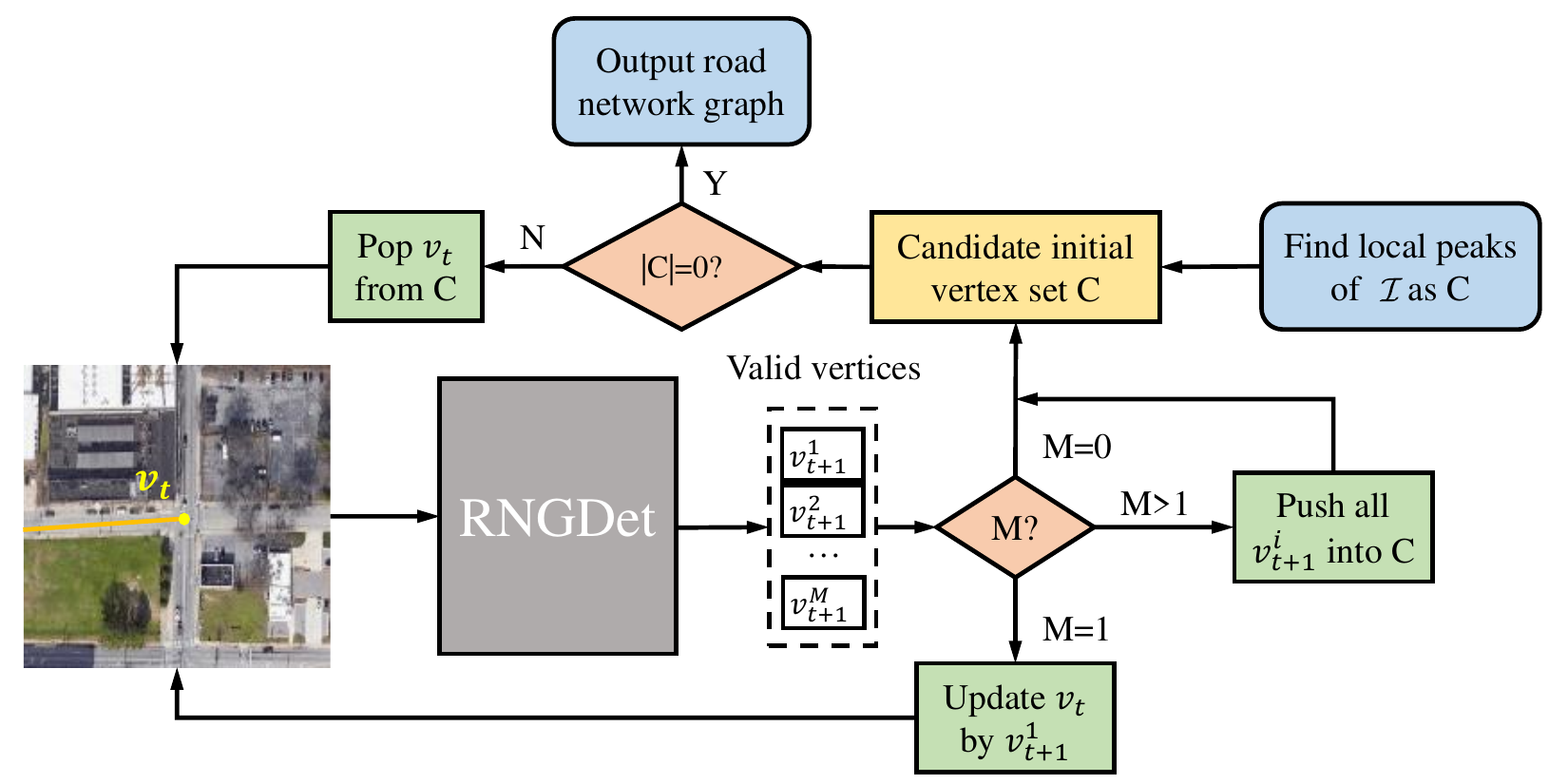}
  \caption{The pipeline of road network graph generation by RNGDet. RNGDet iteratively generates the road network graph vertex-by-vertex. According to the number of valid vertices at each step, RNGDet takes different actions to update the graph. Please zoom in for details.}
  \label{RNGDet pipeline}
\end{figure}

 \begin{figure}[t]

 \centering
    \begin{subfigure}[t]{0.117\textwidth}
        \begin{subfigure}[t]{\textwidth}
            \includegraphics[width=\textwidth]{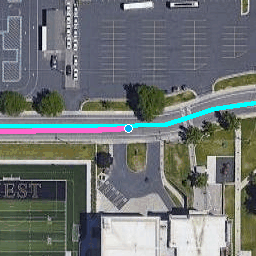}
        \end{subfigure}\vspace{.6ex}
        \begin{subfigure}[t]{\textwidth}
            \includegraphics[width=\textwidth]{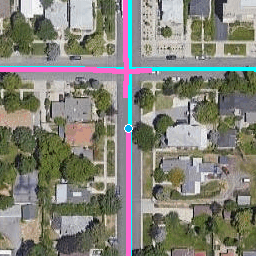}
        \end{subfigure}\vspace{.6ex}
        \begin{subfigure}[t]{\textwidth}
            \includegraphics[width=\textwidth]{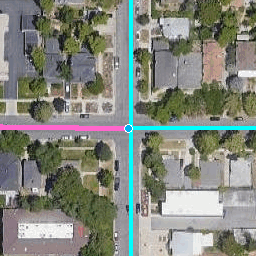}
        \end{subfigure}\vspace{.6ex}
        \caption{Step $t$}
    \end{subfigure}
    \begin{subfigure}[t]{0.117\textwidth}
        \begin{subfigure}[t]{\textwidth}
            \includegraphics[width=\textwidth]{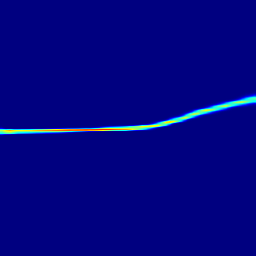}
        \end{subfigure}\vspace{.6ex}
        \begin{subfigure}[t]{\textwidth}
            \includegraphics[width=\textwidth]{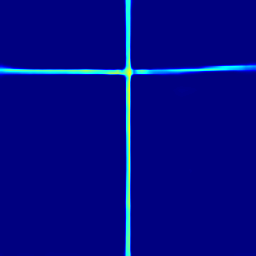}
        \end{subfigure}\vspace{.6ex}
        \begin{subfigure}[t]{\textwidth}
            \includegraphics[width=\textwidth]{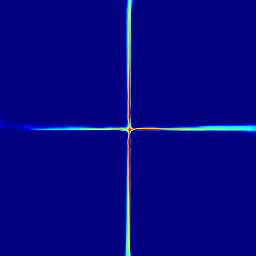}
        \end{subfigure}\vspace{.6ex}
        \caption{$\mathcal{S}$}
    \end{subfigure}
    \begin{subfigure}[t]{0.117\textwidth}
        \begin{subfigure}[t]{\textwidth}
            \includegraphics[width=\textwidth]{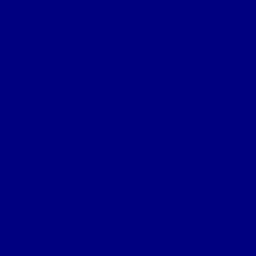}
        \end{subfigure}\vspace{.6ex}
        \begin{subfigure}[t]{\textwidth}
            \includegraphics[width=\textwidth]{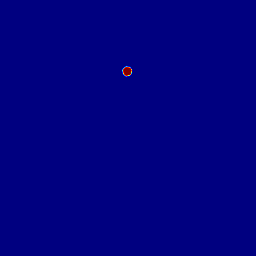}
        \end{subfigure}\vspace{.6ex}
        \begin{subfigure}[t]{\textwidth}
            \includegraphics[width=\textwidth]{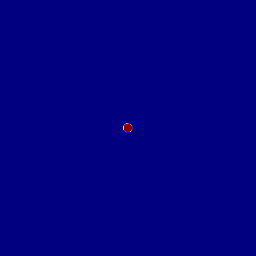}
        \end{subfigure}\vspace{.6ex}
        \caption{$\mathcal{I}$}
    \end{subfigure}
    \begin{subfigure}[t]{0.117\textwidth}
        \begin{subfigure}[t]{\textwidth}
            \includegraphics[width=\textwidth]{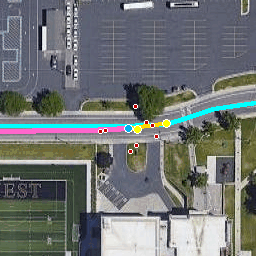}
        \end{subfigure}\vspace{.6ex}
        \begin{subfigure}[t]{\textwidth}
            \includegraphics[width=\textwidth]{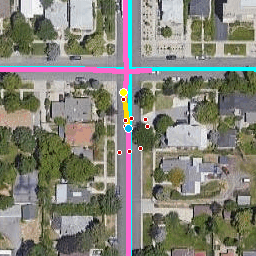}
        \end{subfigure}\vspace{.6ex}
        \begin{subfigure}[t]{\textwidth}
            \includegraphics[width=\textwidth]{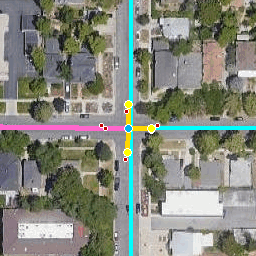}
        \end{subfigure}\vspace{.6ex}
        \caption{Step $t+1$}
    \end{subfigure}
    \caption{Visualization of graph updating. Each row represents an example that shows how RNGDet iteratively generates the road network graph. (a) ROI of the current step $t$. The cyan lines are the ground-truth road network, the pink lines demonstrate history graph and the blue node is $v_t$. (b) The segmentation map of road segments (i.e., $\mathcal{S}$). (c) The segmentation map of road intersections (i.e., $\mathcal{I}$). (d) Updated graph at the next time step $t+1$. Yellow nodes are valid predicted adjacent vertices $v_{t+1}^i$ while red nodes are invalid ones. Valid vertices will be used to update the road network graph. Orange lines are the predicted new edges connecting $v_t$ with $v_{t+1}^i$. This figure is best viewed in color. Please zoom in for details.}
    \label{fig_examples}
\end{figure}
During the inference period, RNGDet generates the road network graph by iterations. First, we initialize the candidate initial vertex set $C$ by finding local peaks of the road intersection segmentation map $\mathcal{I}$. Then, RNGDet pops one vertex $v_t$ from the set $C$. Centering at $v_t$, RNGDet crops input images and predicts valid vertices at the next time step as $\mathcal{V}$. Based on the number of valid vertices $|\mathcal{V}|=M$, RNGDet takes different actions to update the road network graph: (1) $M=0$. It means there is no road ahead, thus RNGDet should stop processing the current road and turns to work on other roads if $C$ is not empty. (2) $M=1$. This happens when RNGDet travels along a single road. RNGDet adds $v_{t+1}^1$ into $V$ and edge $(v_t,v_{t+1}^1)$ into $E$, and then moves to $v_{t+1}^1$. RNGDet keeps updating the graph in this way until intersections or broken roads are met. (3) $M>1$. This indicates that RNGDet encounters road intersections and needs to generate new vertices in multiple directions. RNGDet will update the graph, push all vertices in $\mathcal{V}$ to $C$, and pop one candidate initial vertex from $C$. 

RNGDet keeps running the above iterations to generate the road network graph vertex-by-vertex. If and only if the candidate initial vertex set $C$ is empty, RNGDet stops and outputs the generated road network graph $G=(V,E)$. The pipeline of the road network graph generation process of RNGDet is visualized in Fig. \ref{RNGDet pipeline}. Some example visualizations are shown in Fig. \ref{fig_examples}.

\begin{figure}[t]
  \centering
    \includegraphics[width=\linewidth]{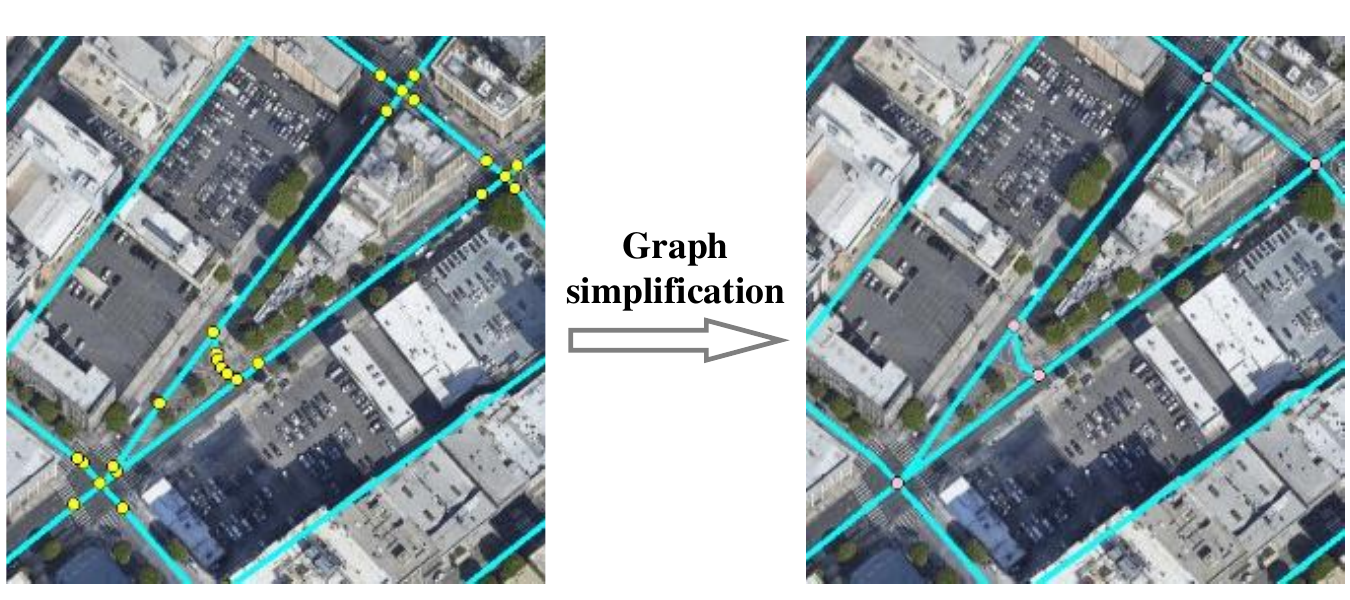}
  \caption{Visualization of the simplification of the ground-truth road network graph. In the raw ground-truth graph, there are various kinds of vertices with different degrees (yellow nodes in the left subfigure). For training label calculation, we remove vertices whose degrees are two. After filtering, only intersection vertices (degrees larger than 2) and endpoint vertices (degrees are 1) remain (pink nodes in the right subfigure). The road connecting two adjacent vertices is defined as a road segment. The agent must finish exploring a road segment before switching to other road segments. In this way, we can guarantee the correctness of automatically generated labels. This figure is best viewed in color.}
  \label{vertex extraction}
\end{figure}
 
\subsection{Training label calculation}\label{expert}
Based on the ground-truth road network graph and the current location of the agent $v_t$, we can automatically generate the training label for RNGDet (i.e., $\mathcal{S}^*$,$\mathcal{I}^*$,$\mathcal{V}^*=\{(v_{t+1}^i)^*\}_{i=1}^M$). The segmentation labels can be simply cropped from the ground-truth segmentation masks centering at $v_t$. For $\mathcal{V}^*$, we need to obtain coordinates of the ground-truth vertices at the next time step. To achieve that, we (1) simplify the ground-truth road network graph by removing vertices whose degree is 2 and (2) generate the label vertices at the next time step based on the ground-truth road network graph.

The raw ground-truth graph consists of various kinds of vertices, such as endpoint vertices (degrees are 1), vertices in the middle of roads (degrees are 2) and intersection vertices (degrees are larger than 2). Among them, vertices whose degrees are 2 are not uniquely defined and could be removed without harming the road network topology. Therefore, for simplicity, we remove these vertices from the ground-truth graph. This process is visualized in Fig. \ref{vertex extraction}. After the graph simplification, there are only endpoint vertices and intersection vertices remaining.

\begin{figure}[t]
    \begin{subfigure}[t]{0.5\textwidth}
        \begin{subfigure}[t]{0.24\textwidth}
            \includegraphics[width=\textwidth]{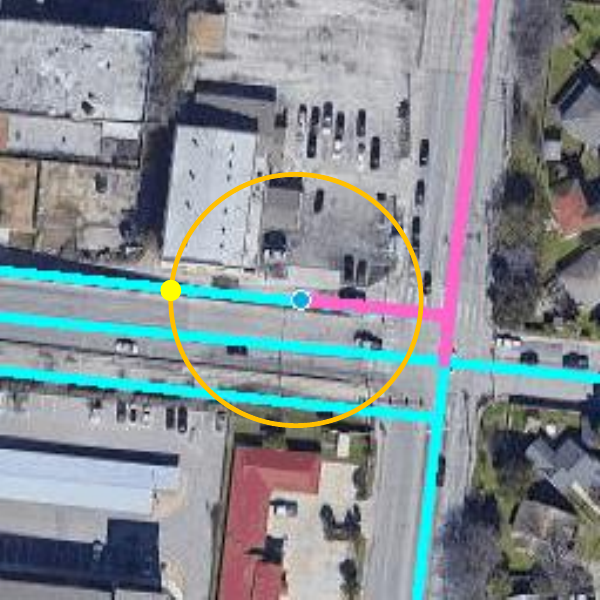}
            \caption{}
        \end{subfigure}
        \vspace{.6ex}
        \begin{subfigure}[t]{0.24\textwidth}
            \includegraphics[width=\textwidth]{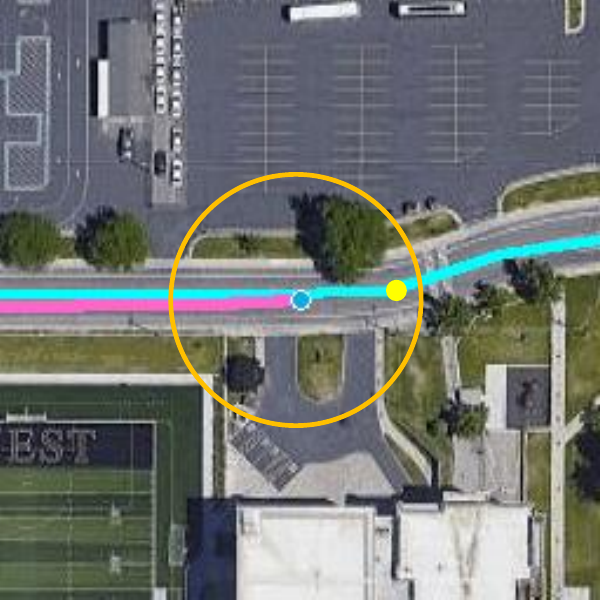}
            \caption{}
        \end{subfigure}
        \vspace{.6ex}
        \begin{subfigure}[t]{0.24\textwidth}
            \includegraphics[width=\textwidth]{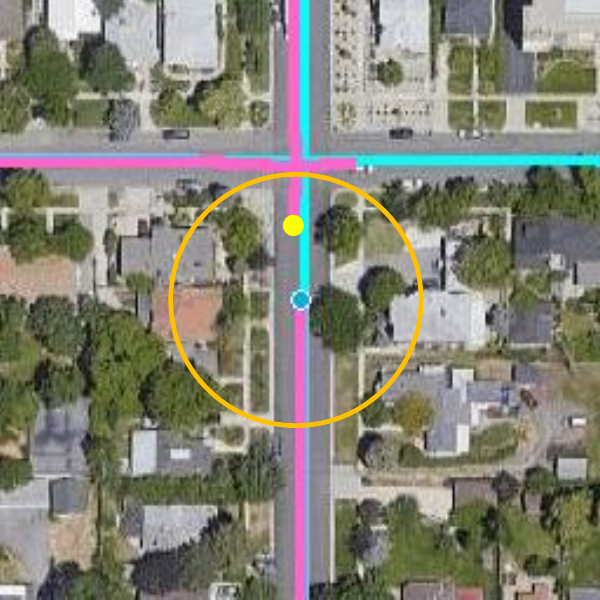}
            \caption{}
        \end{subfigure}
        \begin{subfigure}[t]{0.24\textwidth}
            \includegraphics[width=\textwidth]{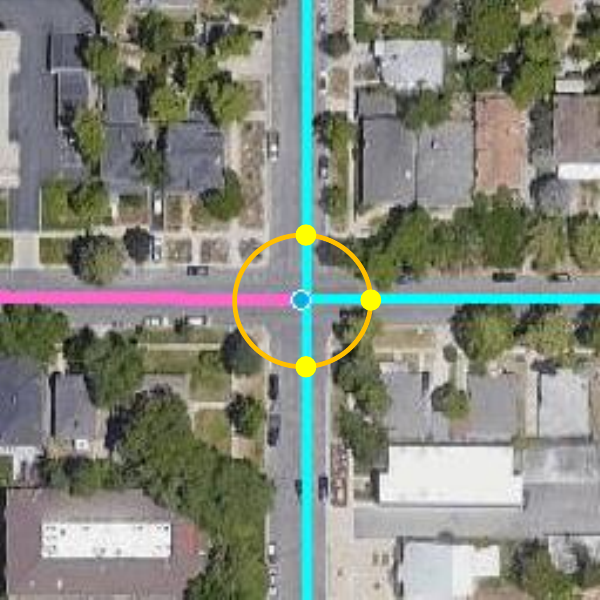}
            \caption{}
        \end{subfigure}
    \end{subfigure}
        
    \caption{Examples of training label generation. Cyan lines are ground-truth road network, pink lines are history road network, blue nodes represents $v_t$ and $(v_{t+1}^i)^*$ is denoted by yellow nodes. The radiant of the orange circle in (a)-(c) is $\tau$ while the radiant of the orange circle in (d) is $\tau'$. $\tau'$ is smaller than $\tau$ in our experiment. (a) Road-segment-mode. The road ahead is straight. (b) Road-segment-mode. The road ahead has turning points with large enough curvature. (c) Road-segment-mode. The agent should connect $v_t$ with a previously generated candidate initial vertex. (d) Intersection-mode. There are three new road segments that are incident to the current intersection point.  This figure is best viewed in color.}
    \label{labels}
\end{figure}
To facilitate the calculation of labels, we define every road connecting adjacent vertices as a road segment. The agent will be either in road-segment-mode or intersection-mode. The road-segment-mode indicates that the agent is currently traveling along a road segment, and there will be only one ground-truth valid vertex at each step (i.e., $\mathcal{V}^*=\{(v_{t+1}^1)^*\}$). Three examples are visualized in subfigure (a), (b) and (c) of Fig. \ref{labels}. To prevent the agent from being trapped in an infinite loop, the training label should encourage the agent to move forward (i.e., $(v_{t+1}^1)^*$ should be far enough from $v_t$ and move to the unexplored part of the current road segment). If the road ahead is straight (i.e., no turning points with large curvature within $\tau$ distance to $v_t$), we select the point whose distance is $\tau$ away from $v_t$ as $(v_{t+1}^1)^*$ (subfigure (a) of Fig. \ref{labels}). If there are some turning points ahead, we find the turning point that is closest to $v_t$ as $(v_{t+1}^1)^*$ (subfigure (b) of Fig. \ref{labels}). If there is some candidate initial vertices within $\tau$ distance to $v_t$, the closest candidate initial vertex is treated as $(v_{t+1}^1)^*$ (subfigure (c) of Fig. \ref{labels}). Note that the above algorithm is only utilized to generate the training label and does not affect the graph updating process during inference.

When the agent finishes exploring the current road segment, it switches to the intersection-mode and finds incident road segments. For each incident road segment, the point whose distance is $\tau'$ away from $v_t$ is defined as the label vertex (subfigure (d) of Fig. \ref{labels}). Usually there will be multiple $(v_{t+1}^i)^*$ when RNGDet is in the intersection-mode (i.e., $\mathcal{V}^*=\{(v_{t+1}^i)^*\}_{i=1}^M$ and $M>1$).


\subsection{Loss functions}
At each step, RNGDet predicts two segmentation maps as well as coordinates and valid probability of vertices at the next step. Thus, in our experiment three loss functions are utilized to train RNGDet. Suppose the predicted segmentation maps are $\hat{\mathcal{S}}$ and $\hat{\mathcal{I}}$ while the ground-truth segmentation masks are $\mathcal{S}^{*}$ and $\mathcal{I}^{*}$, then we have 
\begin{equation}
    \mathcal{L}_{seg} = L(\hat{\mathcal{S}},\mathcal{S}^*) + L(\hat{\mathcal{I}},\mathcal{I}^*),
\end{equation}
where $L$ is focal loss \cite{lin2017focal}. Similarly, suppose the ground-truth vertices at the next step $t+1$ are $\{v^*_j\}_{j=1}^M$ and the predictions are $\{ \hat{v}_i \}_{i=1}^N$, where $M\leq N$. They can be matched by solving a bipartite matching problem through minimizing the following function:
\begin{equation}
    \hat{\sigma}=\argmin_{\sigma}\sum_{i}^{N}\mathcal{L}_{match}(\hat{v}_i,v^*_{\sigma}),
\end{equation}
where $\sigma$ is the index of $v^*$ matched with $\hat{v}_i$, and $\mathcal{L}_{match}$ calculates pair wise Euclidean distance. After the vertex matching, we have the L1 loss as
\begin{equation}
    \mathcal{L}_{coord}=\frac{1}{N}\sum_{i}^{N}|\hat{v}_i-v^*_{\sigma}|.
\end{equation}
RNGDet also predicts the valid probability $\hat{p}_i$ of each $\hat{v}_i$, and only vertices with high enough $\hat{p}_i$ will be used to update the road network graph. The ground-truth value of $\hat{p}_i$ is 1 if $\hat{v}_i$ is matched with a $v^*_{\sigma}$, otherwise the ground-truth value of $\hat{p}_i$ is 0 (i.e., $\hat{v}_i$ does not matched with any $v^*$). Binary cross entropy loss is utilized to optimize $\hat{p}_i$:
\begin{equation}
    \mathcal{L}_{valid}=\text{BCELoss}(\hat{\mathcal{P}},\mathcal{P}^*),
\end{equation}
where $\hat{\mathcal{P}}=\{\hat{p}_i\}_{i=1}^M$ and $\mathcal{P}^*=\{p^*_i\}_{i=1}^M$. The final loss function training RNGDet is the weighted summation of the aforementioned loss functions:
\begin{equation}
    \mathcal{L} = \mathcal{L}_{seg} + \alpha \mathcal{L}_{coord} + \beta \mathcal{L}_{valid},
\end{equation}
where, $\alpha$ and $\beta$ balance the loss function. We have $\alpha=5$ and $\beta=1$ in our experiments.

\subsection{Training data sampling}
To obtain the training dataset for RNGDet, we propose a behavior-cloning-based algorithm to explore the ground-truth road network graph to generate training samples. Suppose the agent is now at $v_t$, we then crop the ROI on $I_A$, crop $\mathcal{S}^*$ on the ground-truth road segment segmentation mask, crop $\mathcal{I}^*$ on the ground-truth road intersection segmentation mask, crop and rasterize $G$ for $\mathcal{H}$, and calculate the ground-truth vertices as well as valid probability at the next time step (i.e., $\mathcal{V}^*$ and $\mathcal{P}^*$) based on aforementioned approaches. Then, one training sample (ROI,$\mathcal{H}$,$\mathcal{S}^*$,$\mathcal{I}^*$,$\mathcal{V}^*$,$\mathcal{P}^*$) is obtained. If the agent is in the road-segment-mode and the ground-truth vertex at the next time step $(v_{t+1}^1)^*$ is an intersection point, we use  $(v_{t+1}^1)^*$ to update the graph (i.e., the agent moves to $(v_{t+1}^1)^*$); otherwise we add $(v_{t+1}^i)^*$ with Gaussian noises $\Delta$ to update the graph (i.e., the agent moves to $(v_{t+1}^i)^*+\Delta$). The agent is driven to explore road networks by this behavior-cloning-based sampling algorithm, and the generated samples are used to train RNGDet.

\section{Experimental Results and Discussions}
\subsection{Dataset}
In this paper, all the experiments are conducted on the RoadTracer dataset \cite{bastani2018roadtracer}. The dataset contains 300 high-resolution aerial images (60cm/pixel) obtained from Google map, and the ground-truth road network graphs are from the OpenStreetMap (OSM). All the data has been converted to the image coordinate system. This dataset covers 40 cities (e.g., Los Angeles and Boston). Each aerial image has 3 channels and is $4096\times4096$-sized. A city may be composed of multiple aerial images. 

\subsection{Implementation}
To obtain the dataset to train RNGDet, we run the proposed data sampling algorithm to explore the ground-truth road network graph of the training aerial images and generate the training samples.
At each step, Gaussian noise is added for graph updating in order to make RNGDet more robust. Finally, the training set contains around 300K samples from different aerial images. We split 10K samples from the training set as the validation data.   

In our experiment, we set the crop size of ROI as 256 (i.e., $L=256$) to trade-off between effectiveness and efficiency. When we generate the training labels, $\tau$ is set as 40 pixels and $\tau'$ is set as 20 pixels. For the transformer, the number of input vertex queries is 10 (i.e., $|Q|=N=10$). RNGDet is trained with a learning rate as $10^{-4}$ and a decay rate as $10^{-5}$ for 50 epochs. We evaluate the performance of RNGDet on the validation set at the end of each epoch. All the experiments are conducted on 4 RTX-3090 GPUs.

\subsection{Baselines}

\begin{figure*}[!th]
 \centering
    \begin{subfigure}[t]{0.135\textwidth}
        \begin{subfigure}[t]{\textwidth}
            \includegraphics[width=\textwidth]{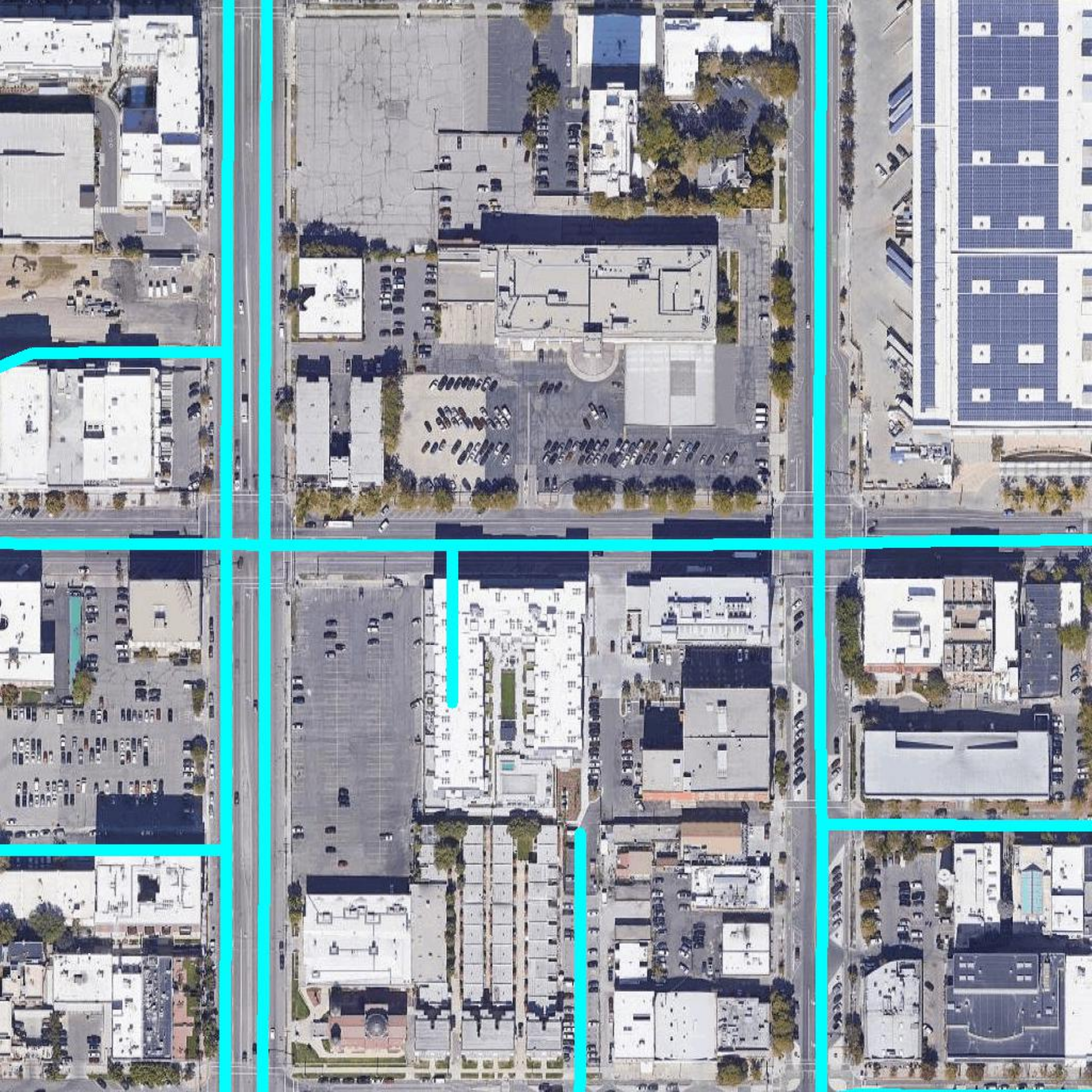}
        \end{subfigure}\vspace{.6ex}
        \begin{subfigure}[t]{\textwidth}
            \includegraphics[width=\textwidth]{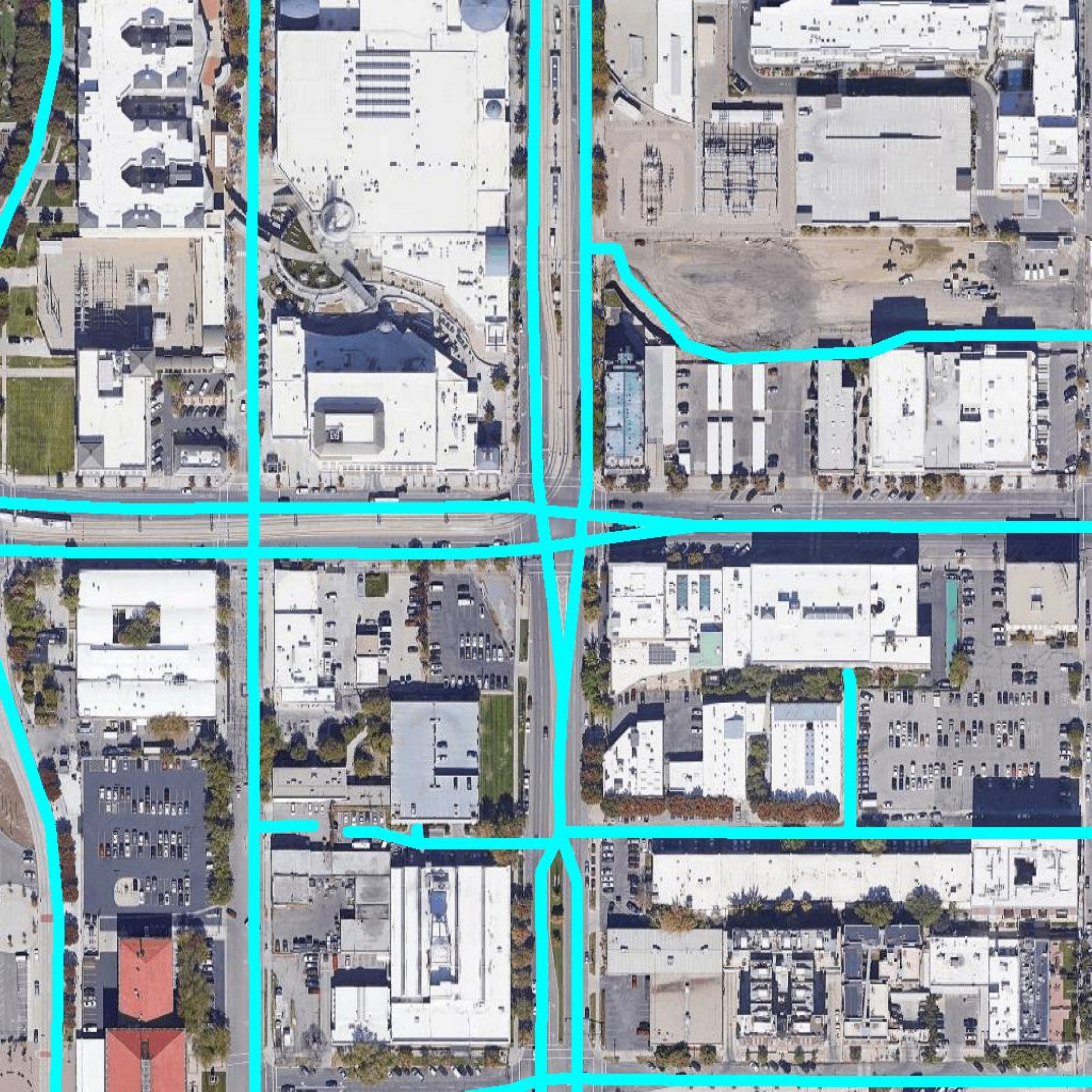}
        \end{subfigure}\vspace{.6ex}
        \begin{subfigure}[t]{\textwidth}
            \includegraphics[width=\textwidth]{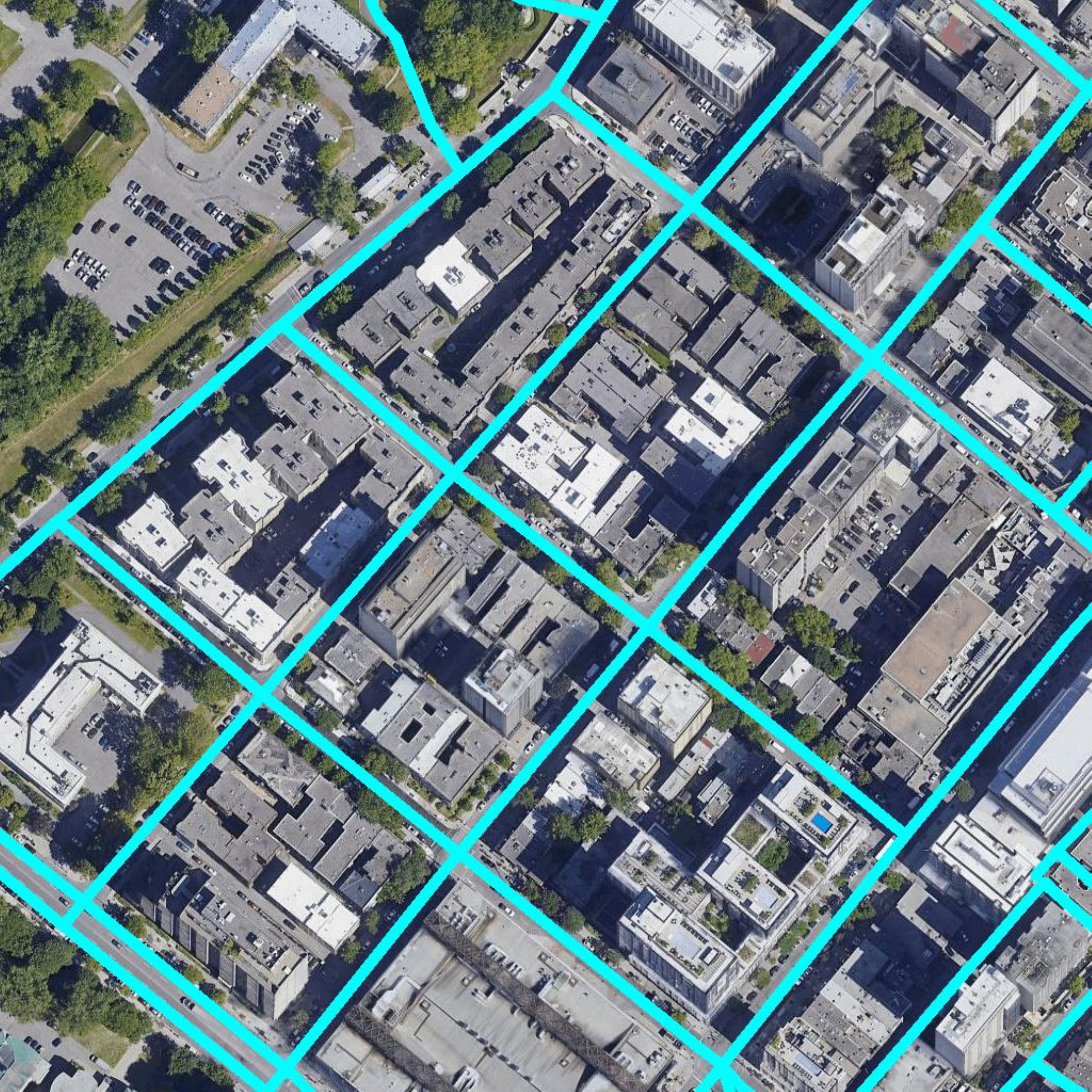}
        \end{subfigure}\vspace{.6ex}
        \begin{subfigure}[t]{\textwidth}
            \includegraphics[width=\textwidth]{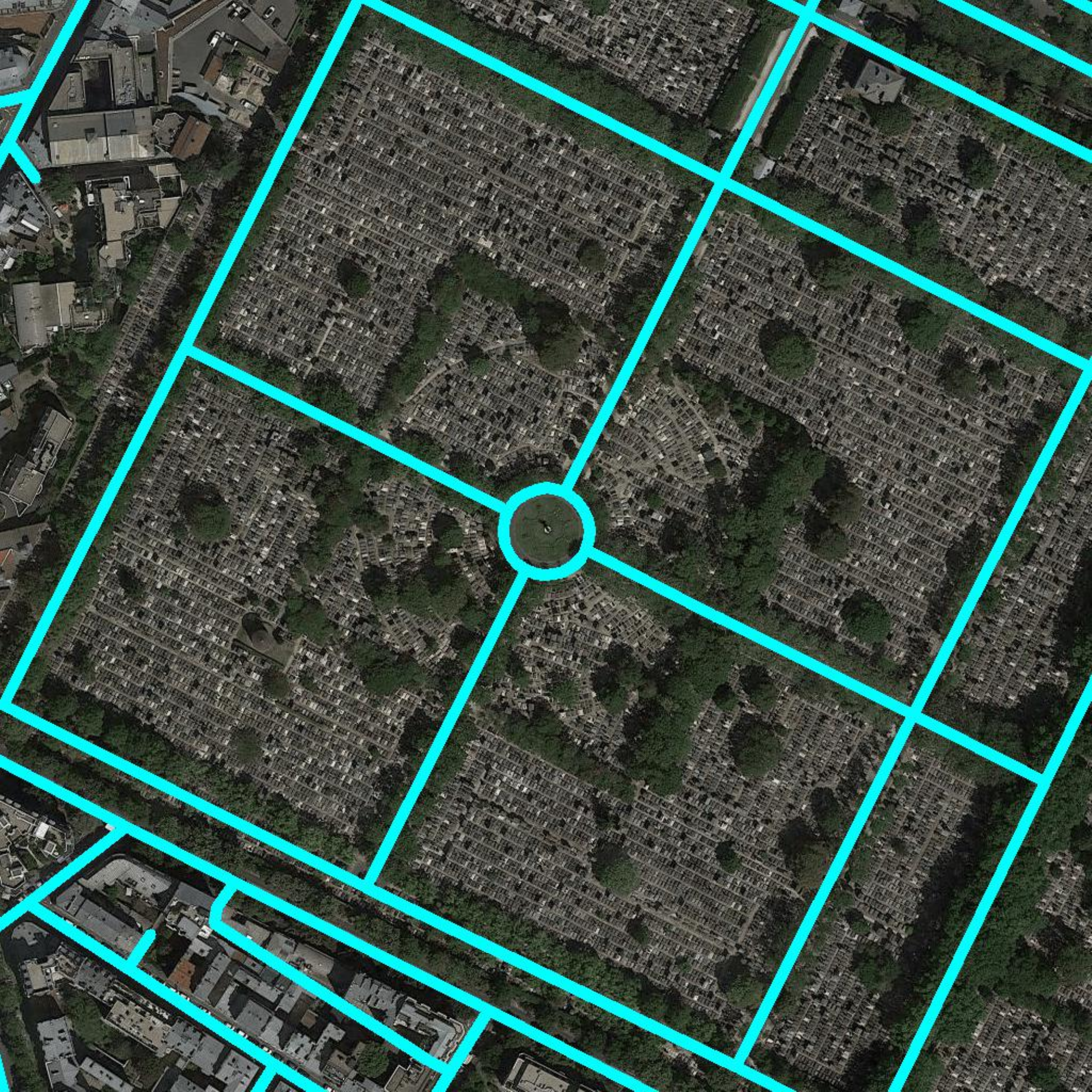}
        \end{subfigure}\vspace{.6ex}
        \begin{subfigure}[t]{\textwidth}
            \includegraphics[width=\textwidth]{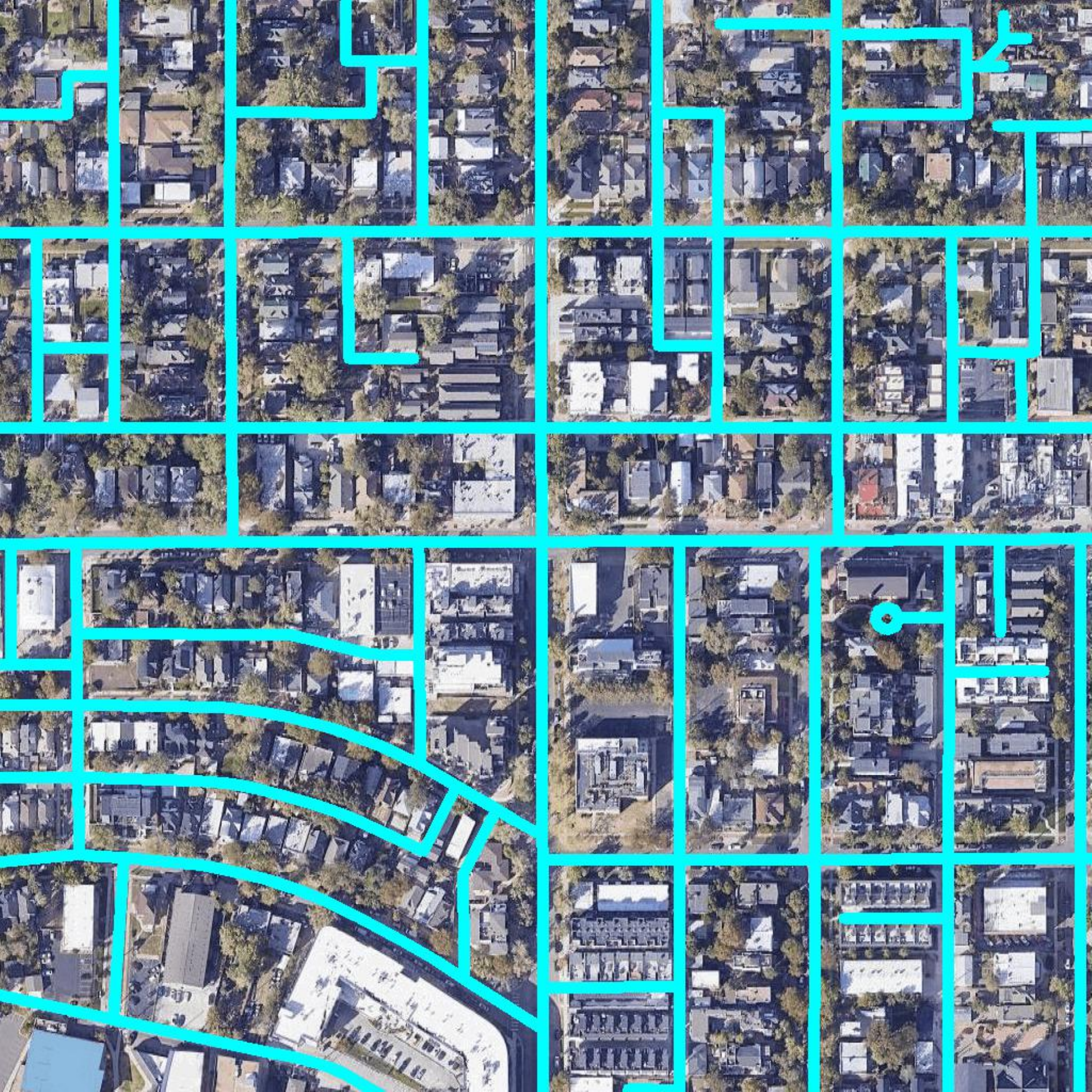}
        \end{subfigure}\vspace{.6ex}
        \begin{subfigure}[t]{\textwidth}
            \includegraphics[width=\textwidth]{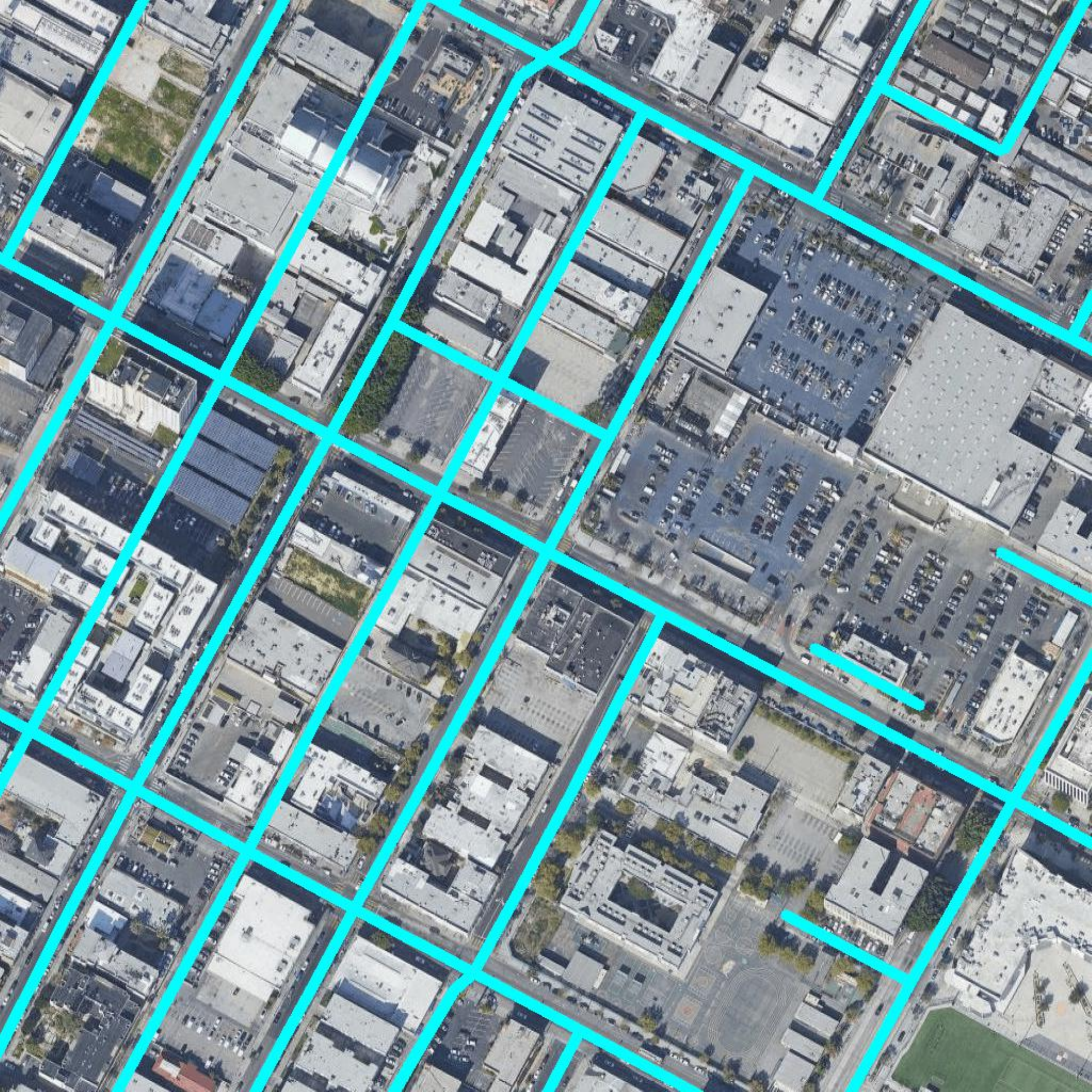}
        \end{subfigure}\vspace{.6ex}
        \begin{subfigure}[t]{\textwidth}
            \includegraphics[width=\textwidth]{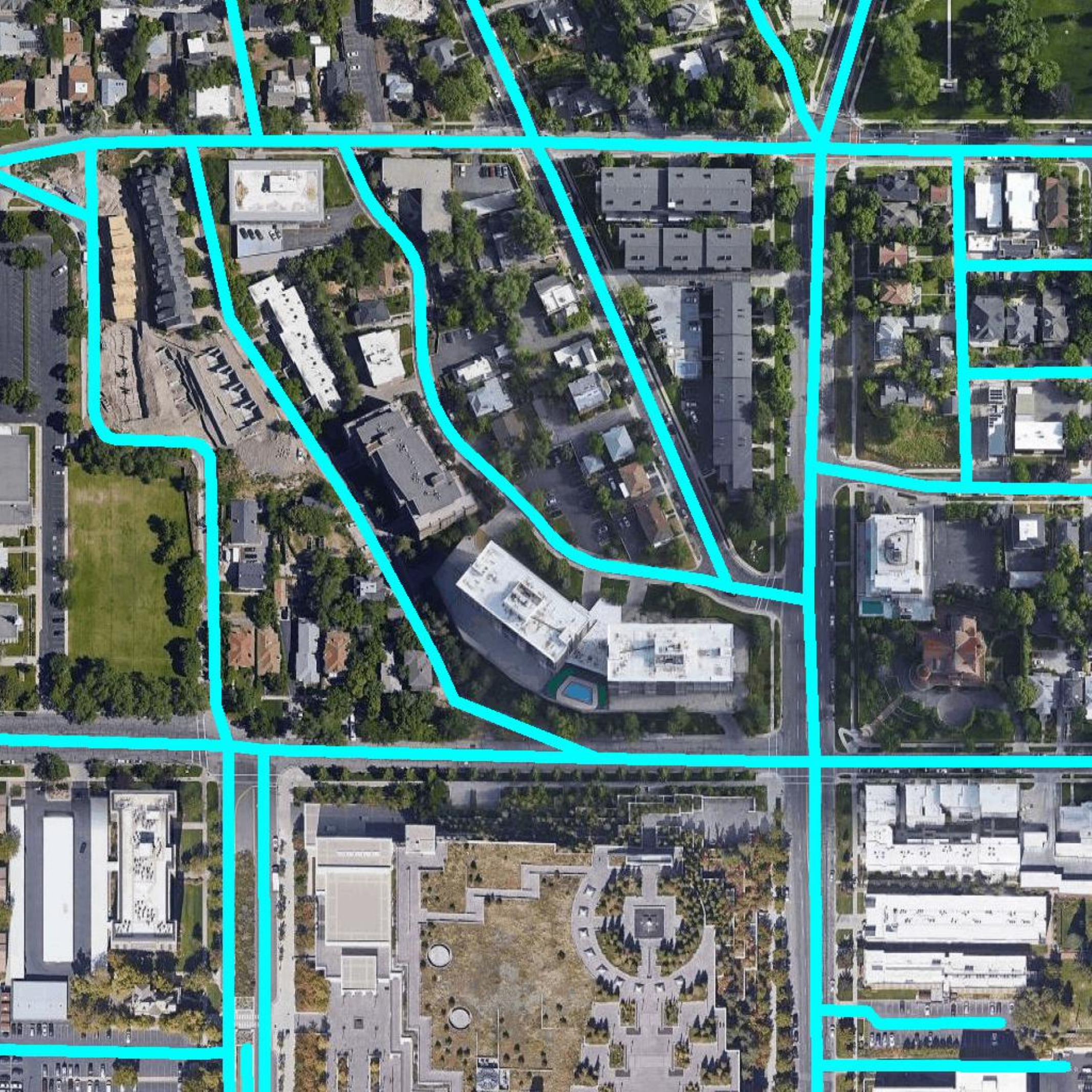}
        \end{subfigure}\vspace{.6ex}
        \caption{Ground truth}
        \label{fig_qualitative_1st}
    \end{subfigure}
    \begin{subfigure}[t]{0.135\textwidth}
        \begin{subfigure}[t]{\textwidth}
            \includegraphics[width=\textwidth]{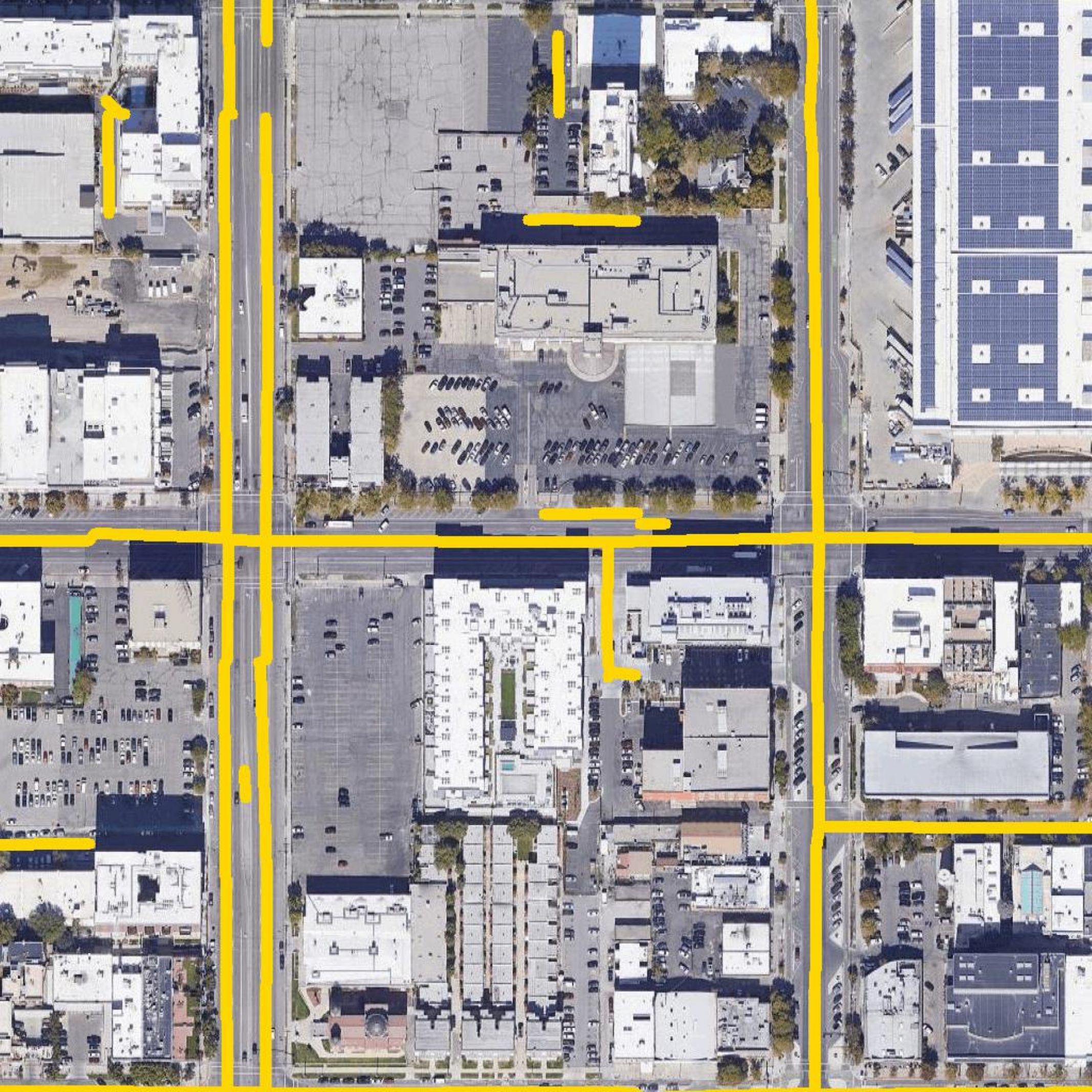}
        \end{subfigure}\vspace{.6ex}
        \begin{subfigure}[t]{\textwidth}
            \includegraphics[width=\textwidth]{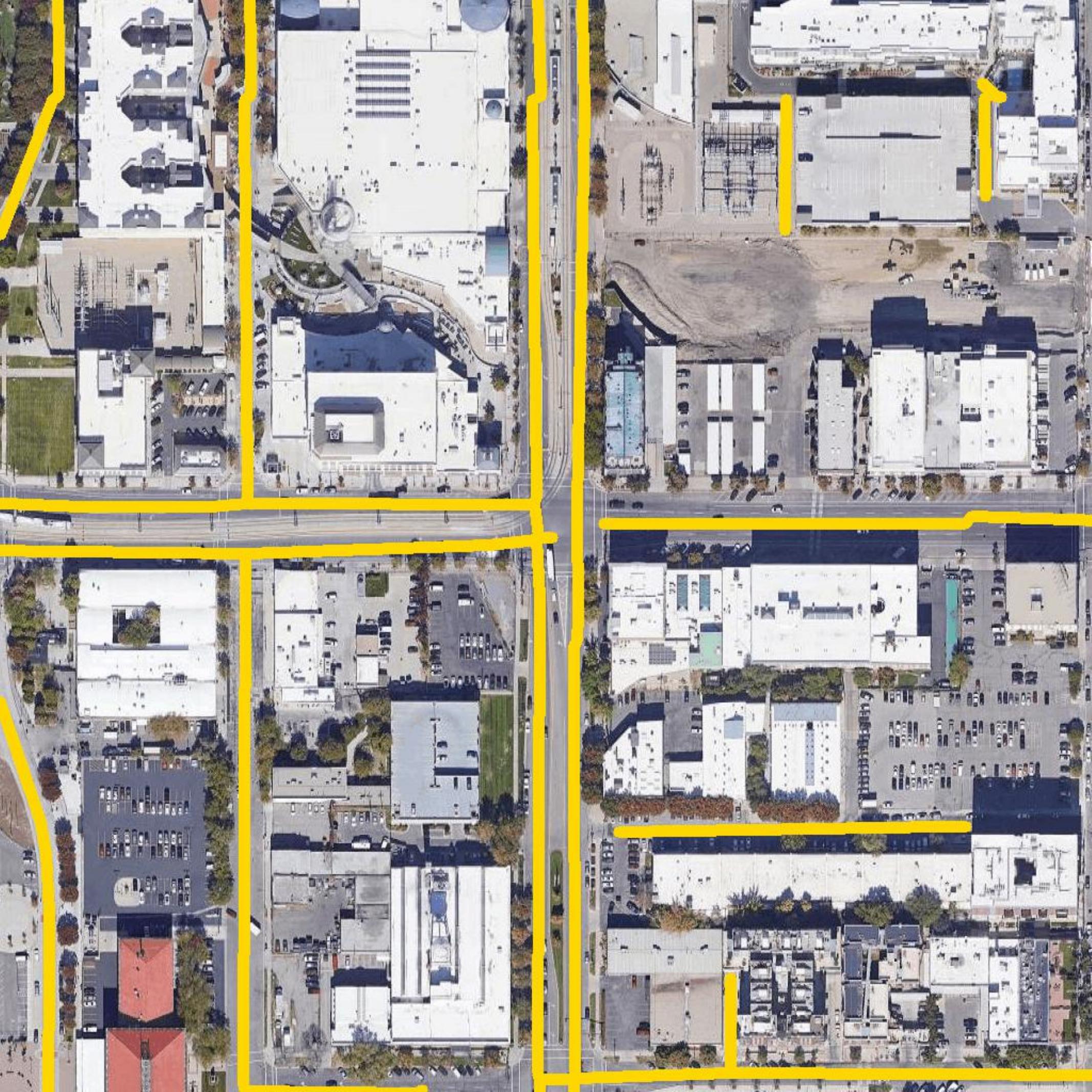}
        \end{subfigure}\vspace{.6ex}
        \begin{subfigure}[t]{\textwidth}
            \includegraphics[width=\textwidth]{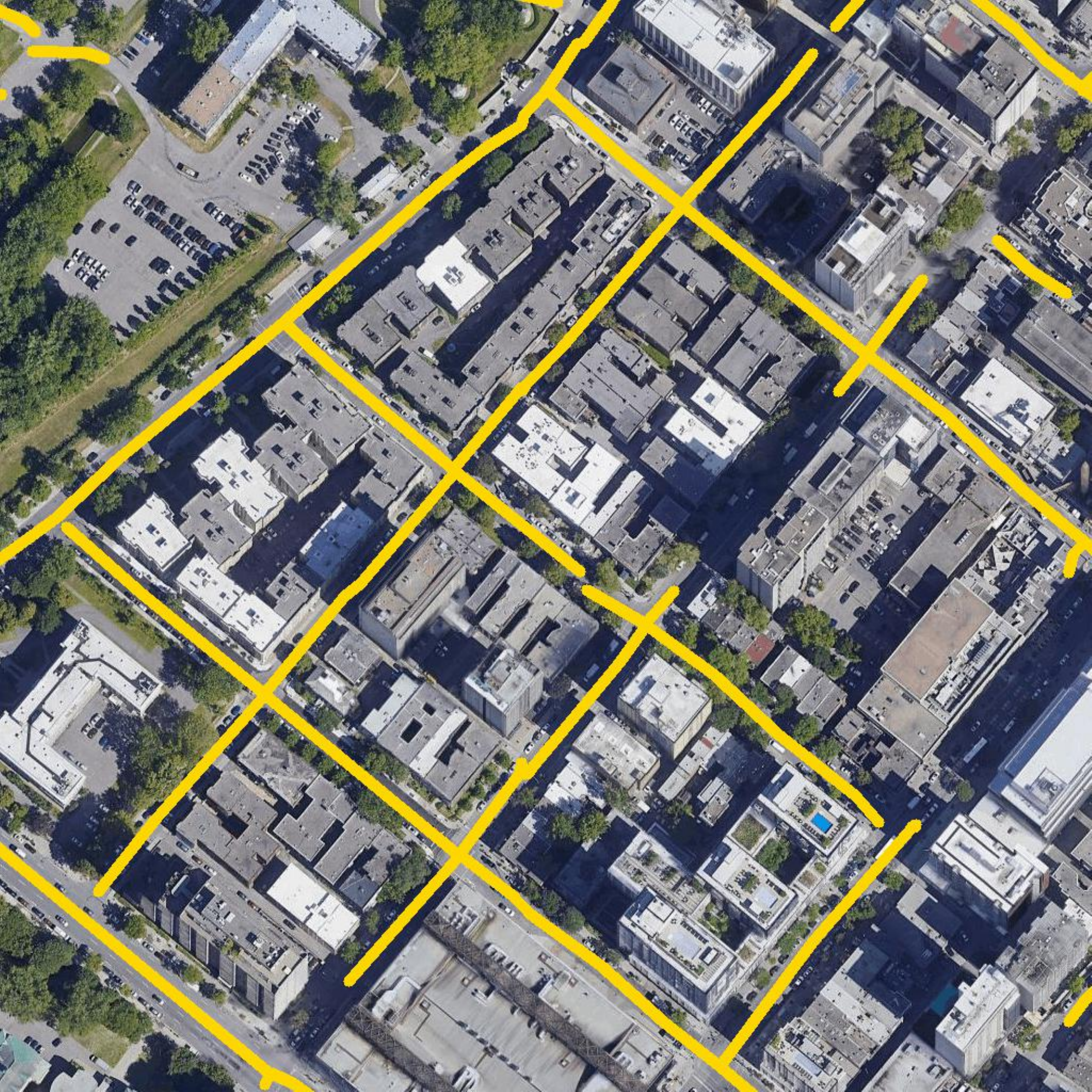}
        \end{subfigure}\vspace{.6ex}
        \begin{subfigure}[t]{\textwidth}
            \includegraphics[width=\textwidth]{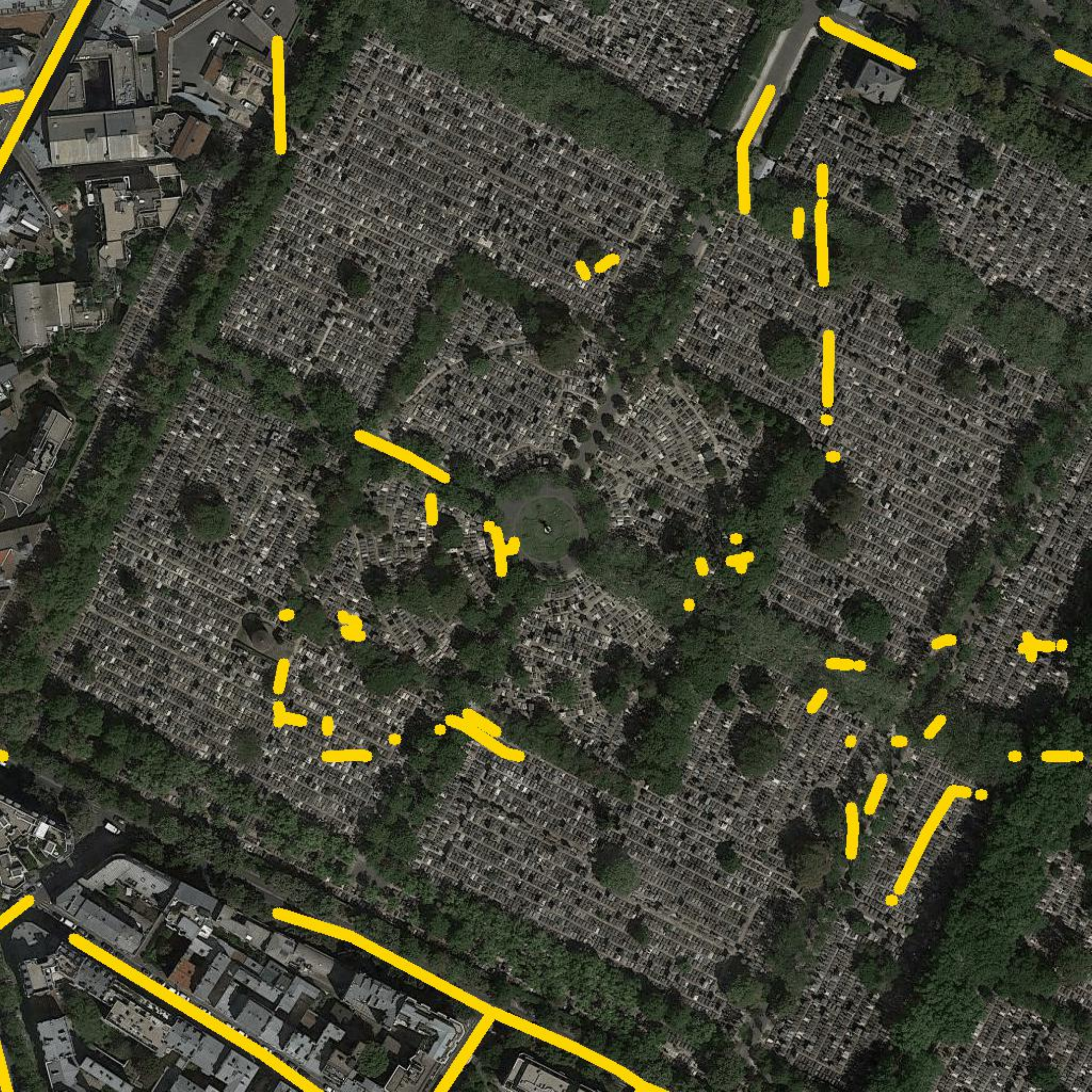}
        \end{subfigure}\vspace{.6ex}
        \begin{subfigure}[t]{\textwidth}
            \includegraphics[width=\textwidth]{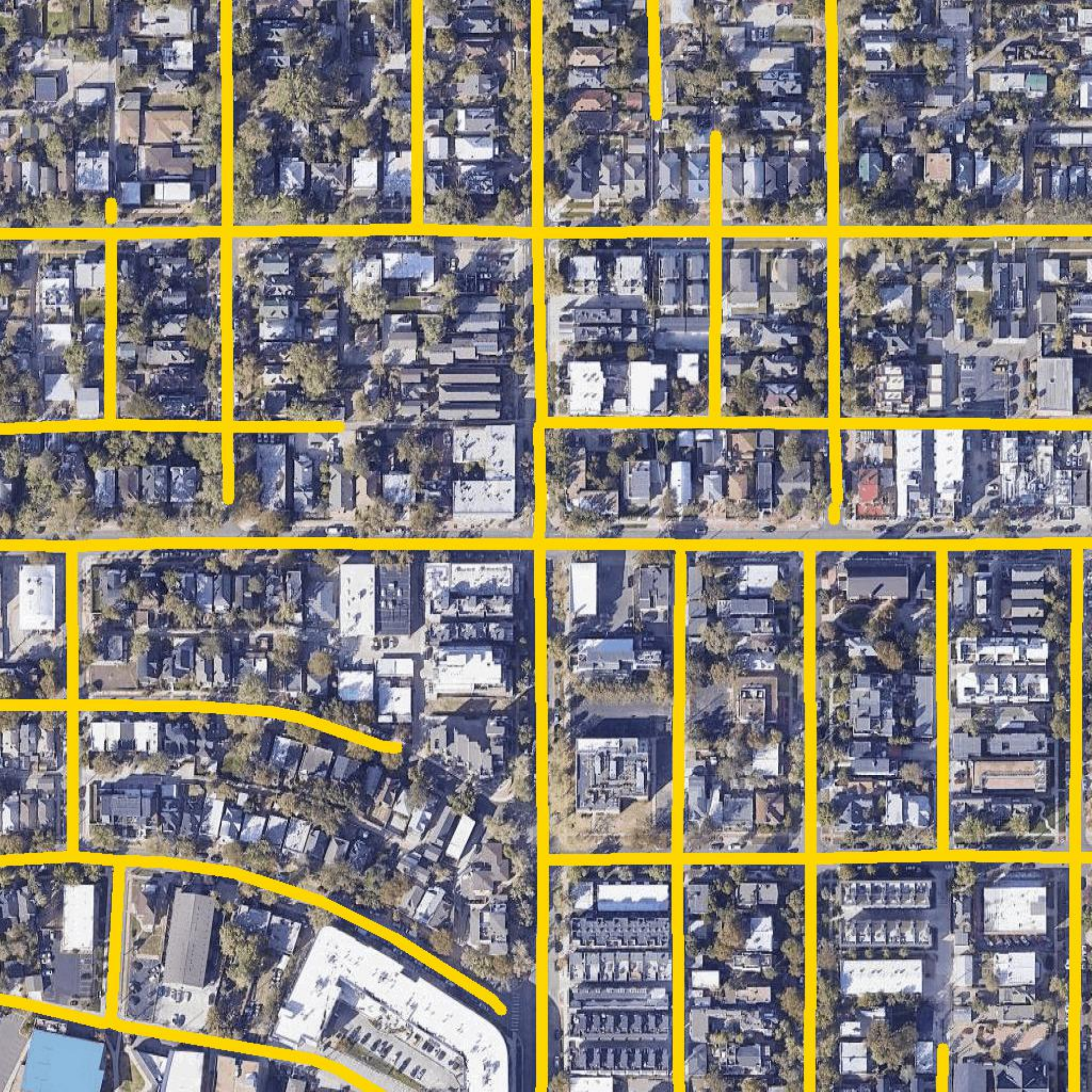}
        \end{subfigure}\vspace{.6ex}
        \begin{subfigure}[t]{\textwidth}
            \includegraphics[width=\textwidth]{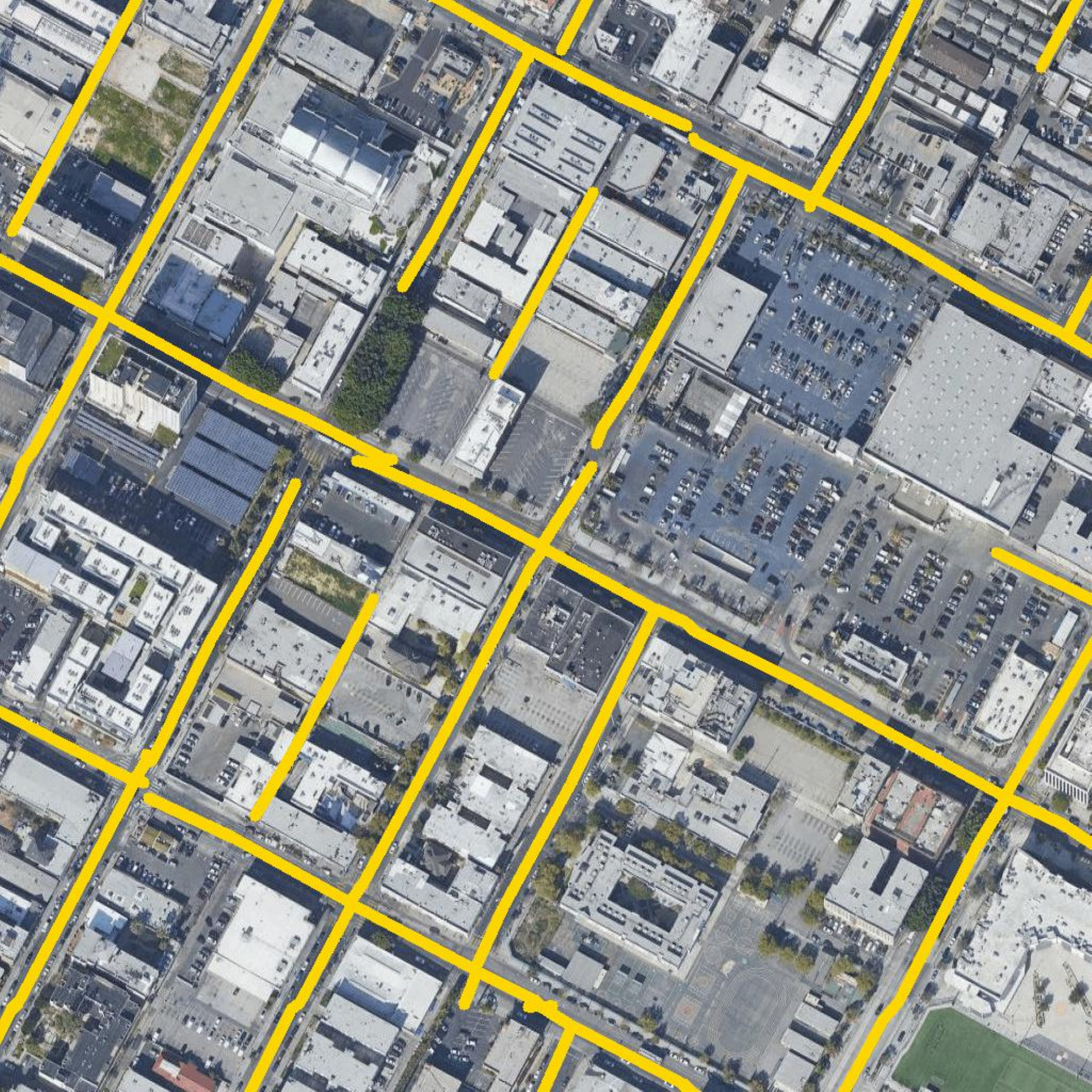}
        \end{subfigure}\vspace{.6ex}
        \begin{subfigure}[t]{\textwidth}
            \includegraphics[width=\textwidth]{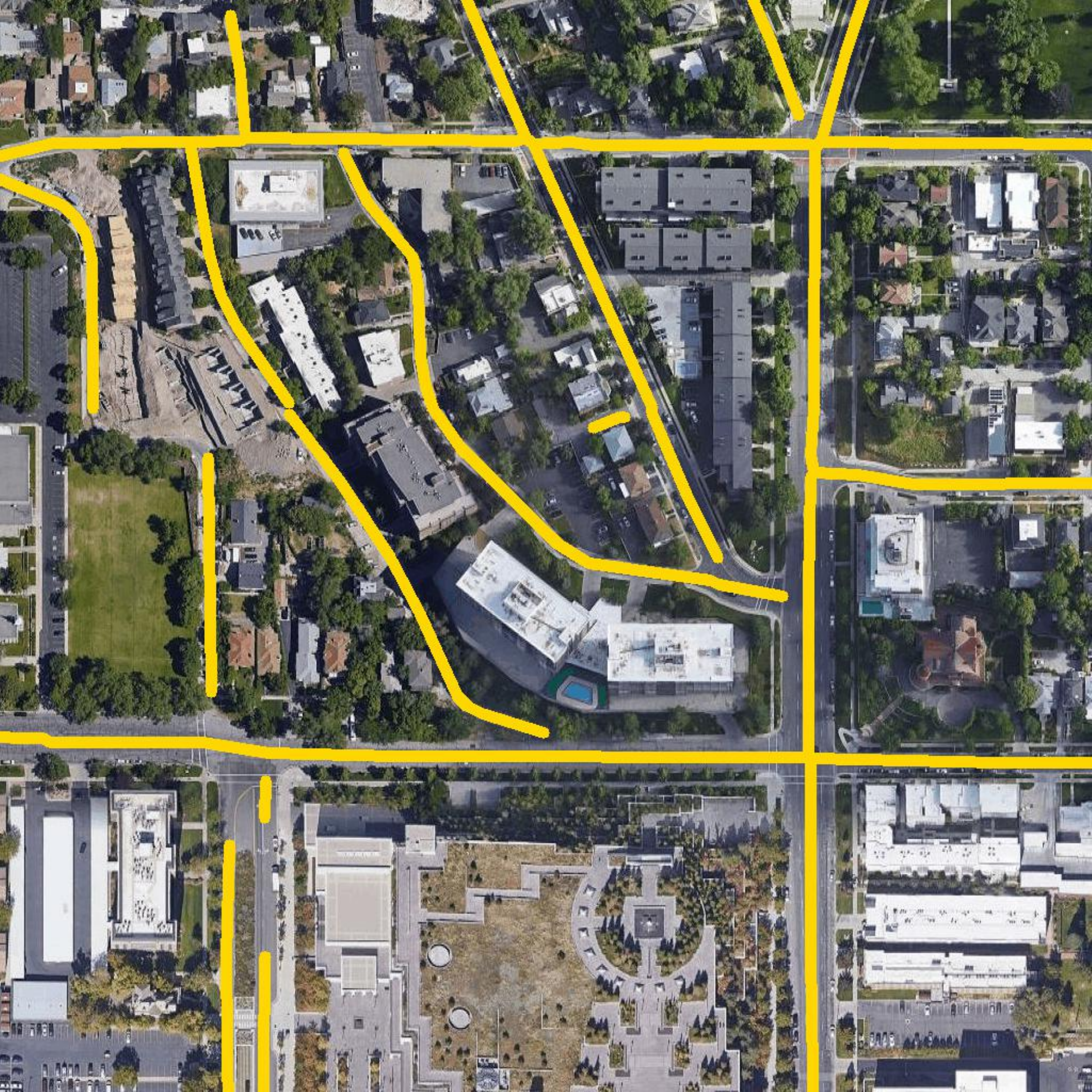}
        \end{subfigure}\vspace{.6ex}
        \caption{\scriptsize{ImprovedRoad \cite{batra2019improved}}}
        \label{fig_qualitative_1st}
    \end{subfigure}
    \begin{subfigure}[t]{0.135\textwidth}
        \begin{subfigure}[t]{\textwidth}
            \includegraphics[width=\textwidth]{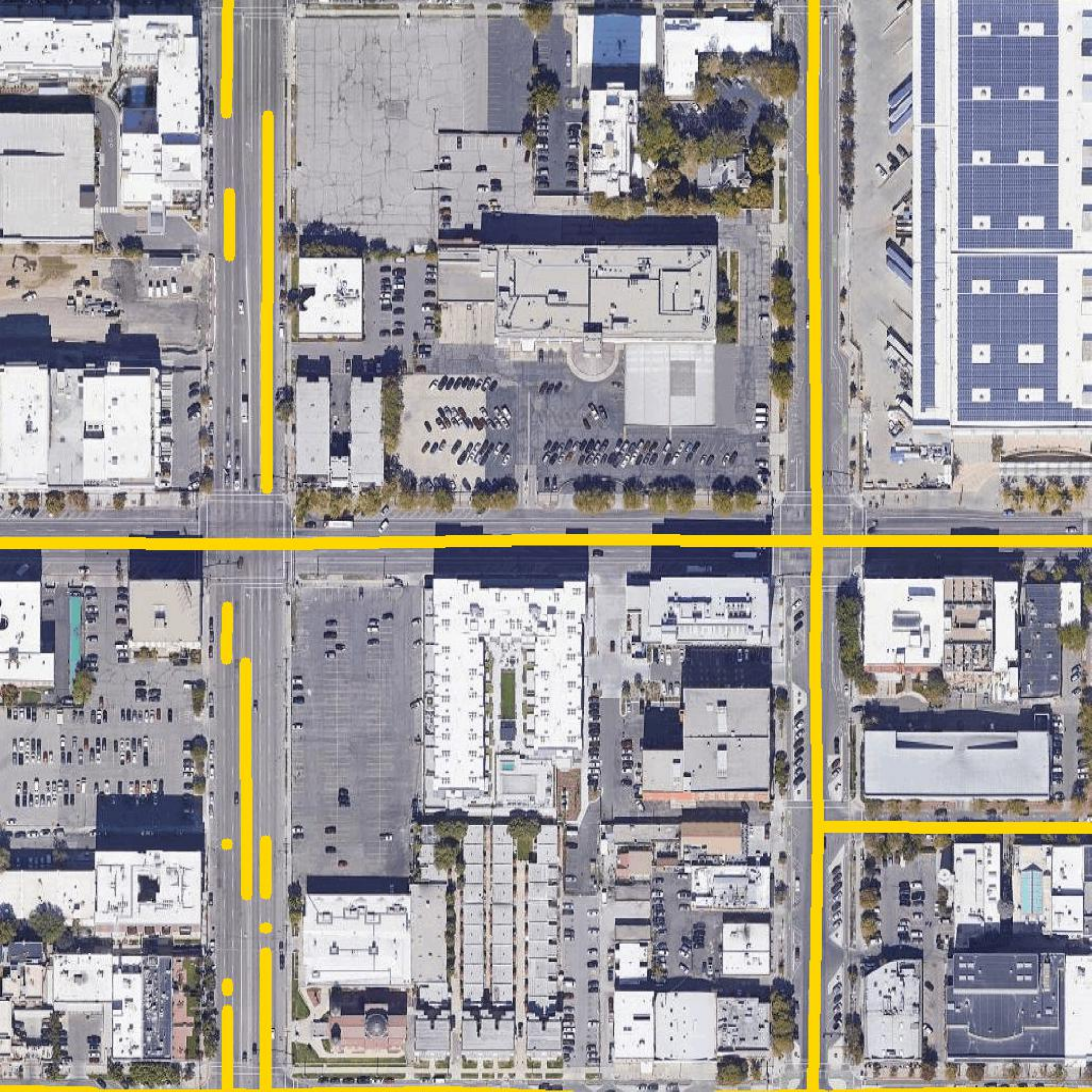}
        \end{subfigure}\vspace{.6ex}
        \begin{subfigure}[t]{\textwidth}
            \includegraphics[width=\textwidth]{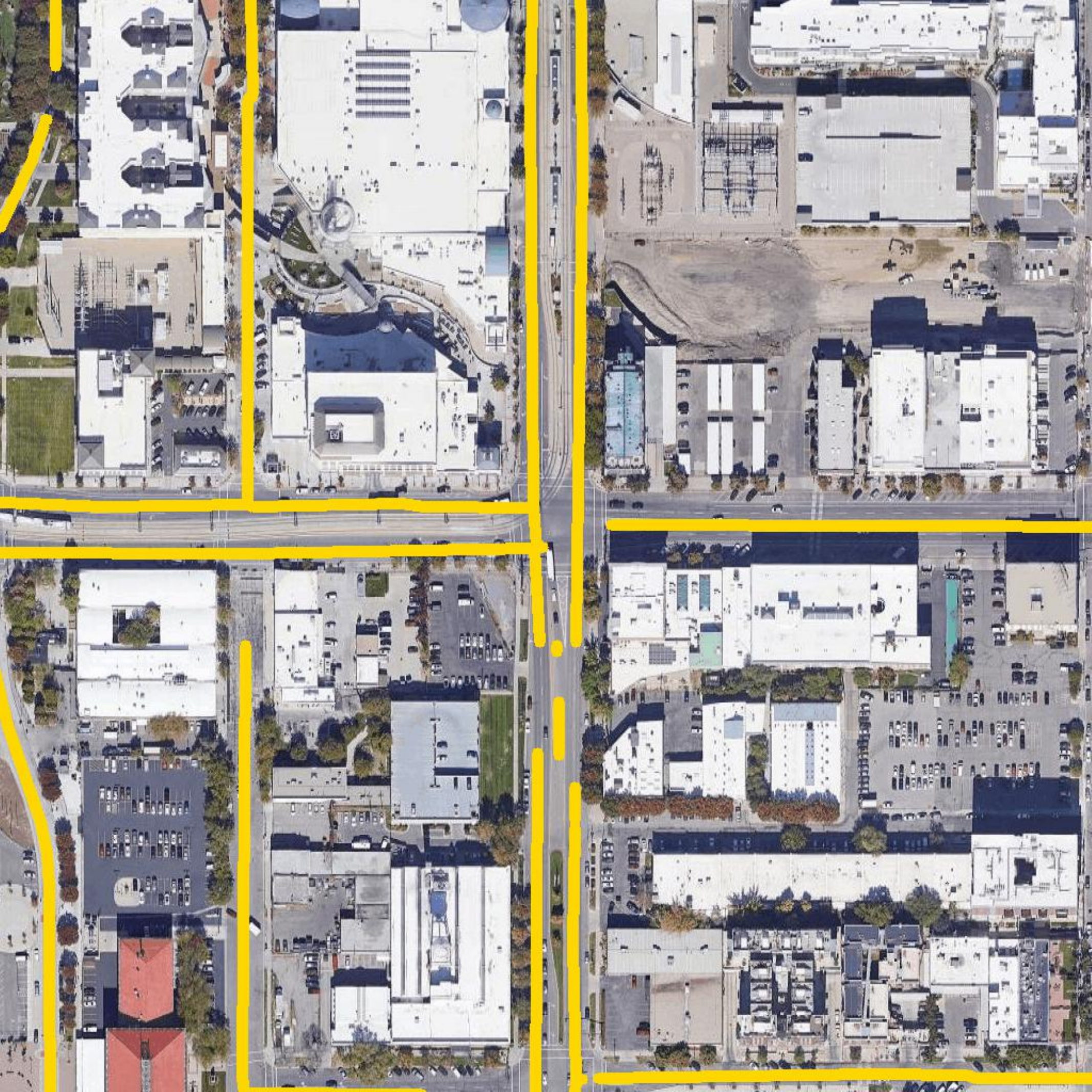}
        \end{subfigure}\vspace{.6ex}
        \begin{subfigure}[t]{\textwidth}
            \includegraphics[width=\textwidth]{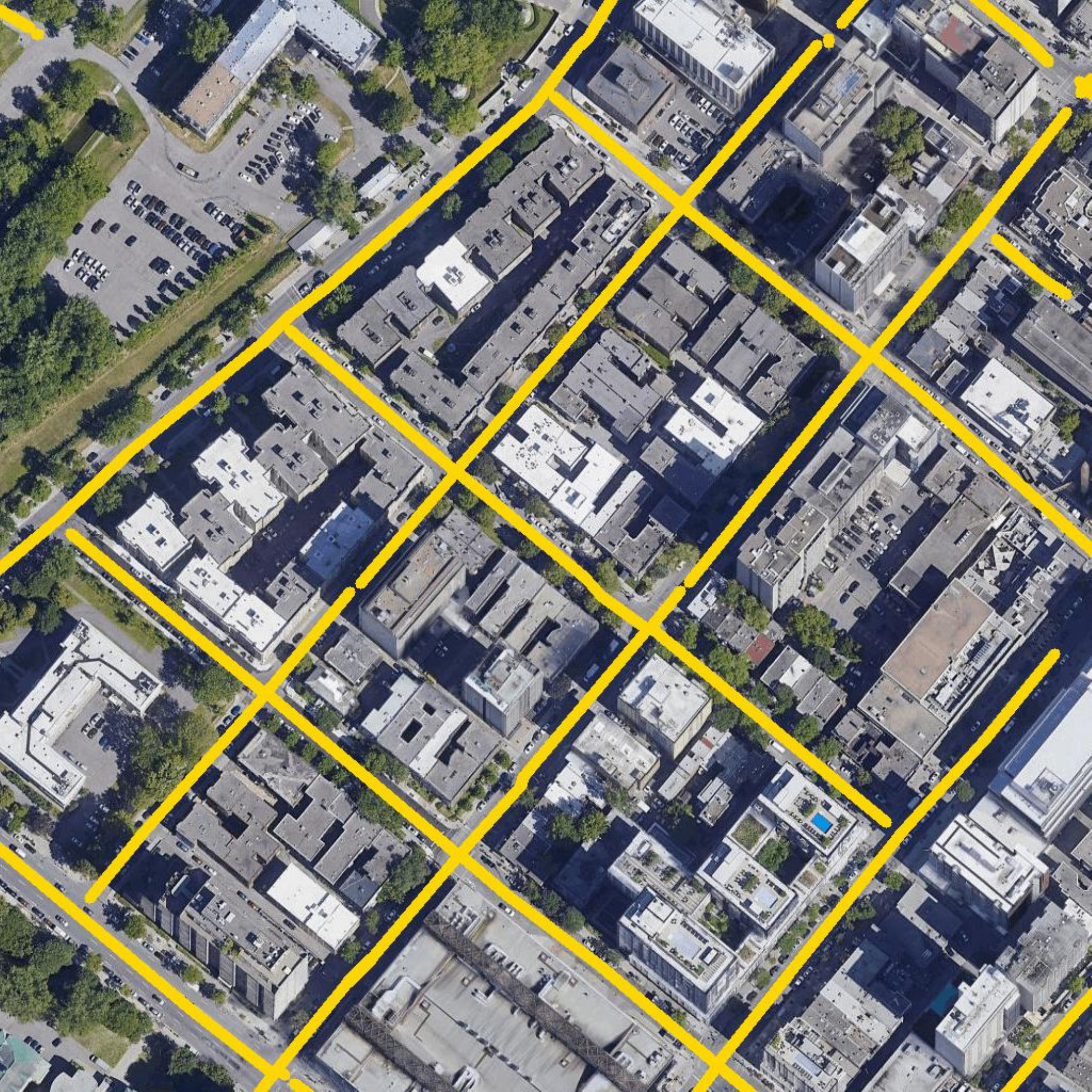}
        \end{subfigure}\vspace{.6ex}
        \begin{subfigure}[t]{\textwidth}
            \includegraphics[width=\textwidth]{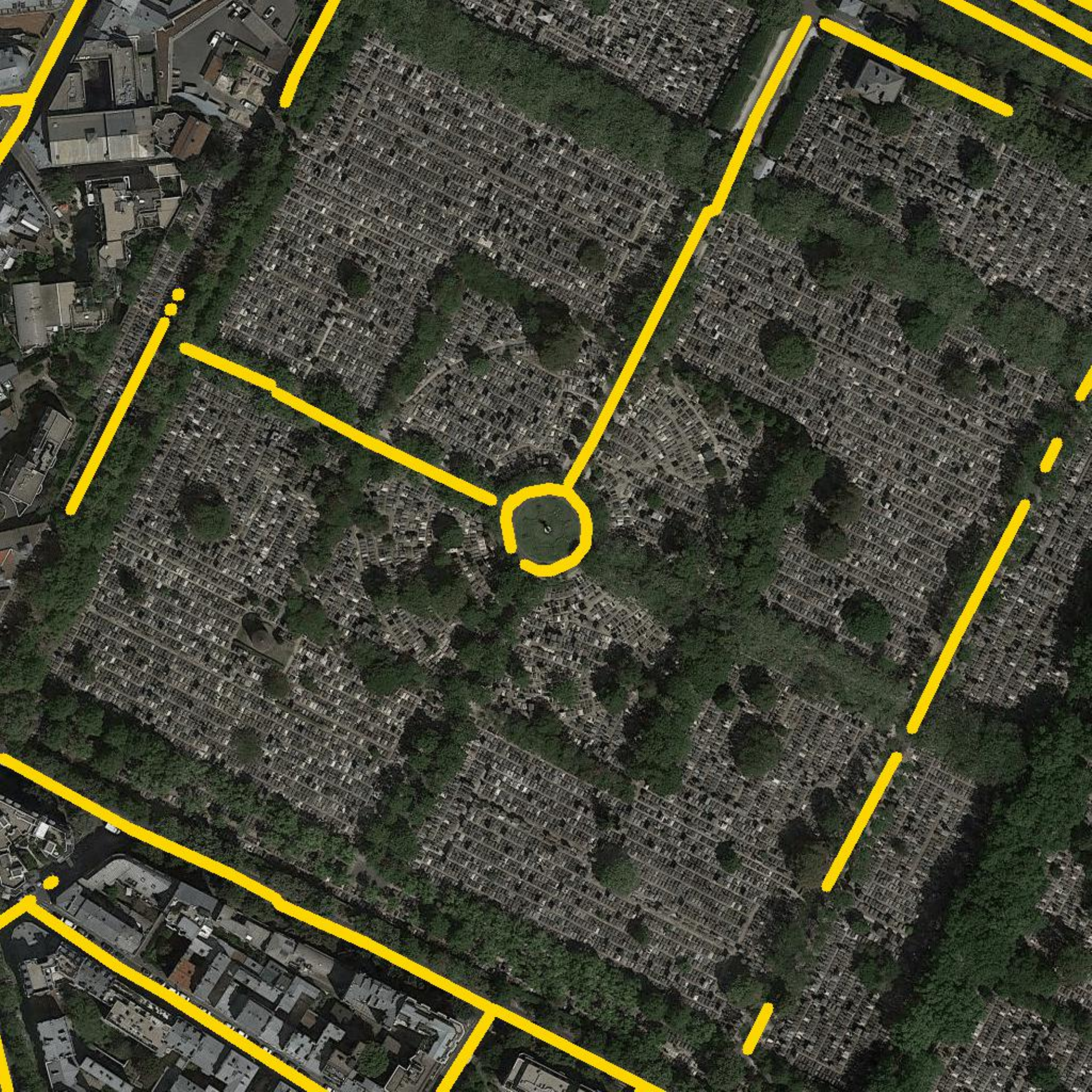}
        \end{subfigure}\vspace{.6ex}
        \begin{subfigure}[t]{\textwidth}
            \includegraphics[width=\textwidth]{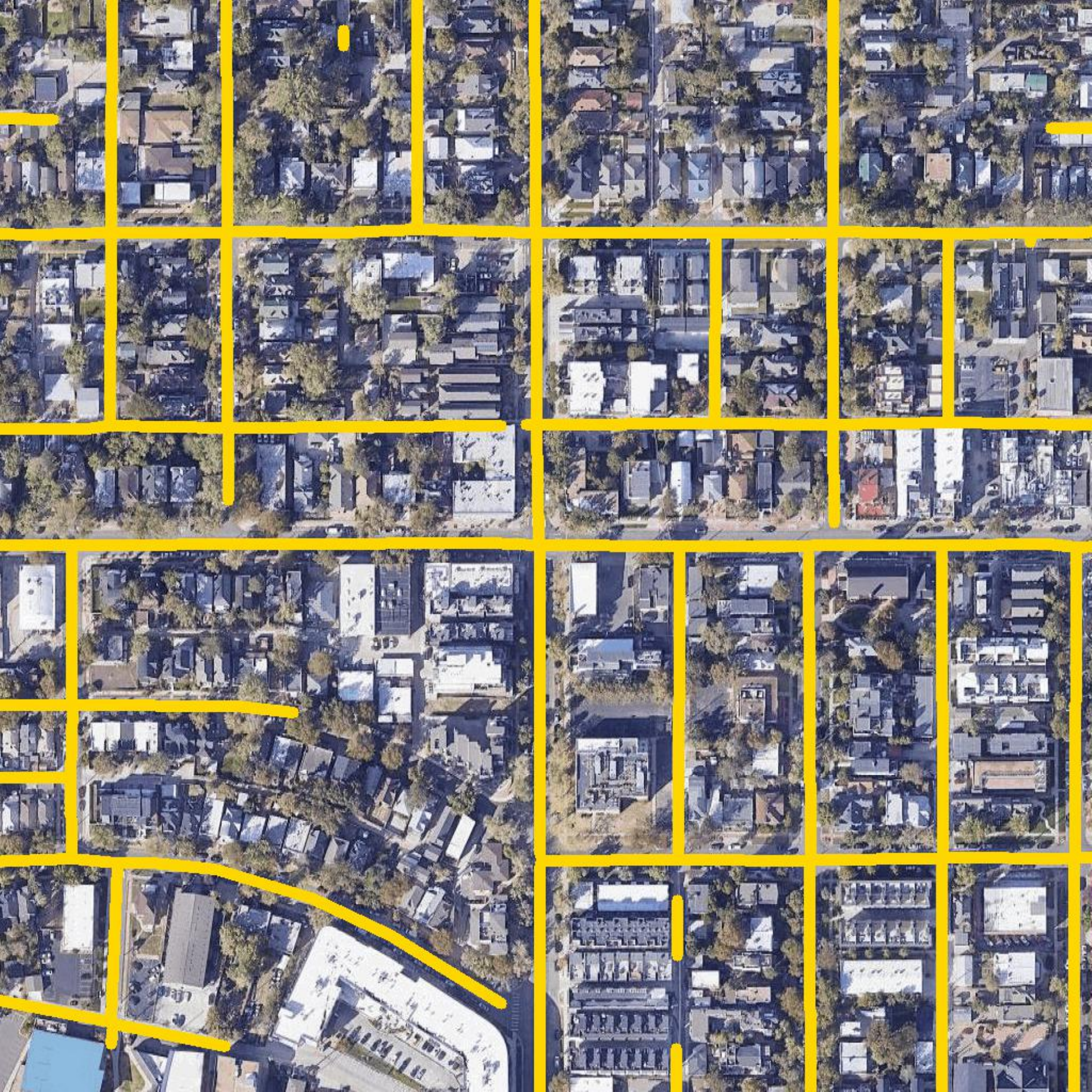}
        \end{subfigure}\vspace{.6ex}
        \begin{subfigure}[t]{\textwidth}
            \includegraphics[width=\textwidth]{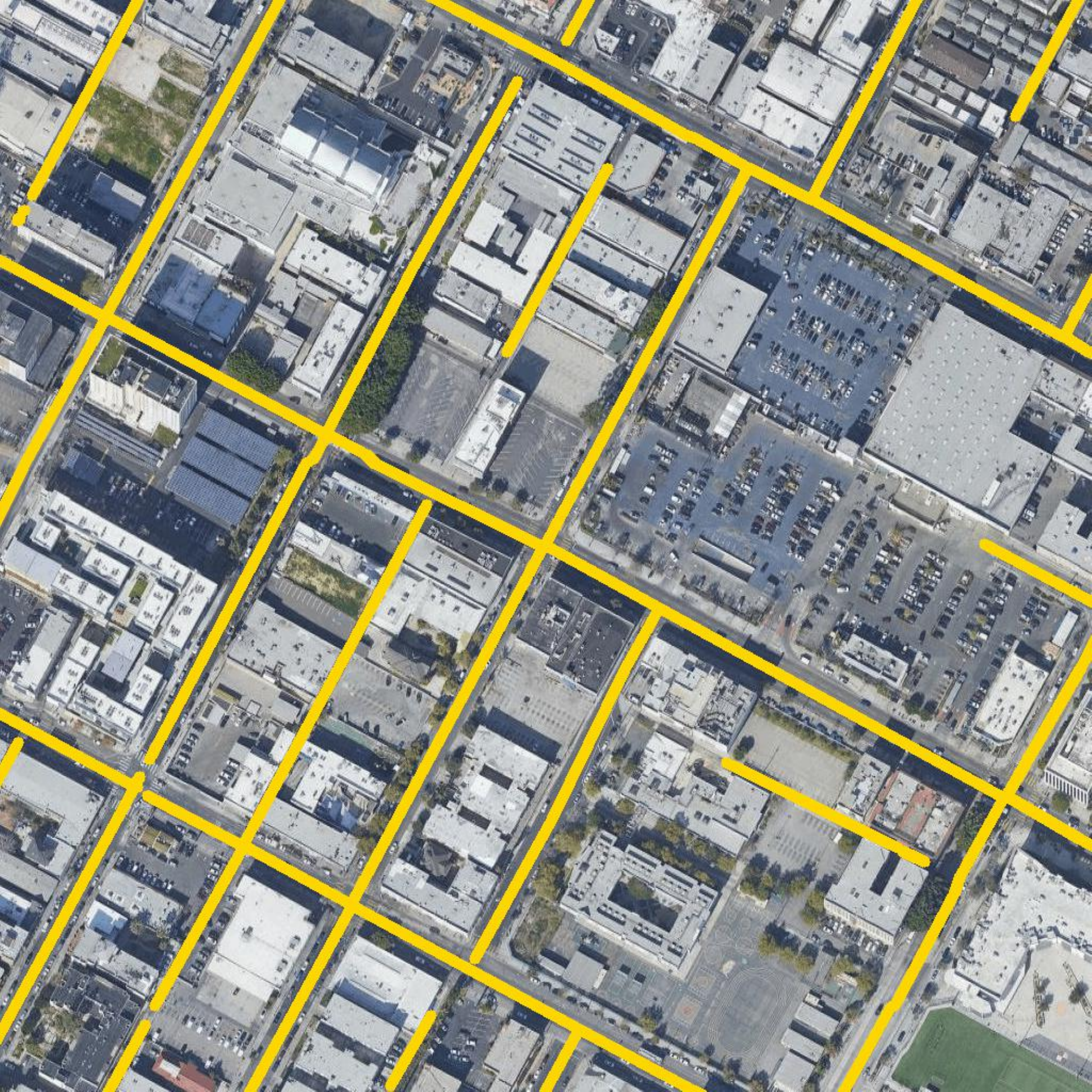}
        \end{subfigure}\vspace{.6ex}
        \begin{subfigure}[t]{\textwidth}
            \includegraphics[width=\textwidth]{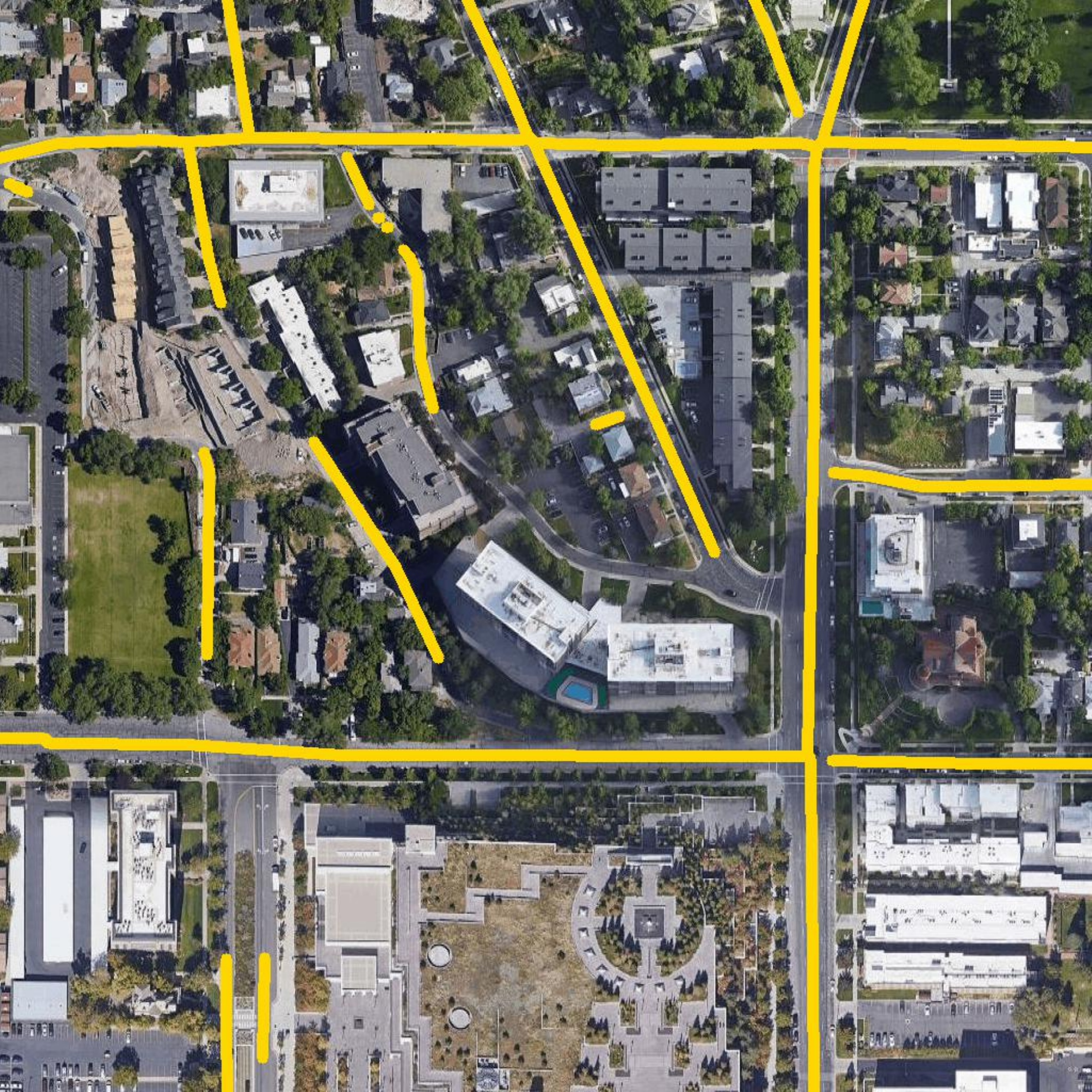}
        \end{subfigure}\vspace{.6ex}
        \caption{SPIN Road\cite{gedara2021spin}}
        \label{fig_qualitative_1st}
    \end{subfigure}
    \begin{subfigure}[t]{0.135\textwidth}
        \begin{subfigure}[t]{\textwidth}
            \includegraphics[width=\textwidth]{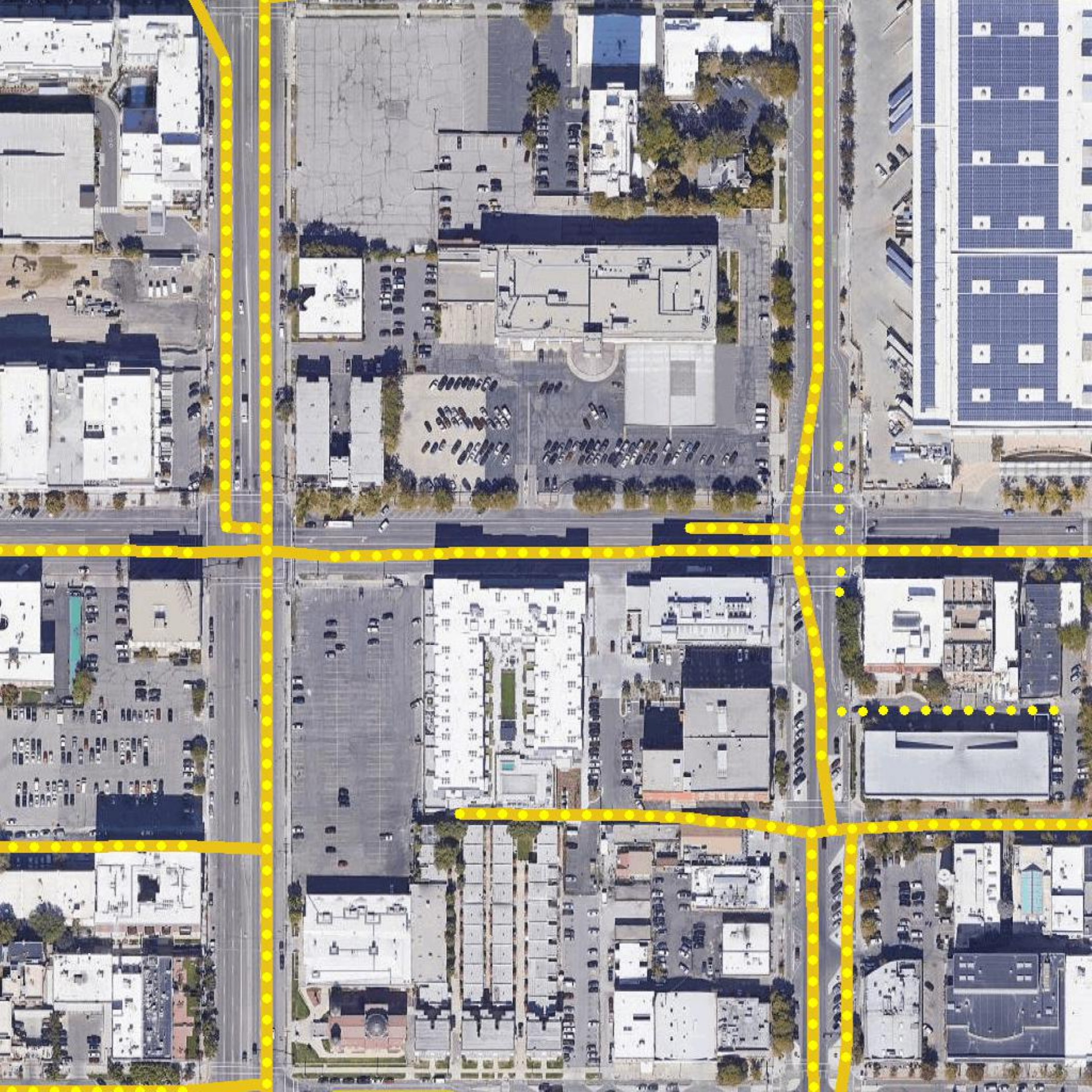}
        \end{subfigure}\vspace{.6ex}
        \begin{subfigure}[t]{\textwidth}
            \includegraphics[width=\textwidth]{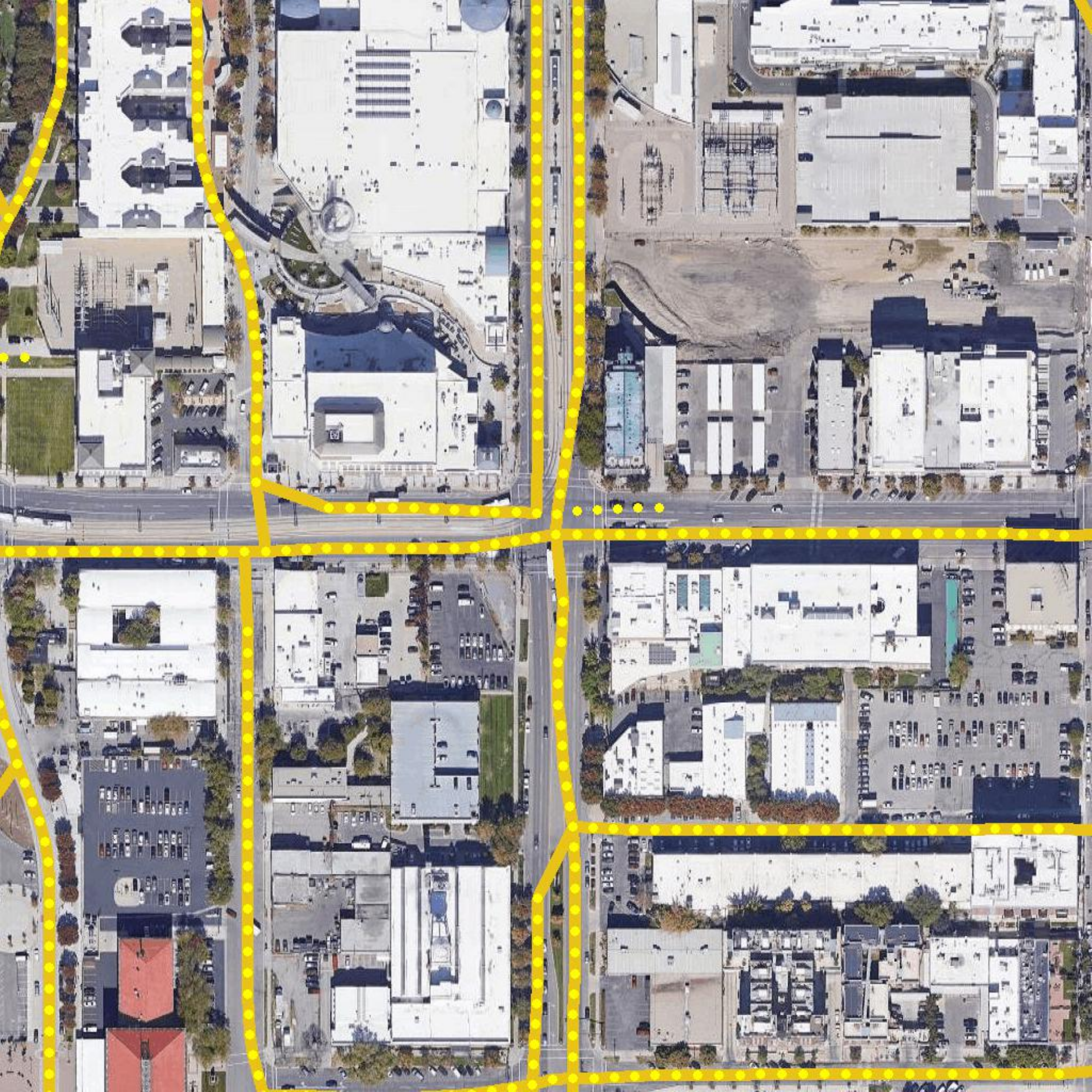}
        \end{subfigure}\vspace{.6ex}
        \begin{subfigure}[t]{\textwidth}
            \includegraphics[width=\textwidth]{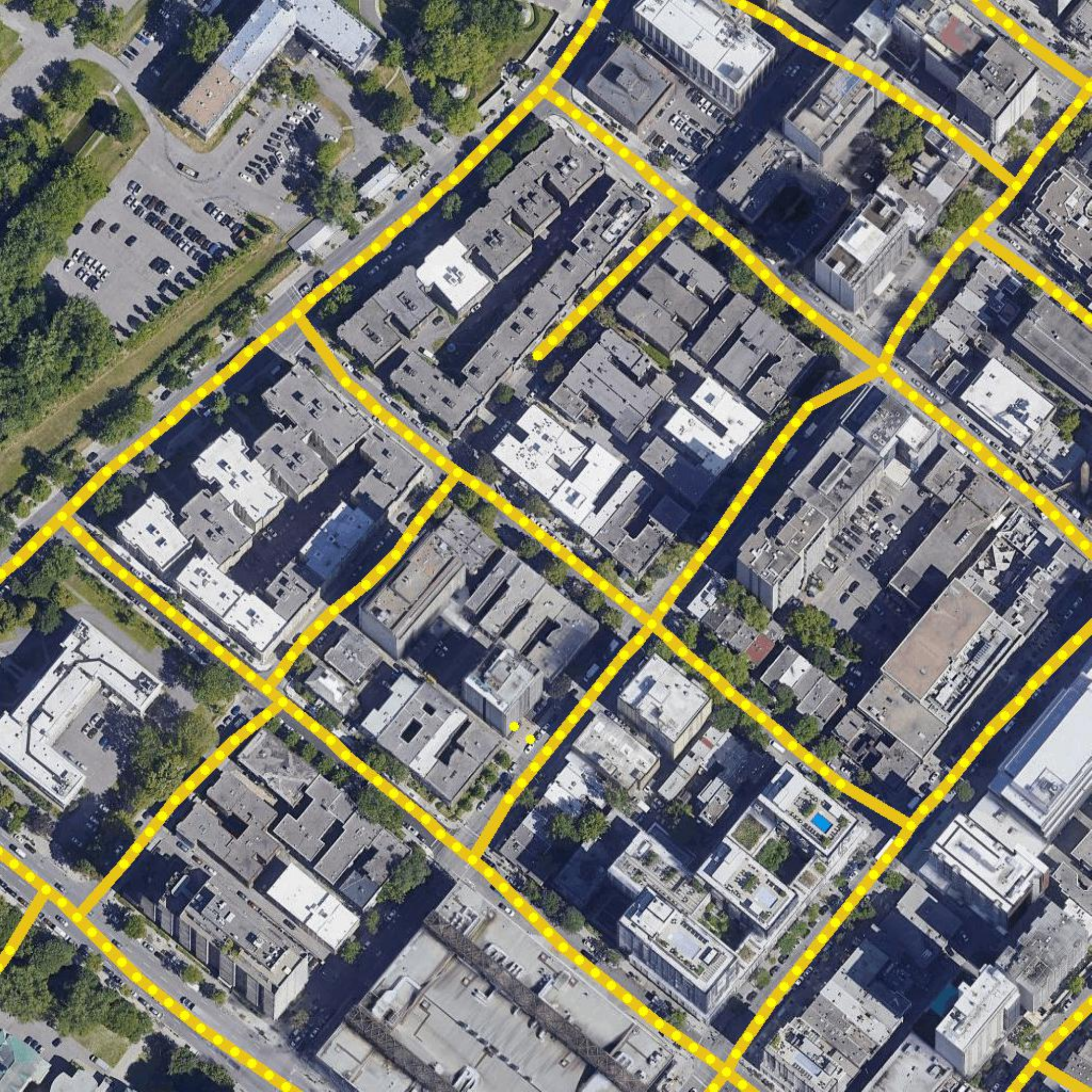}
        \end{subfigure}\vspace{.6ex}
        \begin{subfigure}[t]{\textwidth}
            \includegraphics[width=\textwidth]{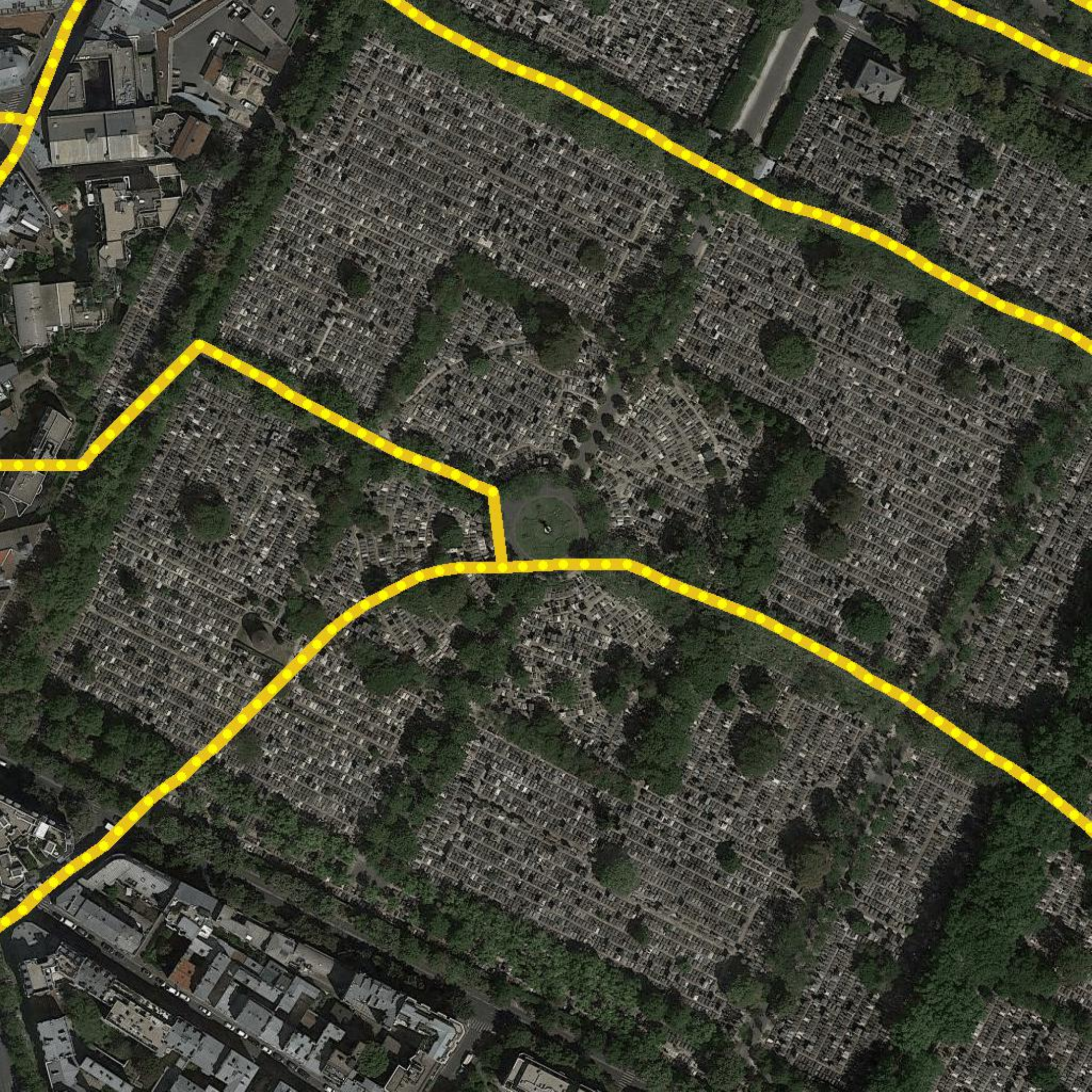}
        \end{subfigure}\vspace{.6ex}
        \begin{subfigure}[t]{\textwidth}
            \includegraphics[width=\textwidth]{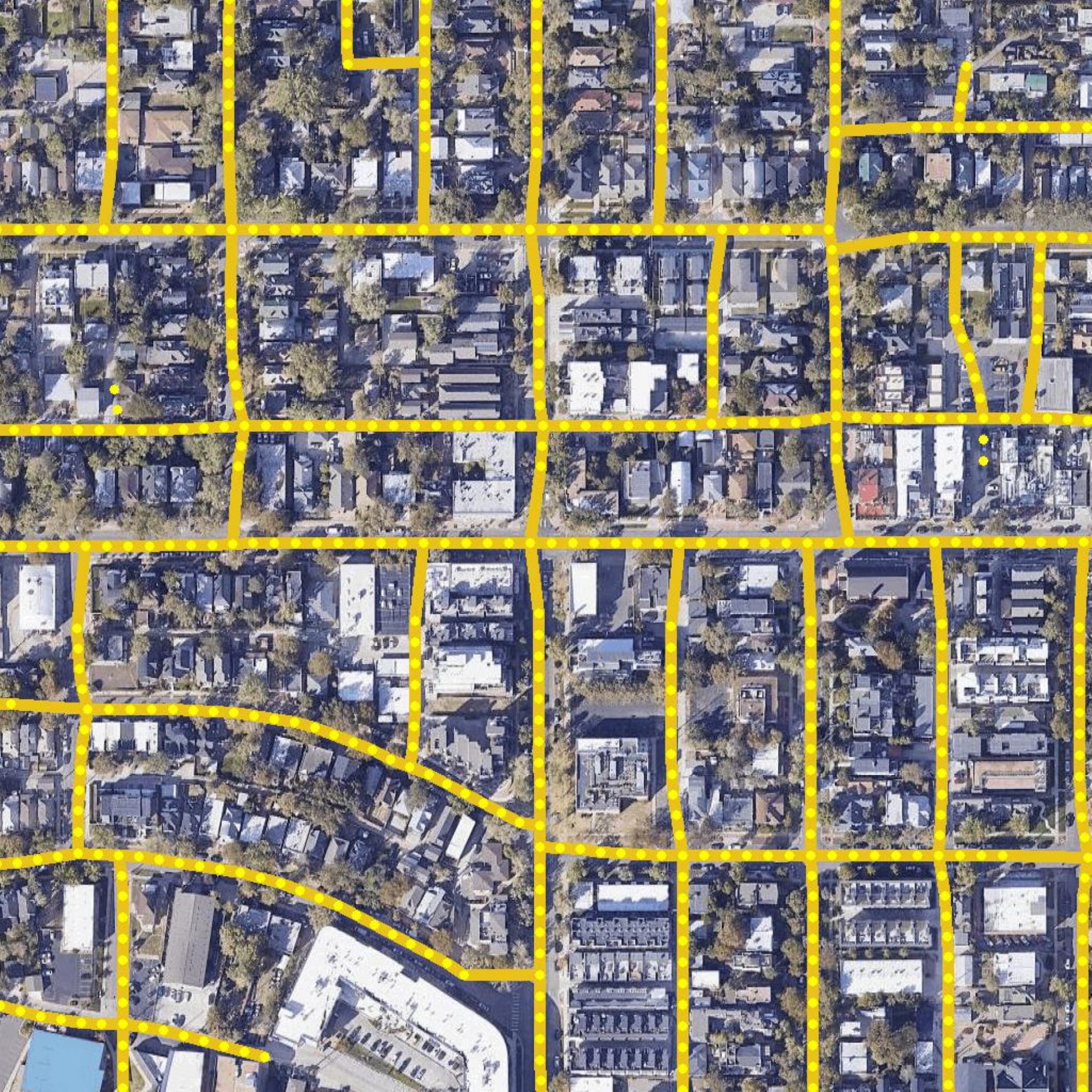}
        \end{subfigure}\vspace{.6ex}
        \begin{subfigure}[t]{\textwidth}
            \includegraphics[width=\textwidth]{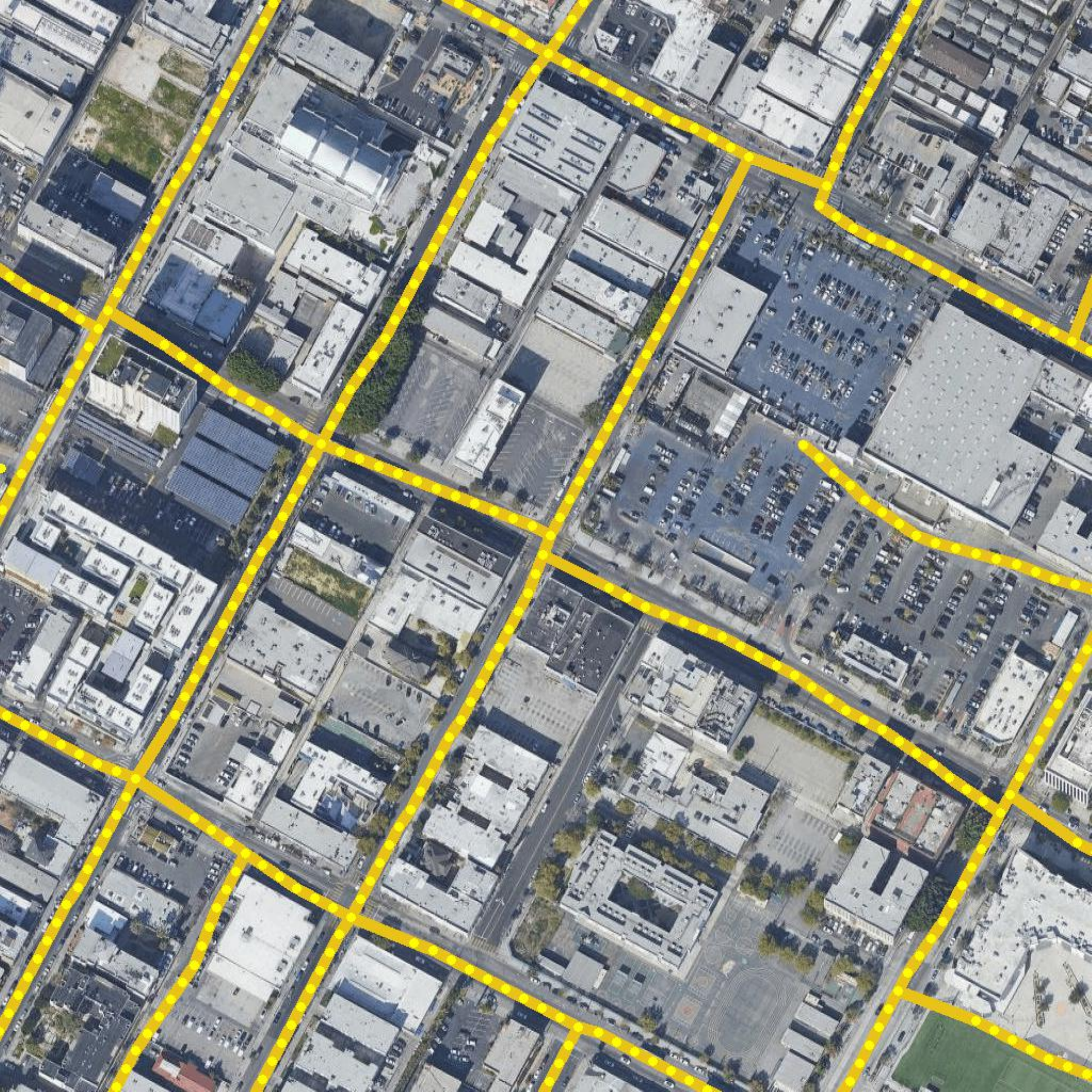}
        \end{subfigure}\vspace{.6ex}
        \begin{subfigure}[t]{\textwidth}
            \includegraphics[width=\textwidth]{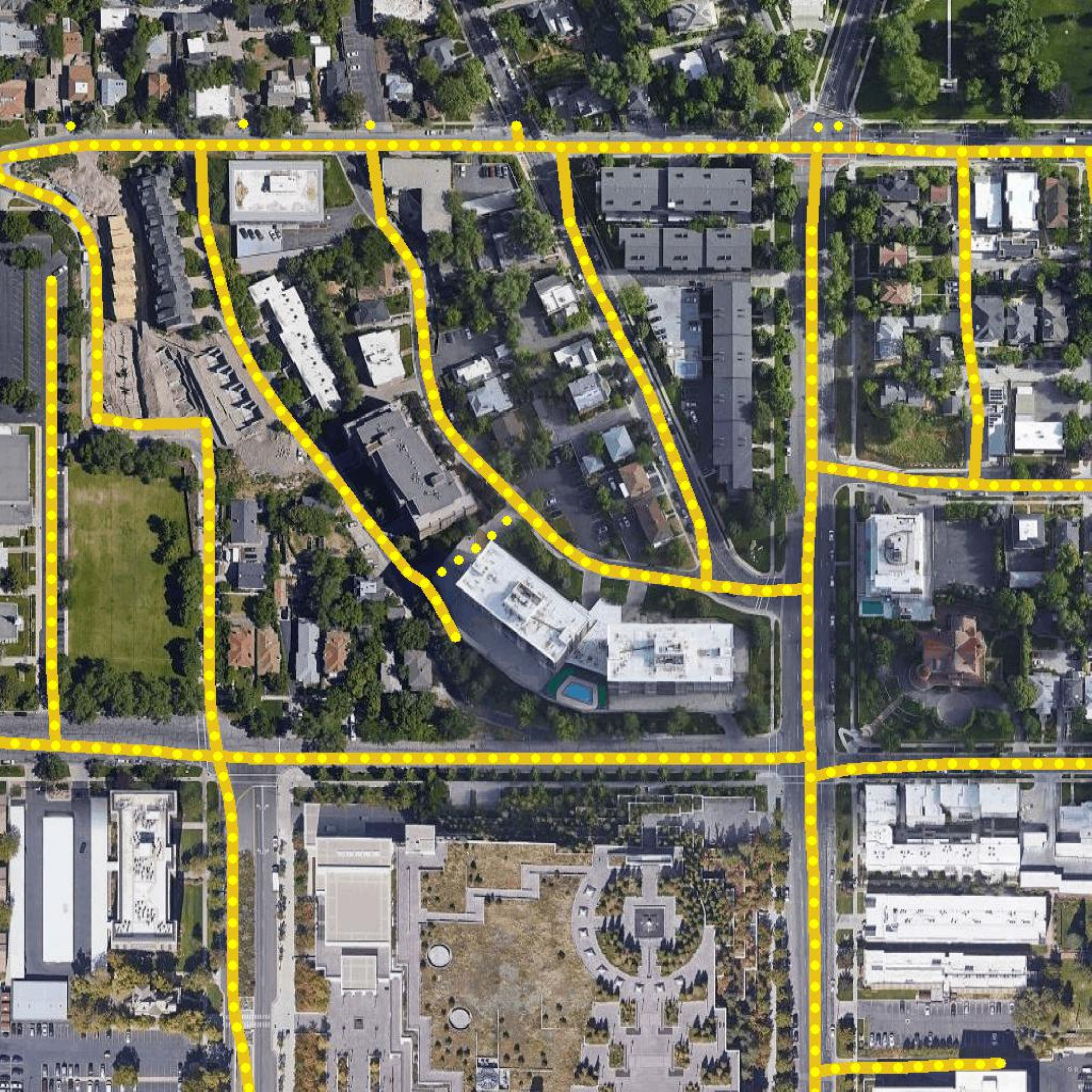}
        \end{subfigure}\vspace{.6ex}
        \caption{\footnotesize{RoadTracer} \cite{bastani2018roadtracer}}
        \label{fig_qualitative_1st}
    \end{subfigure}
    \begin{subfigure}[t]{0.135\textwidth}
        \begin{subfigure}[t]{\textwidth}
            \includegraphics[width=\textwidth]{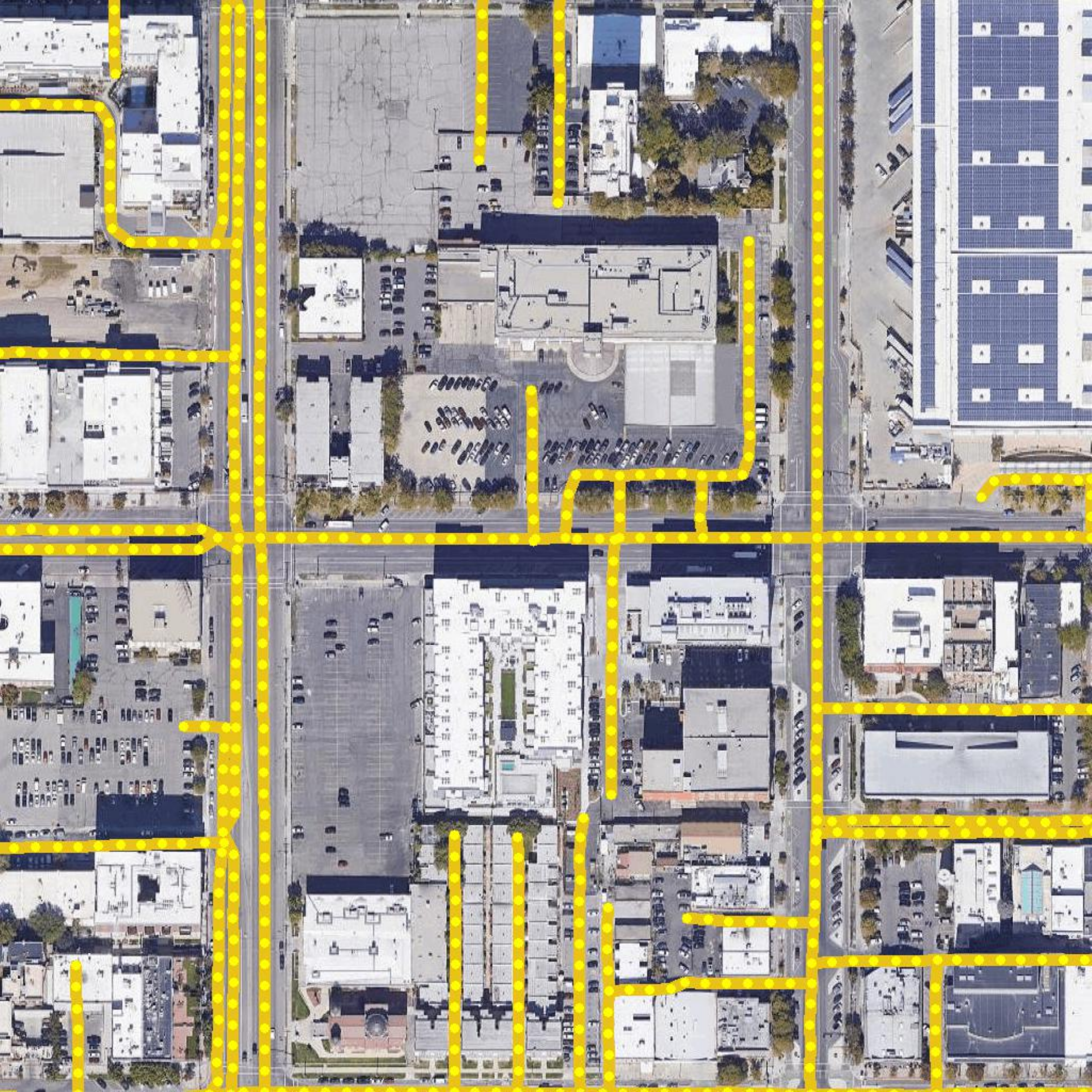}
        \end{subfigure}\vspace{.6ex}
        \begin{subfigure}[t]{\textwidth}
            \includegraphics[width=\textwidth]{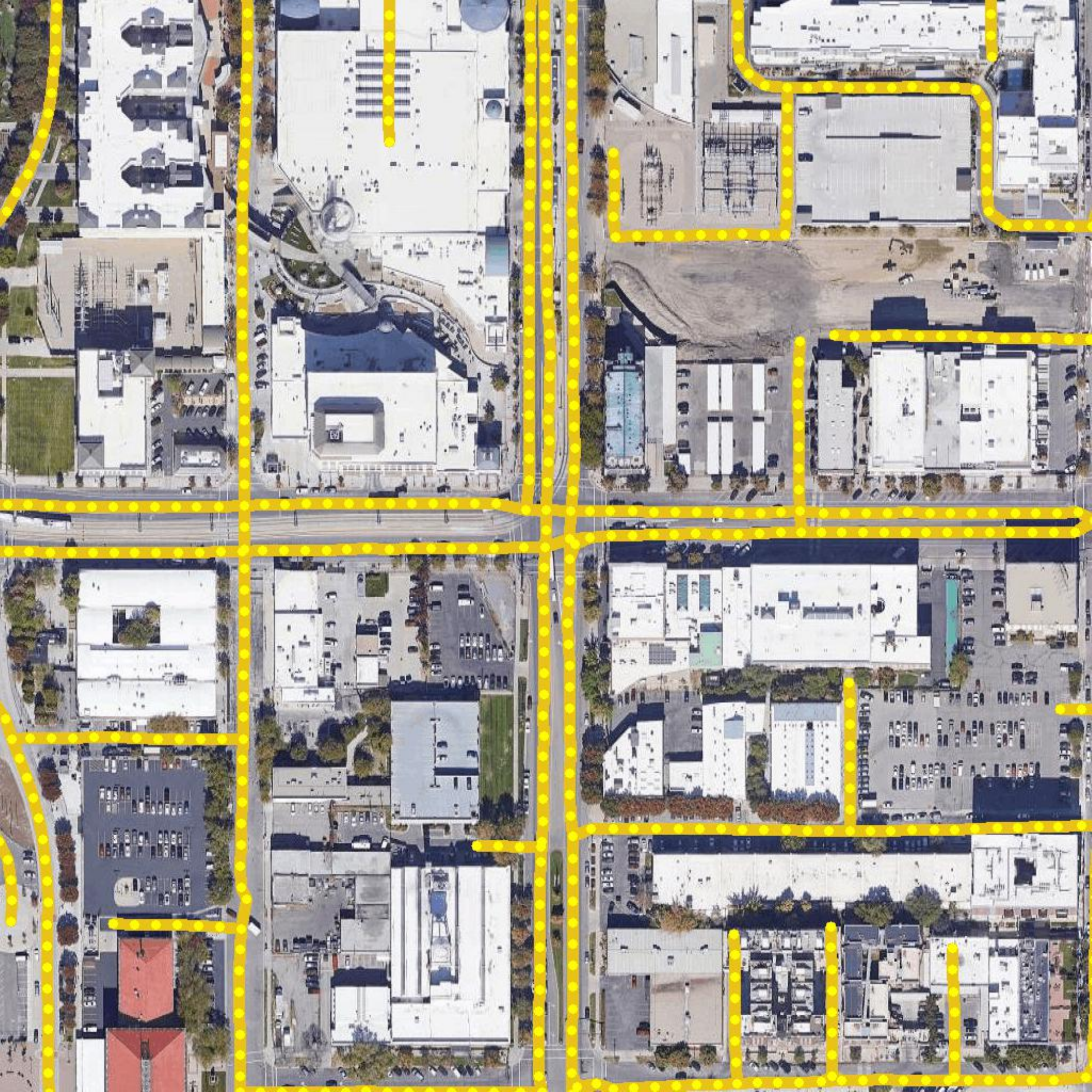}
        \end{subfigure}\vspace{.6ex}
        \begin{subfigure}[t]{\textwidth}
            \includegraphics[width=\textwidth]{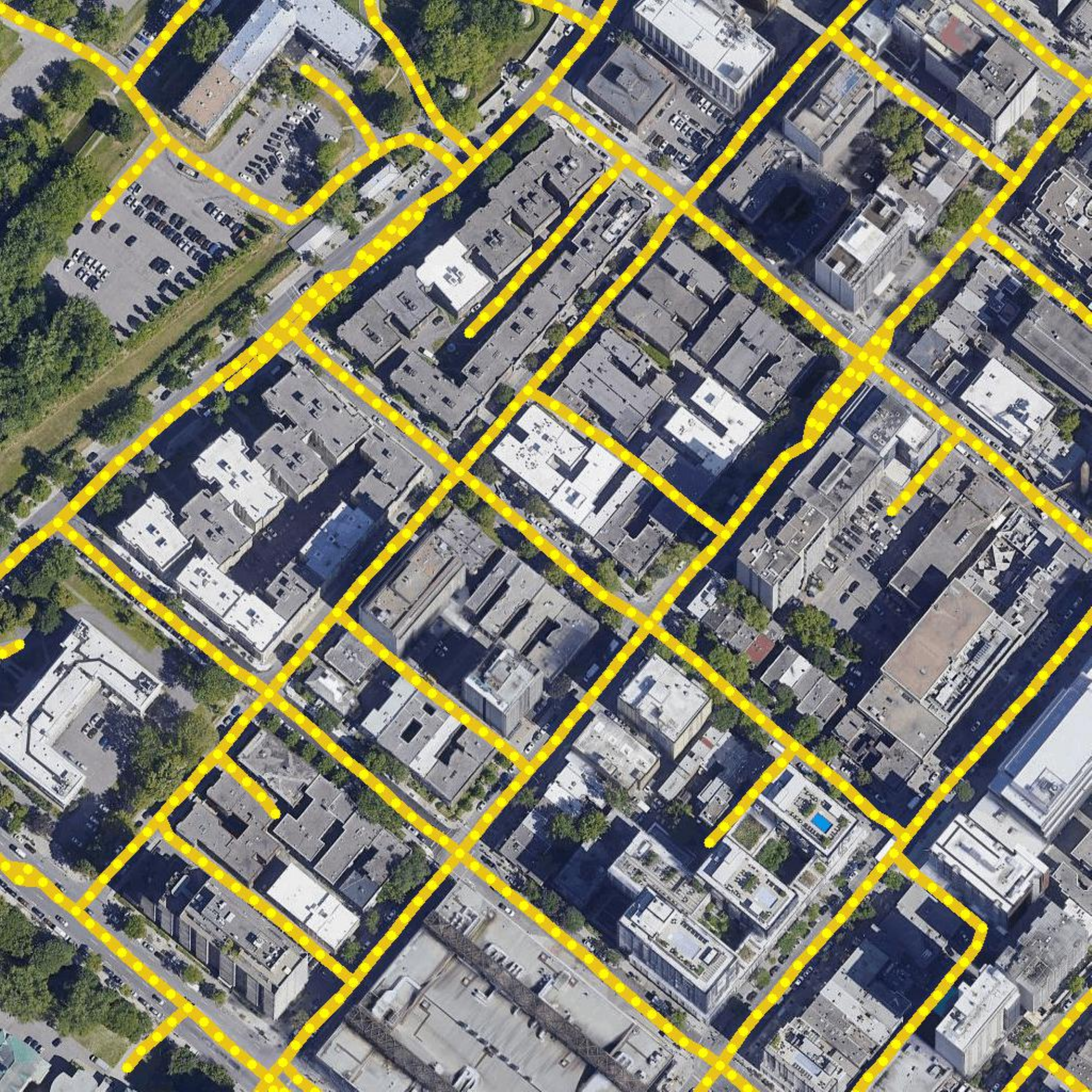}
        \end{subfigure}\vspace{.6ex}
        \begin{subfigure}[t]{\textwidth}
            \includegraphics[width=\textwidth]{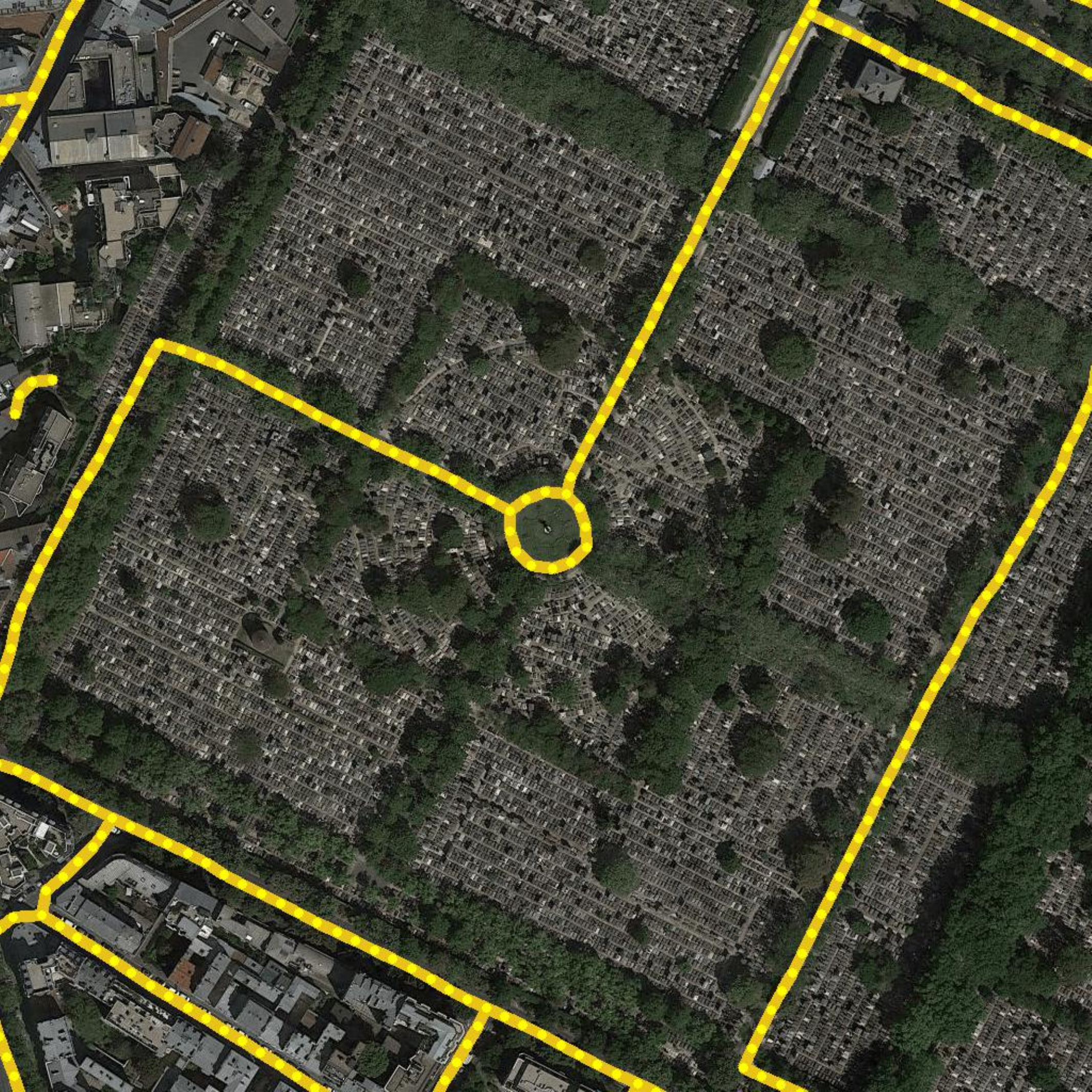}
        \end{subfigure}\vspace{.6ex}
        \begin{subfigure}[t]{\textwidth}
            \includegraphics[width=\textwidth]{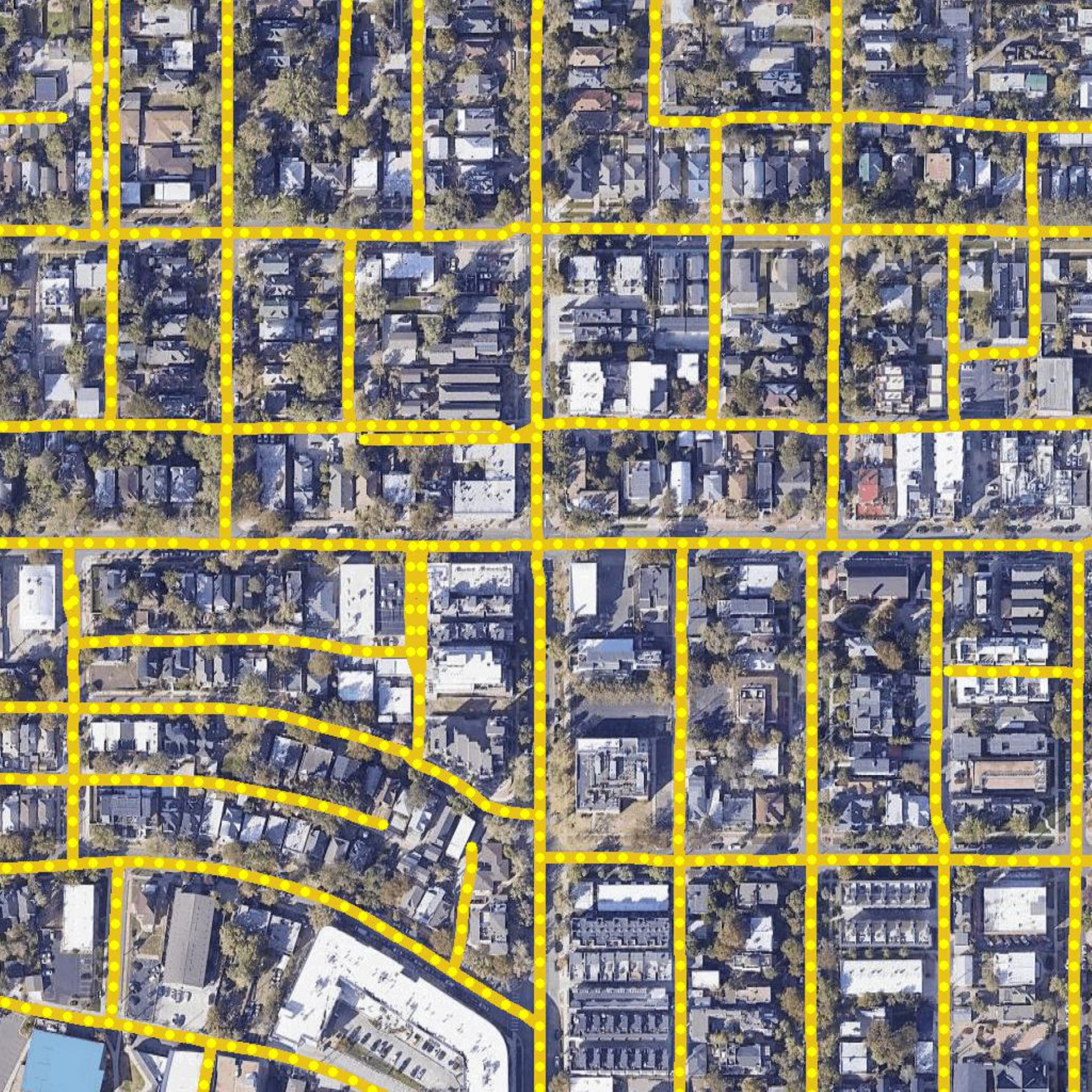}
        \end{subfigure}\vspace{.6ex}
        \begin{subfigure}[t]{\textwidth}
            \includegraphics[width=\textwidth]{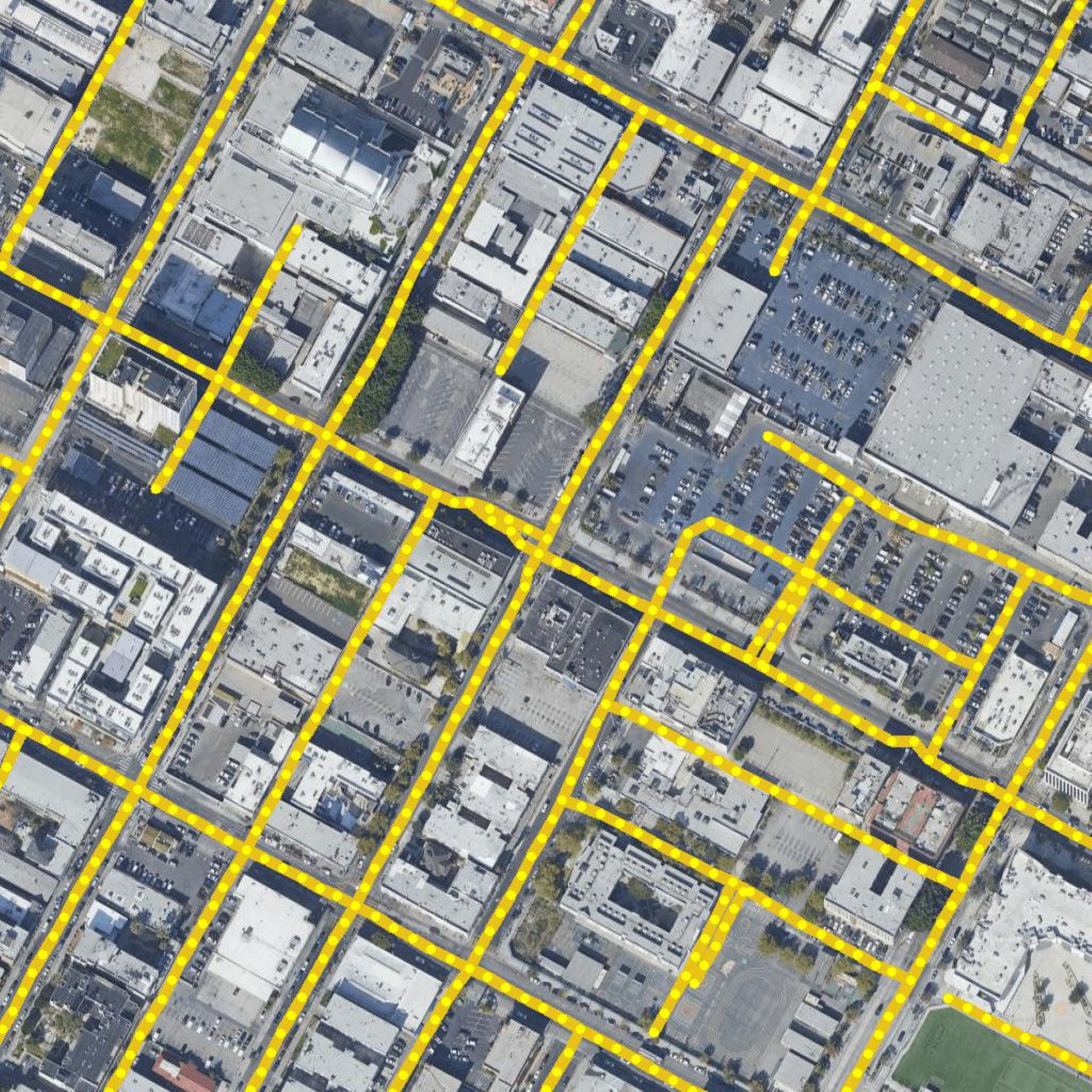}
        \end{subfigure}\vspace{.6ex}
        \begin{subfigure}[t]{\textwidth}
            \includegraphics[width=\textwidth]{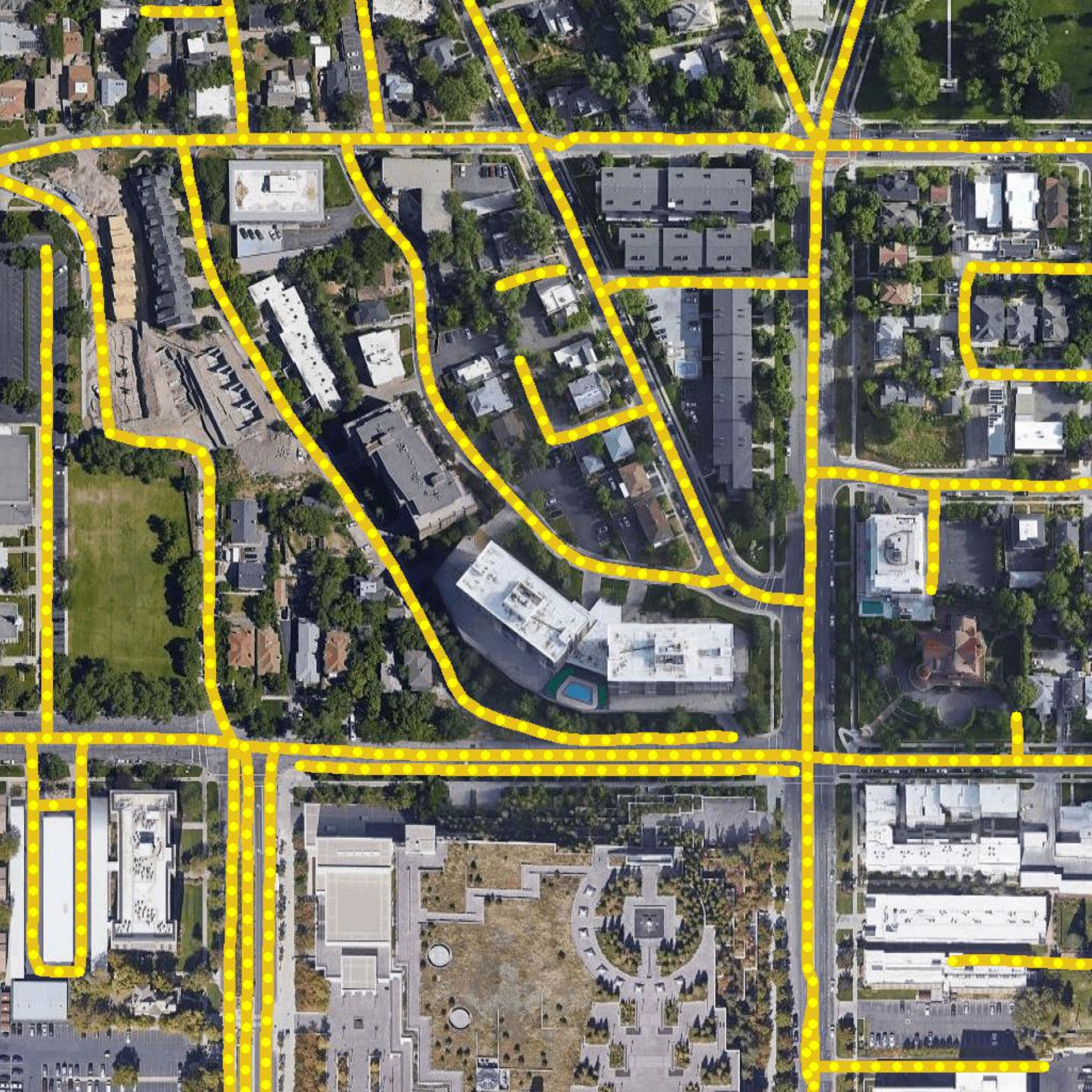}
        \end{subfigure}\vspace{.6ex}
        \caption{VecRoad \cite{tan2020vecroad}}
        \label{fig_qualitative_1st}
    \end{subfigure}
    \begin{subfigure}[t]{0.135\textwidth}
        \begin{subfigure}[t]{\textwidth}
            \includegraphics[width=\textwidth]{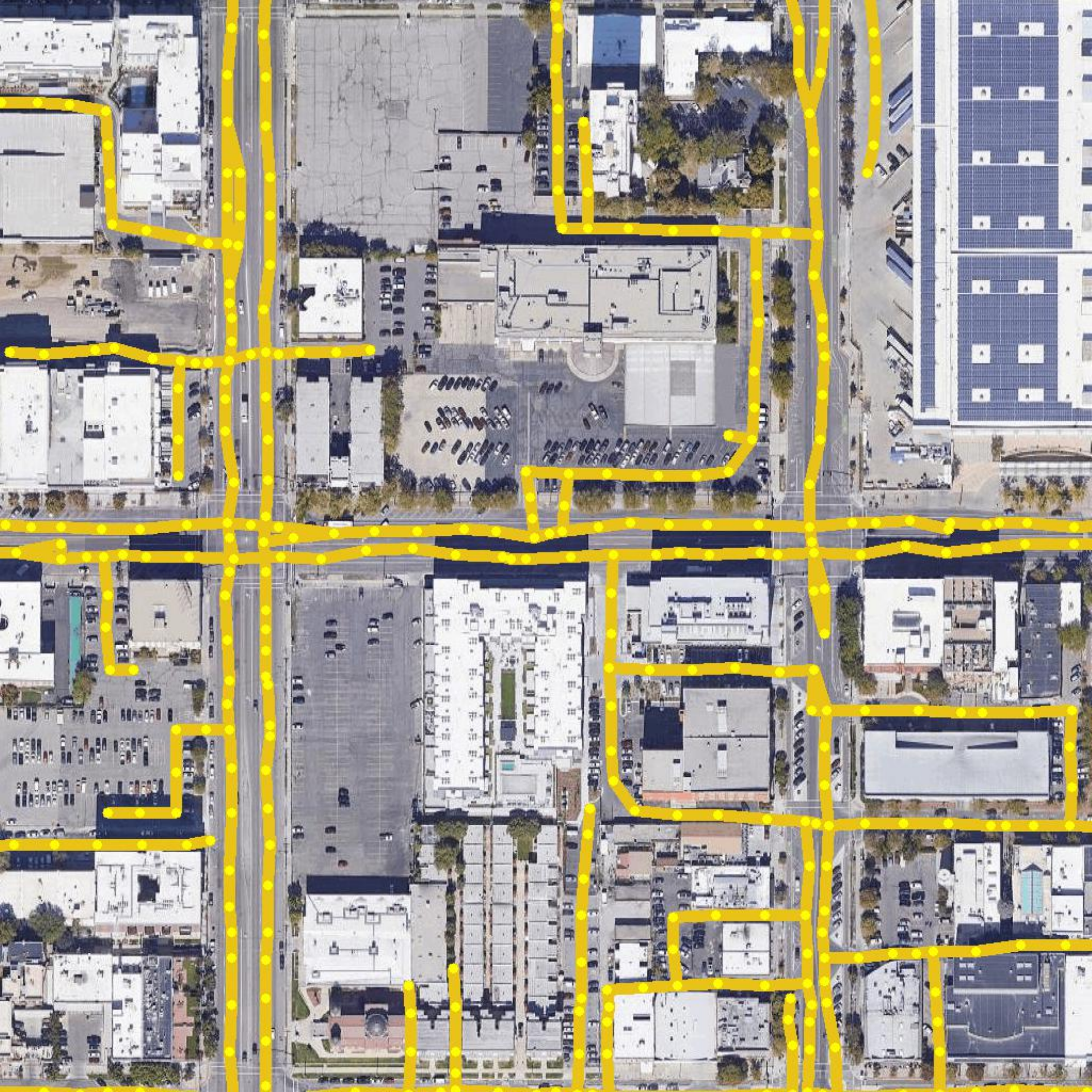}
        \end{subfigure}\vspace{.6ex}
        \begin{subfigure}[t]{\textwidth}
            \includegraphics[width=\textwidth]{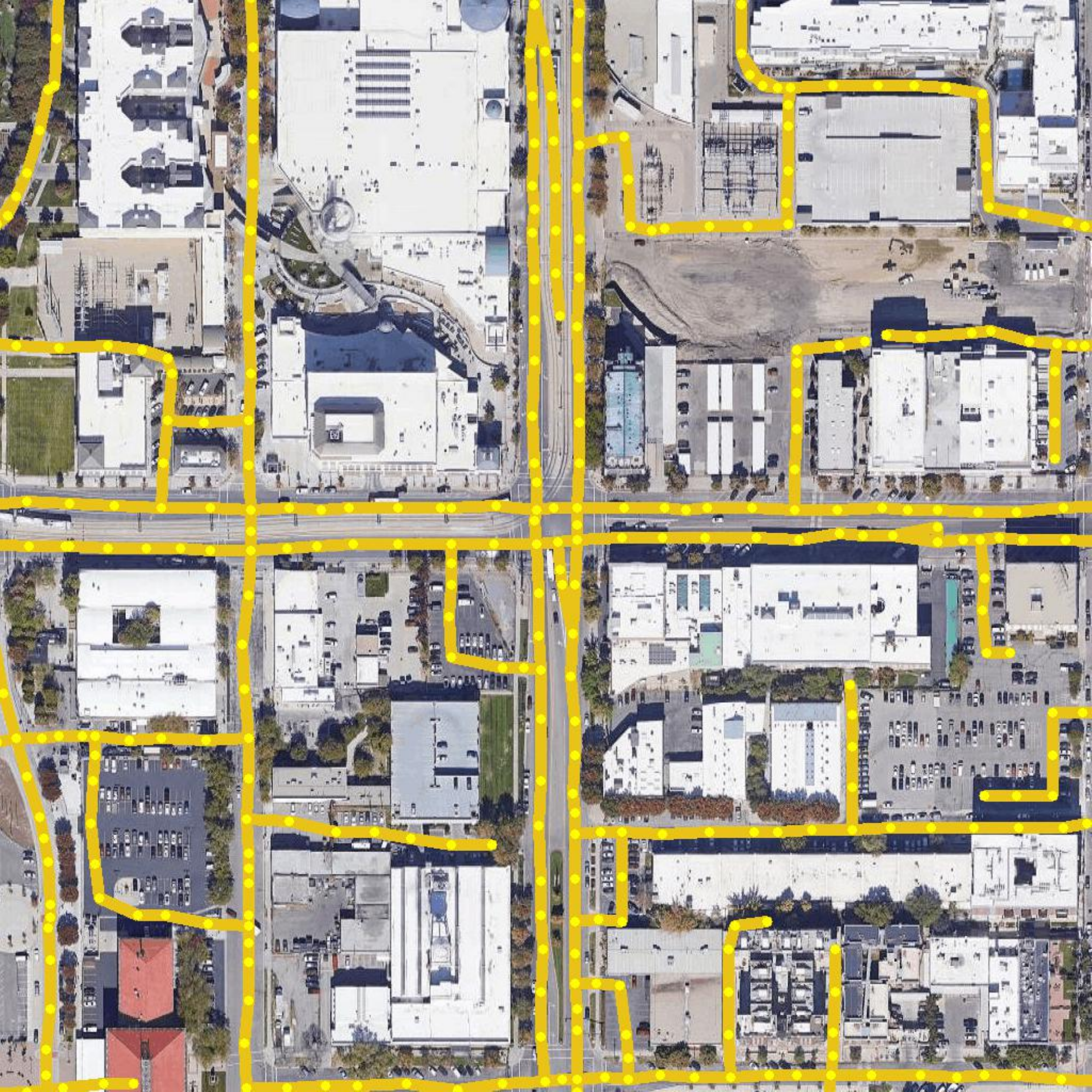}
        \end{subfigure}\vspace{.6ex}
        \begin{subfigure}[t]{\textwidth}
            \includegraphics[width=\textwidth]{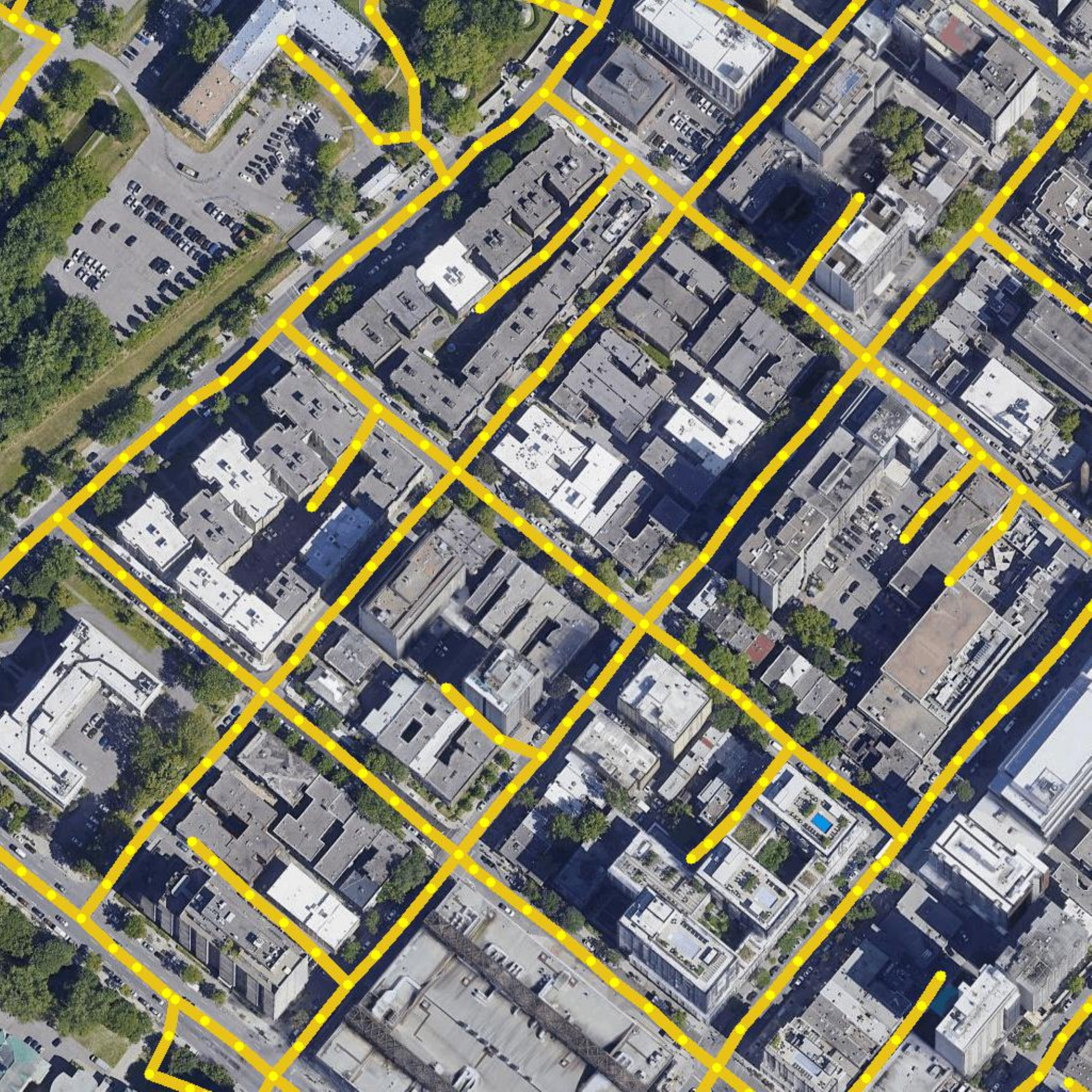}
        \end{subfigure}\vspace{.6ex}
        \begin{subfigure}[t]{\textwidth}
            \includegraphics[width=\textwidth]{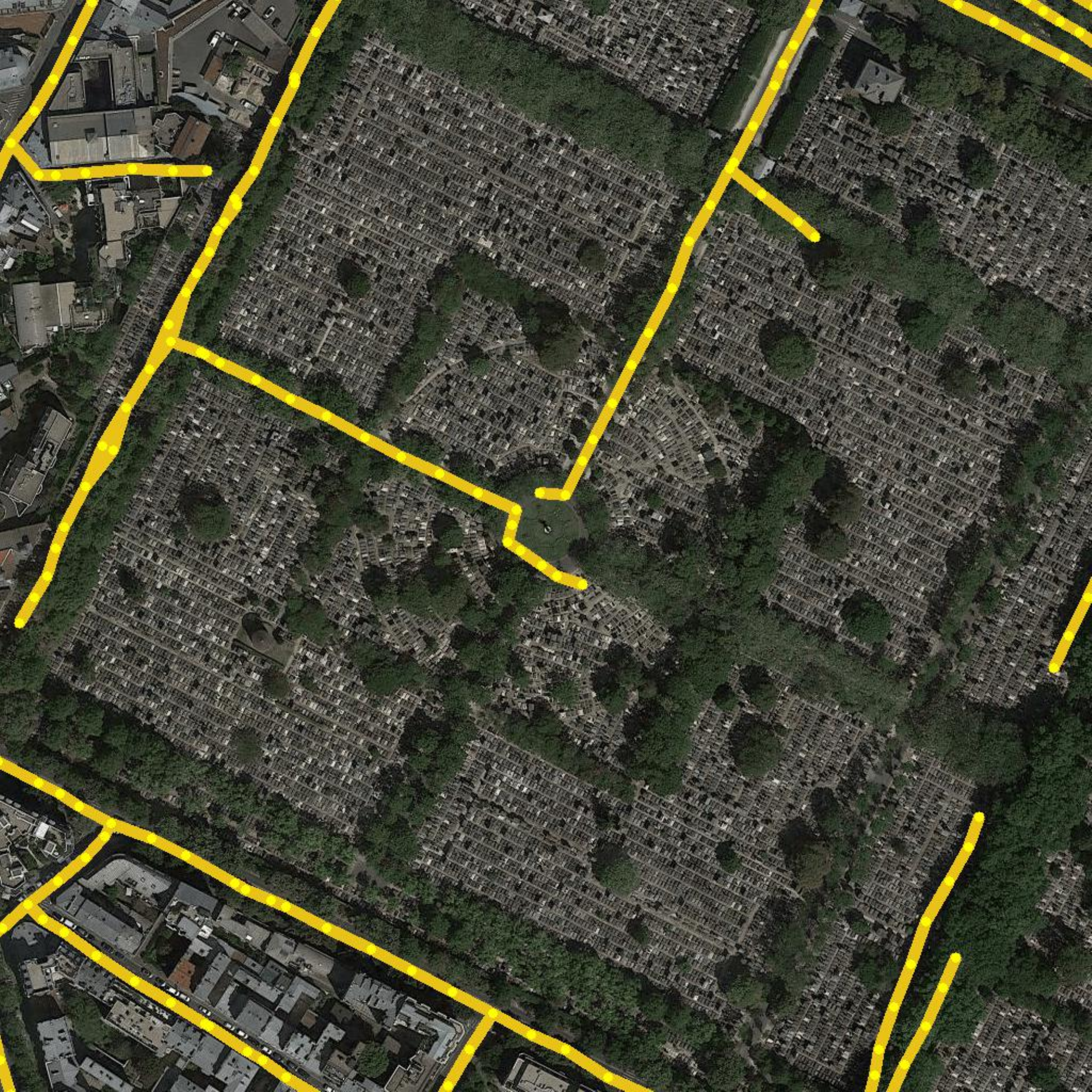}
        \end{subfigure}\vspace{.6ex}
        \begin{subfigure}[t]{\textwidth}
            \includegraphics[width=\textwidth]{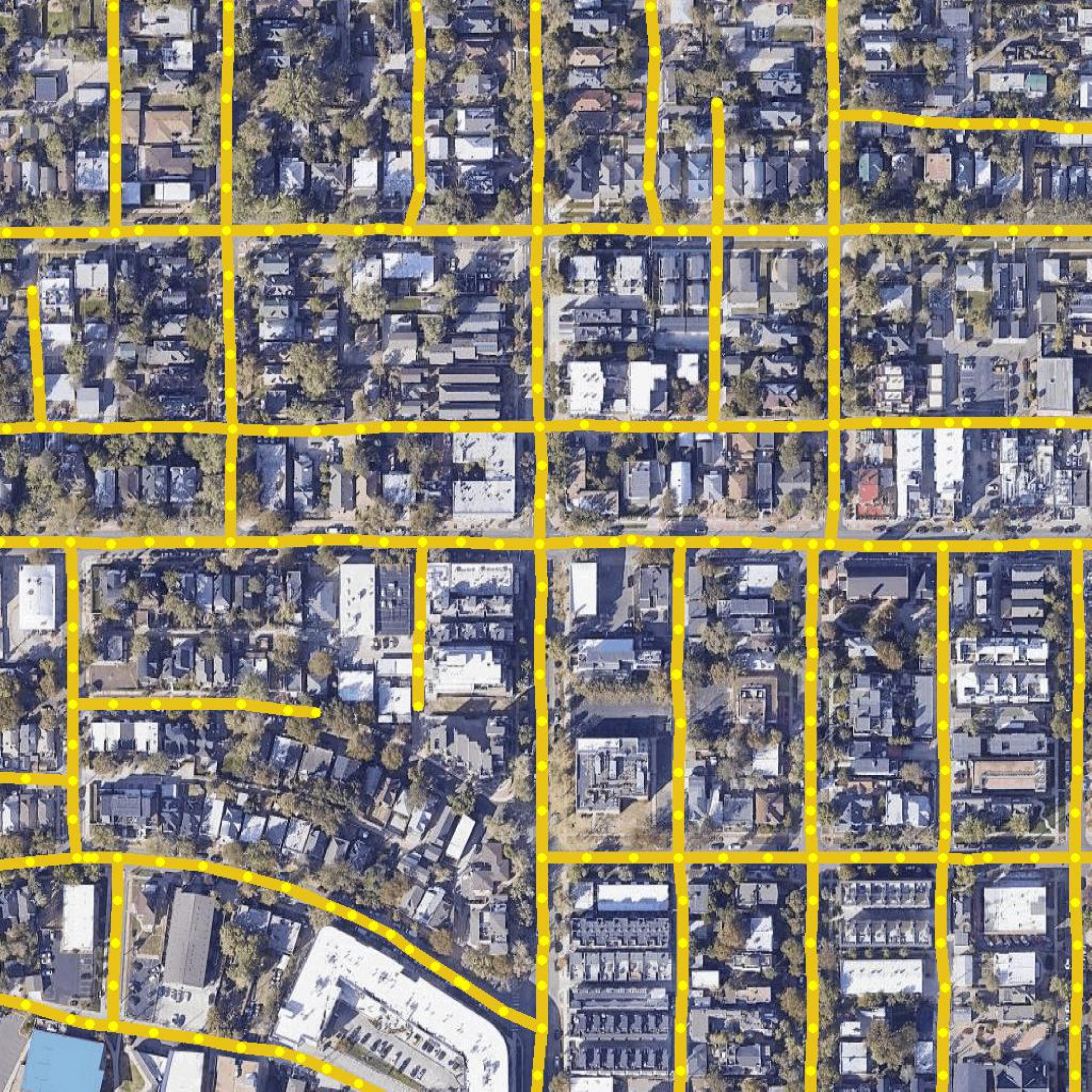}
        \end{subfigure}\vspace{.6ex}
        \begin{subfigure}[t]{\textwidth}
            \includegraphics[width=\textwidth]{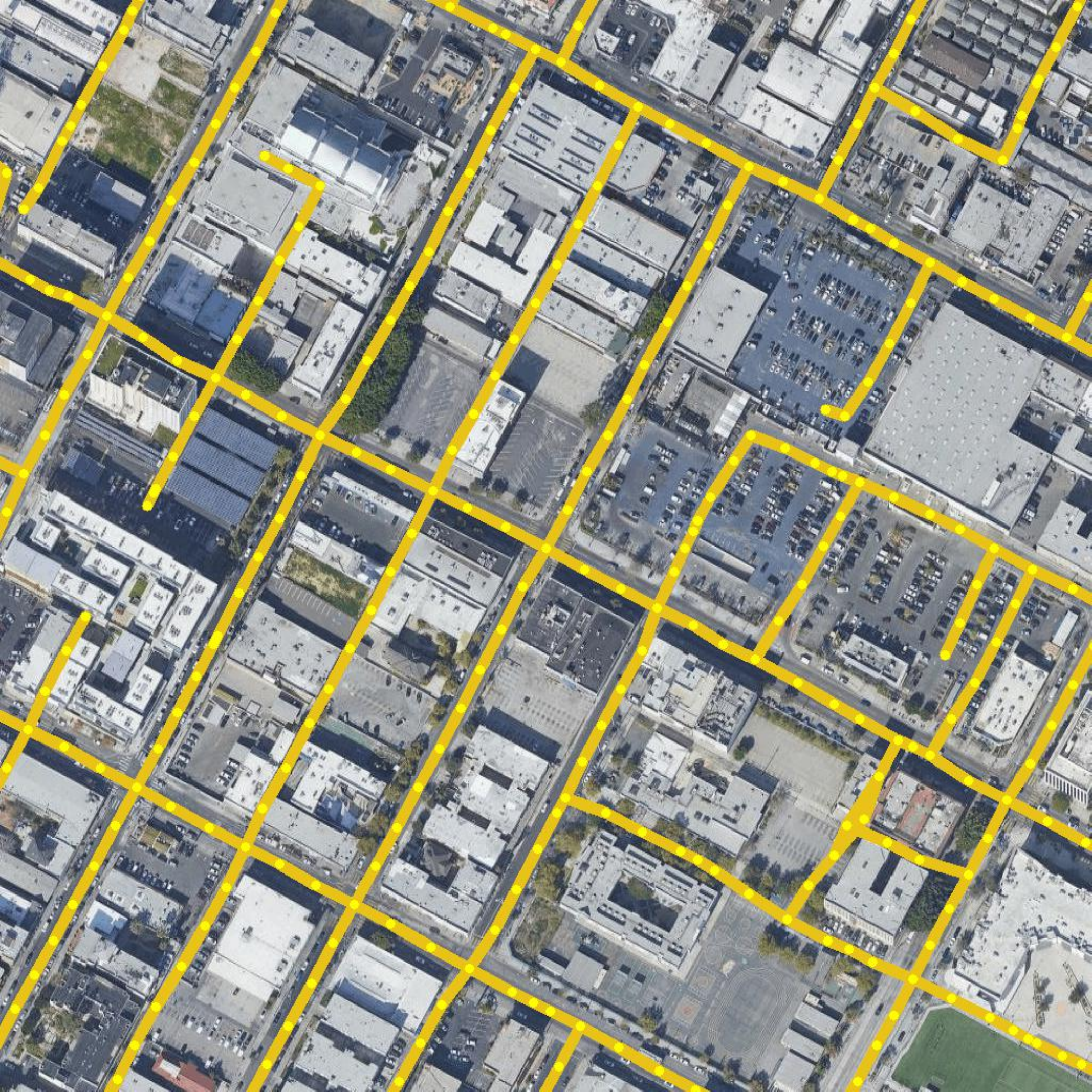}
        \end{subfigure}\vspace{.6ex}
        \begin{subfigure}[t]{\textwidth}
            \includegraphics[width=\textwidth]{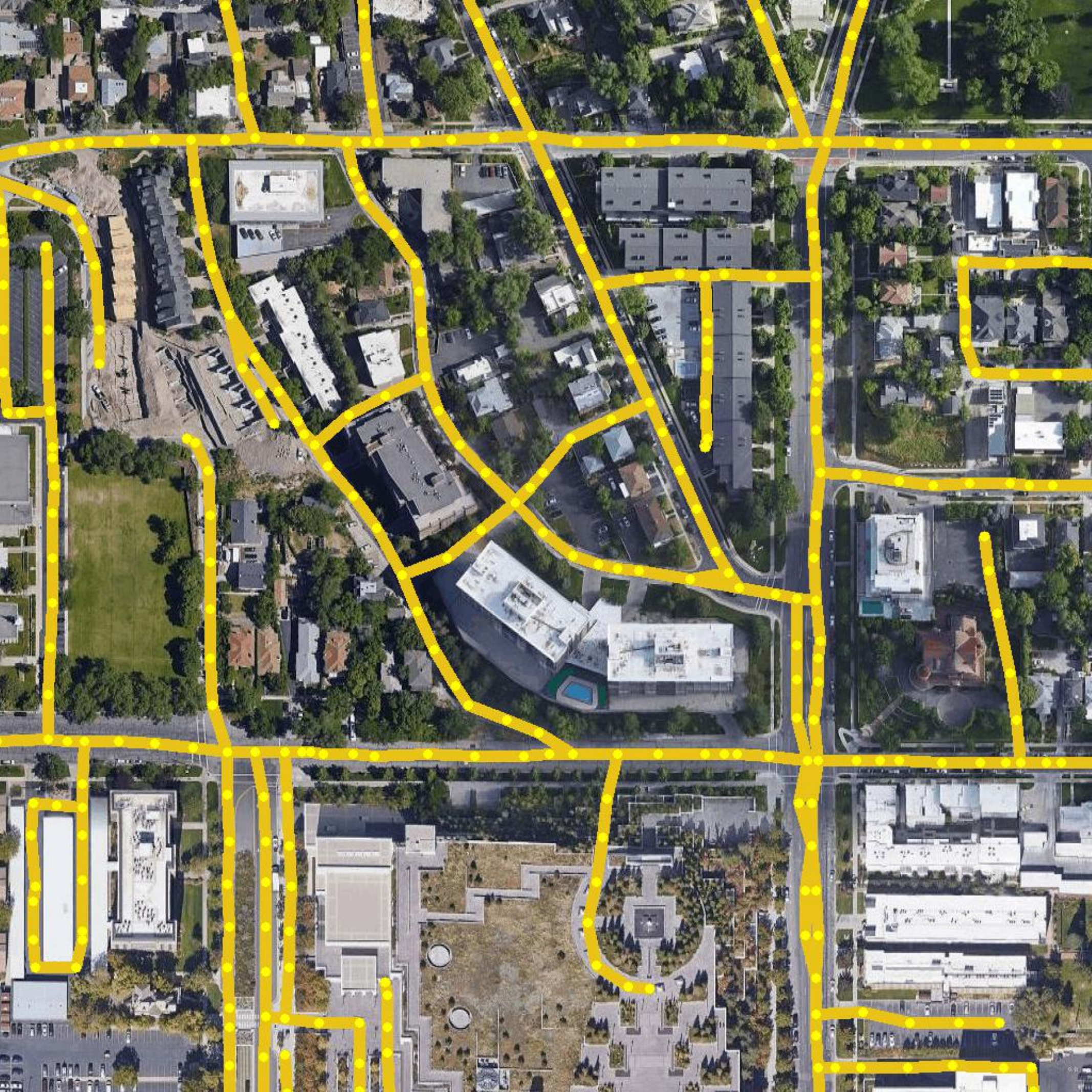}
        \end{subfigure}\vspace{.6ex}
        \caption{Sat2Graph \cite{he2020sat2graph}}
        \label{fig_qualitative_1st}
    \end{subfigure}
    \begin{subfigure}[t]{0.135\textwidth}
        \begin{subfigure}[t]{\textwidth}
            \includegraphics[width=\textwidth]{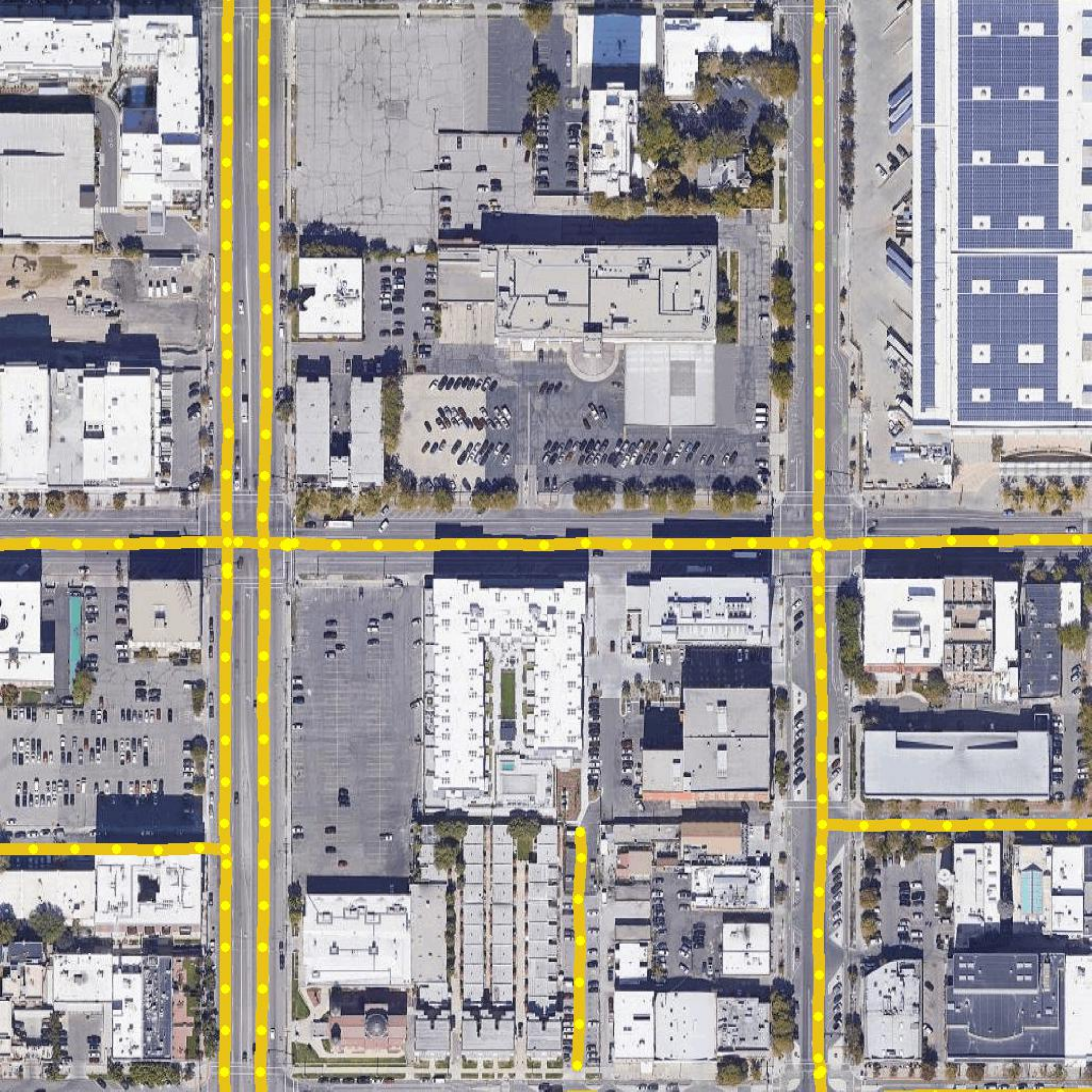}
        \end{subfigure}\vspace{.6ex}
        \begin{subfigure}[t]{\textwidth}
            \includegraphics[width=\textwidth]{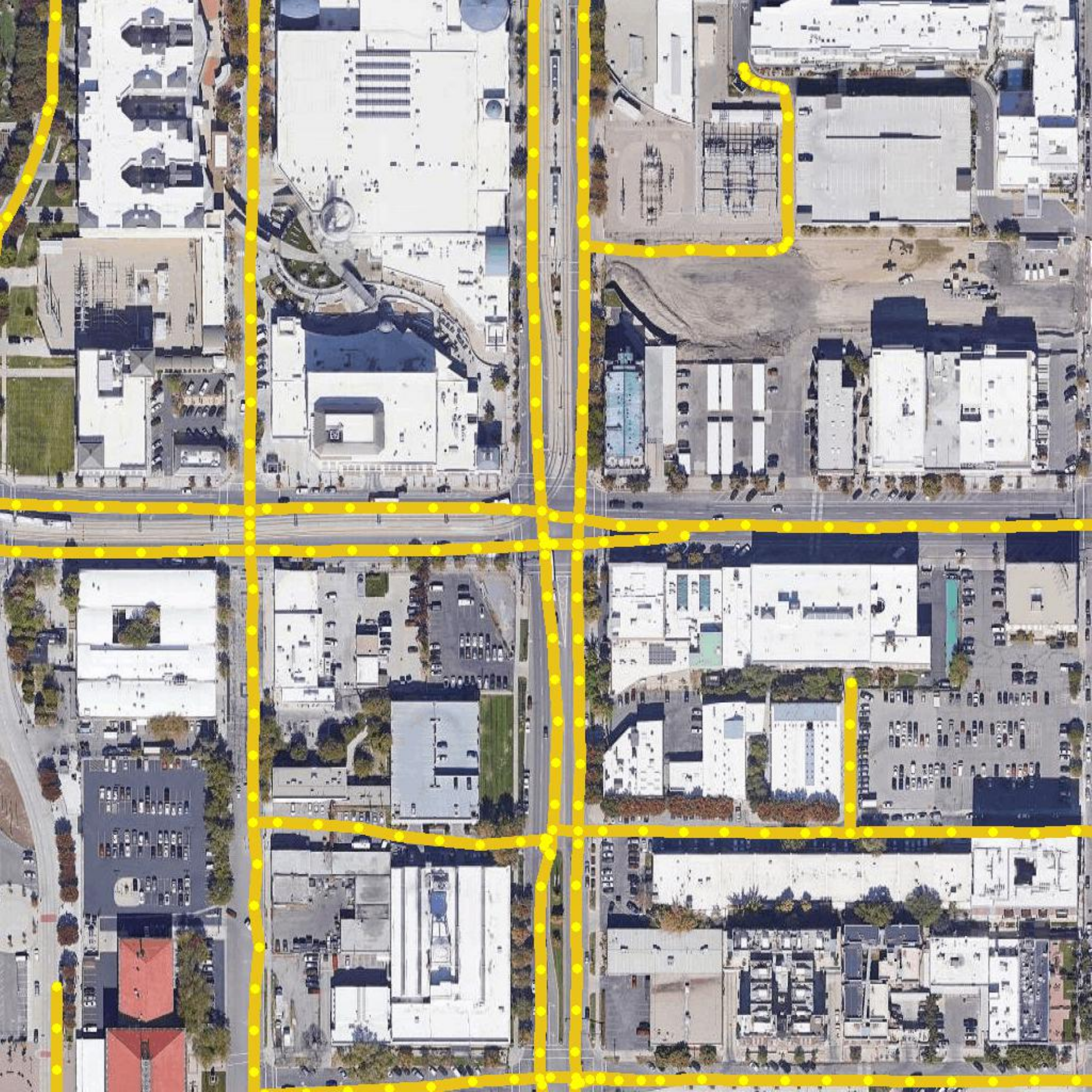}
        \end{subfigure}\vspace{.6ex}
        \begin{subfigure}[t]{\textwidth}
            \includegraphics[width=\textwidth]{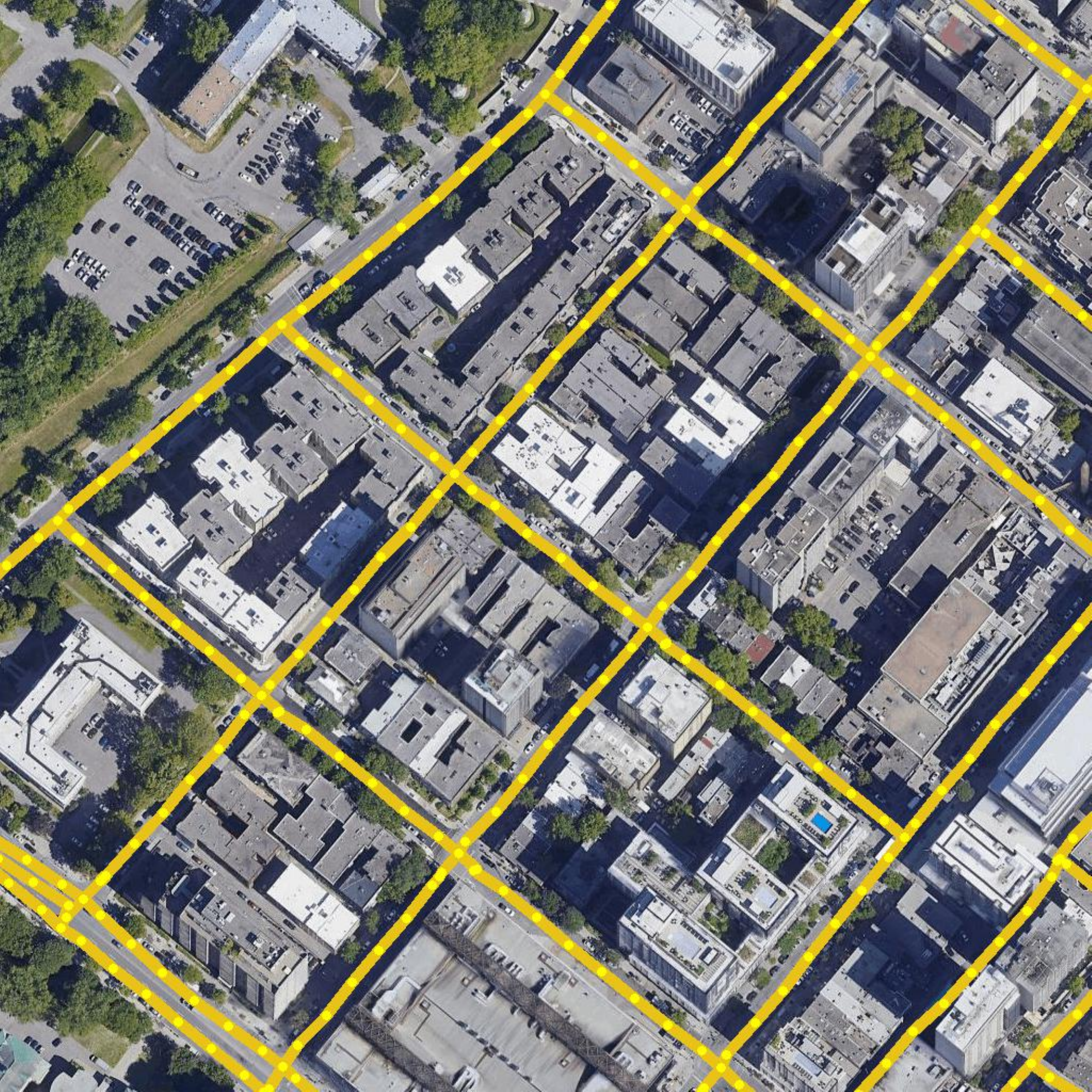}
        \end{subfigure}\vspace{.6ex}
        \begin{subfigure}[t]{\textwidth}
            \includegraphics[width=\textwidth]{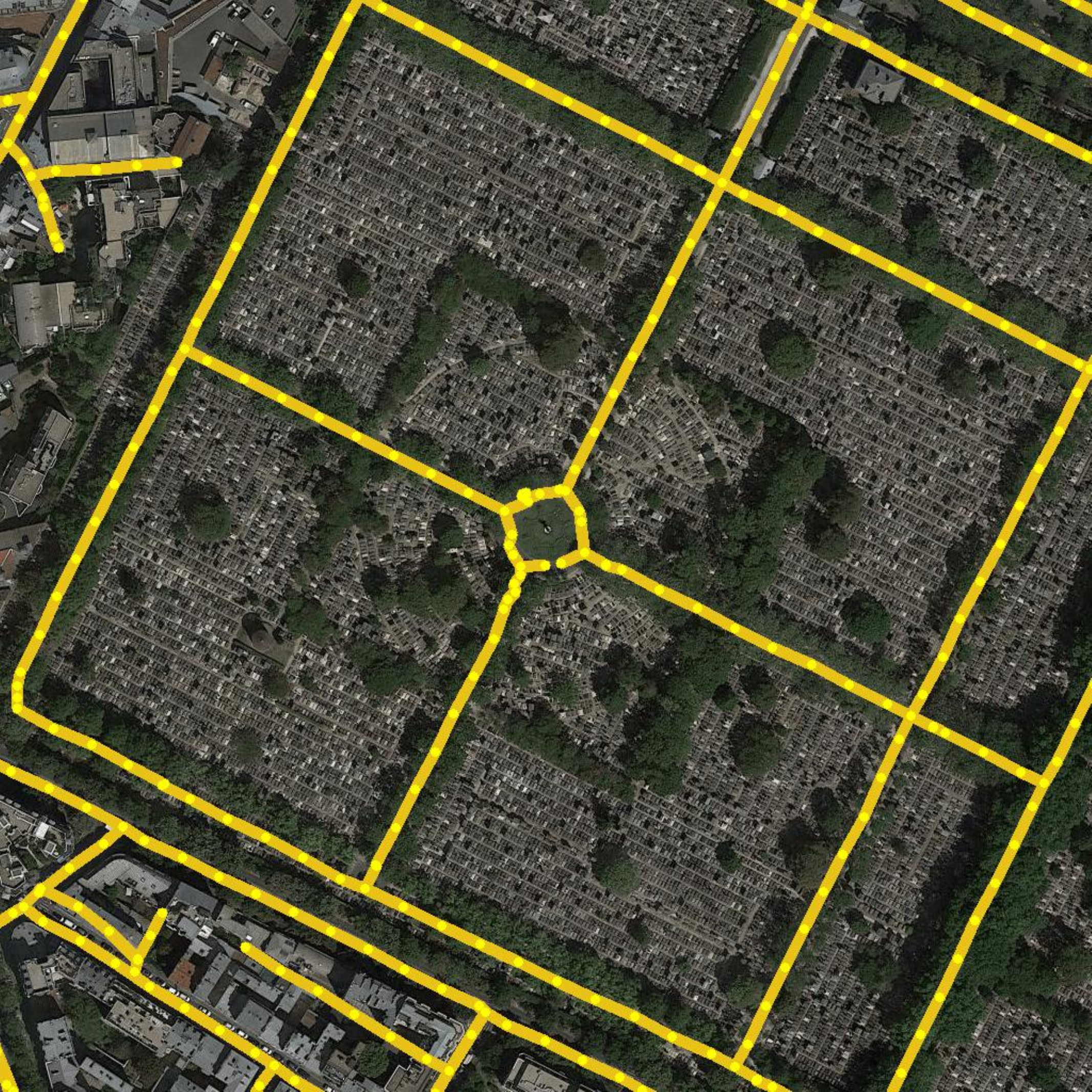}
        \end{subfigure}\vspace{.6ex}
        \begin{subfigure}[t]{\textwidth}
            \includegraphics[width=\textwidth]{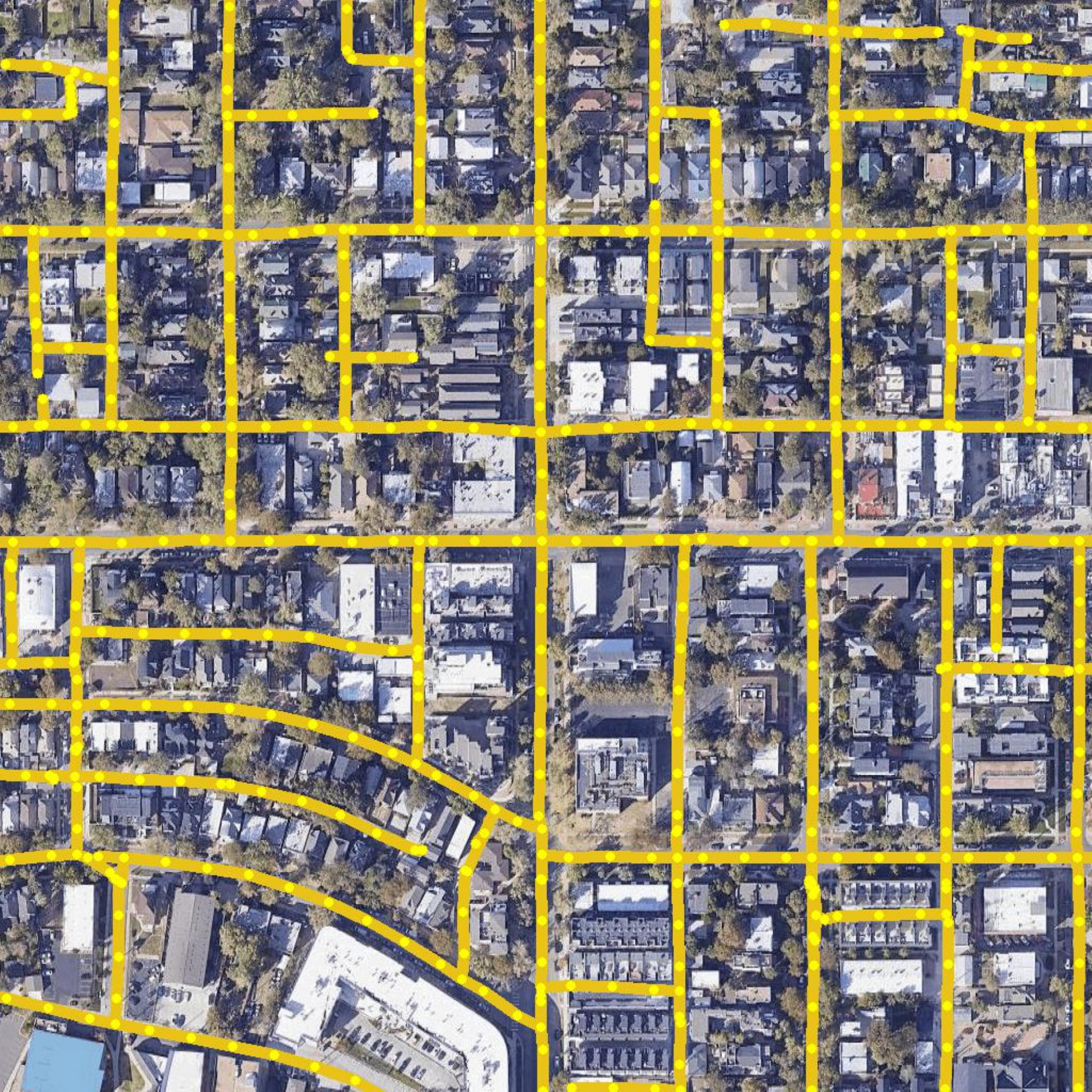}
        \end{subfigure}\vspace{.6ex}
        \begin{subfigure}[t]{\textwidth}
            \includegraphics[width=\textwidth]{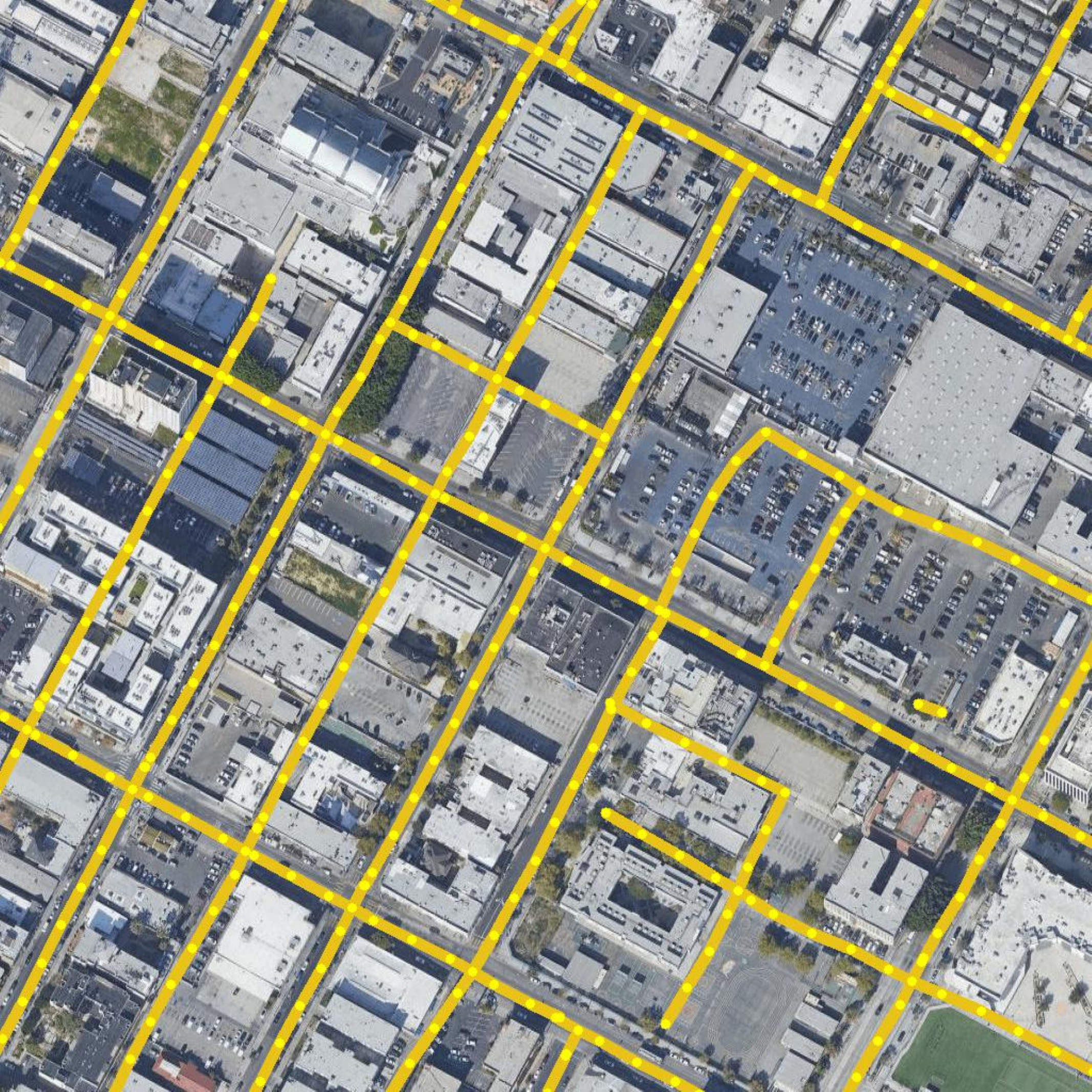}
        \end{subfigure}\vspace{.6ex}
        \begin{subfigure}[t]{\textwidth}
            \includegraphics[width=\textwidth]{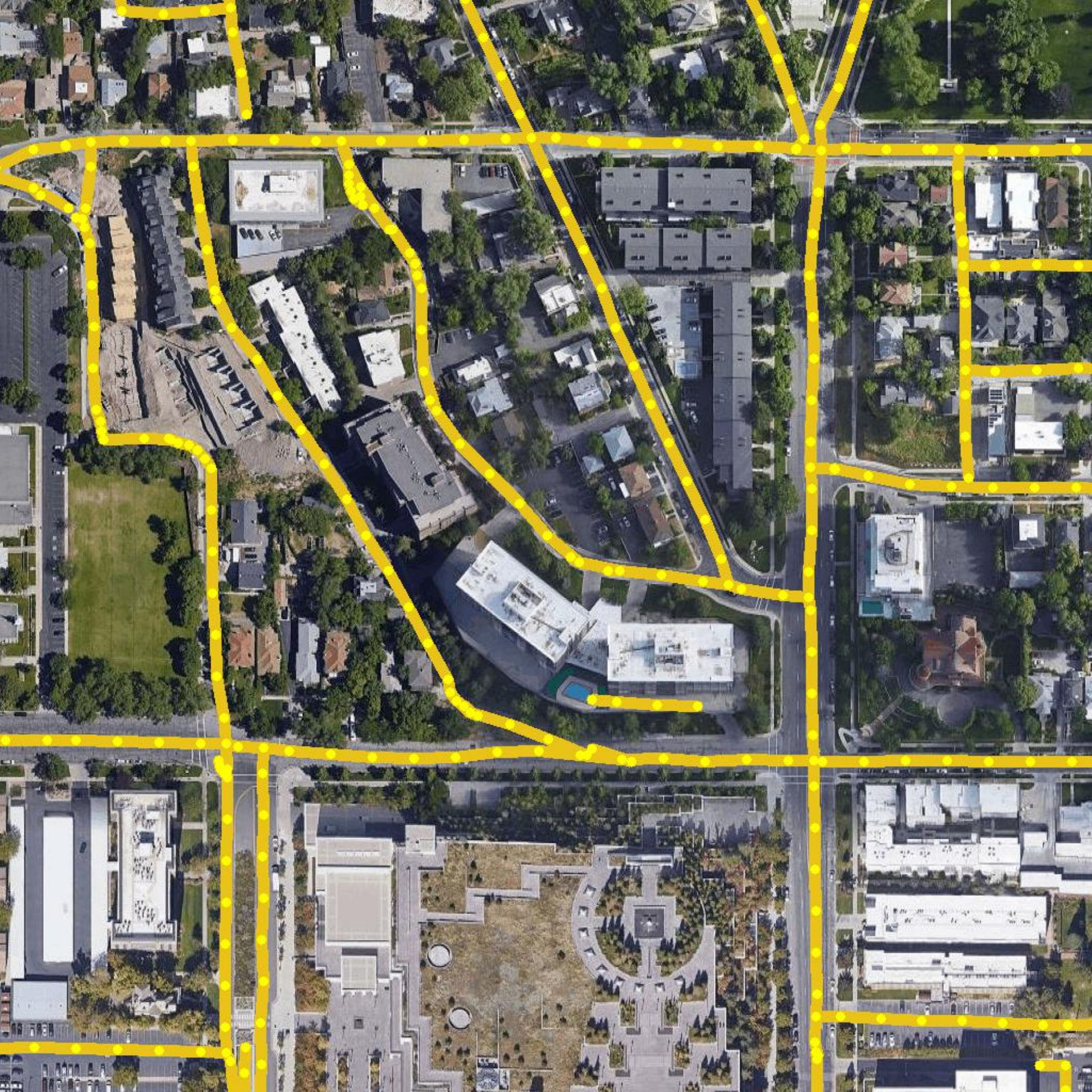}
        \end{subfigure}\vspace{.6ex}
        \caption{RNGDet}
        \label{fig_qualitative_1st}
    \end{subfigure}
    
    \caption{Qualitative demonstrations. We visualize the road network detection results on aerial images. The size of each image is $512\times 512$. (a) Ground-truth road network graph (cyan lines). (b)-(c) The road network graph predicted by segmentation-based approaches (orange lines). These two approaches have poor topology performance such as incorrect disconnections. (d)-(g) The road network graph predicted by graph-based approaches (orange lines as edges and yellow points as vertices). Compared with RoadTracer, VecRoad and Sat2Graph, RNGDet presents more precise graph structures. For better visualization, lines are widened but they are actually of one-pixel width. This figure is best viewed in color. Please zoom in for details.  }
    \label{fig_qualitative}
\end{figure*}

We compare our proposed RNGDet with two segmentation-based approaches and three graph-based approaches.
\begin{itemize}
    \item ImprovedRoad \cite{batra2019improved} (CVPR 2019): ImprovedRoad is one of the state-of-the-art segmentation-based approaches in the past. Orientation information is used to enhance the road segmentation, and it trains an extra refine network to fix incorrect road segmentation predictions.
    
    \item SPIN RoadMapper \cite{gedara2021spin} (ICRA 2022): Based on ImprovedRoad, SPIN RoadMapper proposes a graph reasoning scheme to further capture spatial information of the aerial image. ImprovedRoad and SPIN RoadMapper are trained for 120 epochs. These two approaches usually suffer from poor topology correctness.
    
    \item RoadTracer \cite{bastani2018roadtracer} (CVPR 2018): RoadTracer is believed to be the first graph-based approach. It predicts the directions of the vertices at the next step as a multi-class classification problem. However, it has a fixed step size and the training label generation algorithm may produce inappropriate labels.
    
    \item VecRoad \cite{tan2020vecroad} (CVPR 2020): VecRoad is an improved version of RoadTracer, and is the state-of-the-art graph-based approach. It predicts the distribution of the vertices at the next step, which allows flexible step size. But it is still two-stage and cannot be optimized in an end-to-end manner. Moreover, it may not be able to distinguish vertices that are close to each other.
    
    \item Sat2Graph \cite{he2020sat2graph} (ECCV 2020): Sat2Graph proposes a new graph encoding scheme and designs deep neural networks to predict the graph encoding of the input image. The predicted graph encoding can be decoded into road network graphs with satisfactory accuracy. However, it suffers from the isomorphic encoding issue which is analyzed in \cite{he2020sat2graph}, and it has relatively inferior performance when arc roads are encountered please refer to the fourth row of Fig. \ref{fig_qualitative} as an example).
\end{itemize}

\begin{table*}[!th] 
\setlength{\abovecaptionskip}{0pt} 
\setlength{\belowcaptionskip}{0pt} 
\renewcommand\arraystretch{1.0} 
\renewcommand\tabcolsep{2.3pt} 
\centering 
\begin{threeparttable}
\caption{The quantitative comparison results. The best results are highlighted in bold font. 
For all the metrics, larger values indicate better performance.
}
\begin{tabular}{@{}c c c c c c c c c c c c c c c c c c c c c c c c c@{}}
\toprule
\multirow{3}{*}{Approaches}& \multicolumn{3}{c}{P-P $\uparrow$} & \multicolumn{3}{c}{P-R $\uparrow$} & \multicolumn{3}{c}{P-F $\uparrow$}& \multicolumn{3}{c}{I-P $\uparrow$} & \multicolumn{3}{c}{I-R $\uparrow$} & \multicolumn{3}{c}{I-F $\uparrow$}& \multirow{3}{*}{APLS $\uparrow$}\\ 
\cmidrule(l){2-4} \cmidrule(l){5-7} \cmidrule(l){8-10} \cmidrule(l){11-13} \cmidrule(l){14-16} \cmidrule(l){17-19} 
&   2.0 &  5.0 &  10.0 &   2.0 &  5.0 &  10.0
&    2.0 &  5.0 &  10.0 &    2.0 &  5.0 &  10.0 &   2.0 &  5.0 &  10.0
&    2.0 &  5.0 &  10.0 & \\
\midrule
ImprovedRoad \cite{batra2019improved}


&68.22&75.70&78.72&47.50&54.09&57.80&56.01&63.10&66.66
&29.00&40.62&43.10&23.25&32.60&34.52&25.81&36.17&38.34&41.71\\ 
SPIN RoadMapper \cite{gedara2021spin} &


\textbf{78.53}&\textbf{85.82}&\textbf{89.02}&54.02&60.28&63.71&64.01&70.82&74.27&
\textbf{45.27}& 59.66&62.08&28.87&37.56&38.93&33.67&35.26&46.10&47.85
 \\
 RoadTracer \cite{bastani2018roadtracer} &
 
  57.49 & 68.26 & 74.53 &35.09 &41.81 &46.28 &43.58& 51.86&57.10 
  & 22.79&55.50&\textbf{78.93}&13.18&32.11&44.16&16.59&16.70&40.68&56.63
 \\
VecRoad \cite{tan2020vecroad} &

60.87 & 69.33& 73.97 & 64.91&74.00 &78.88 &62.83 &71.59 &76.35& 37.49&63.87&68.77&37.51&63.70&68.43&37.50&63.78&68.60&65.69
\\
Sat2Graph \cite{he2020sat2graph}
&57.87&65.33&69.43&63.02&71.33 &76.15&60.33&68.20&72.64&29.61&57.02&60.33&34.20&65.90&69.64&31.74&61.14&64.65&63.21
\\
\midrule

RNGDet (ResNet-34) & 58.33 & 69.07 & 73.81& 63.80 & \textbf{75.15} & 79.43&60.94 & 71.98 & 76.52 & 
32.83 & 57.02& 69.22 & 34.19 & 55.96 & 72.16 & 33.50 & 56.49 & 70.66 &  55.64\\
RNGDet (ResNet-50) & 62.11 & 69.54 & 74.39& 66.07 & 74.92 & 80.05&64.03 & 72.12 & 77.13 & 
 36.51 & 63.97&75.20 & 39.98 & 65.48 & 62.84 & 38.17 & 64.72 & 68.47 &  63.76\\
RNGDet (ResNet-101) &65.63 & 72.31 & 77.08& \textbf{66.42} & 75.08& \textbf{82.13}&\textbf{66.02} &\textbf{73.67} & \textbf{79.52} & 
42.37 & \textbf{65.30}& 72.18 & \textbf{40.40} & \textbf{66.50} & \textbf{73.23} & \textbf{41.36} & \textbf{65.89} & \textbf{72.70} &  \textbf{67.88}
\\
\bottomrule 
\label{tab_comparative_patch}
\end{tabular} 
\end{threeparttable}
\end{table*}
\begin{table*}[!h] 
\setlength{\abovecaptionskip}{0pt} 
\setlength{\belowcaptionskip}{0pt} 
\renewcommand\arraystretch{1.0} 
\renewcommand\tabcolsep{3.4pt} 
\centering 
\begin{threeparttable}
\caption{The quantitative results for the ablation study. The best results are highlighted in bold font. For all the metrics, larger values indicate better performance. We assess the road segment segmentation ($\mathcal{S}$), the road intersection segmentation ($\mathcal{I}$) and the transformer structure (T).} 
\begin{tabular}{c c c c c c c c c c c c c c c c c c c c c c}
\toprule
&&& \multicolumn{3}{c}{P-P $\uparrow$} & \multicolumn{3}{c}{P-R $\uparrow$} & \multicolumn{3}{c}{P-F $\uparrow$}& \multicolumn{3}{c}{I-P $\uparrow$} & \multicolumn{3}{c}{I-R $\uparrow$} & \multicolumn{3}{c}{I-F $\uparrow$}& \multirow{3}{*}{APLS $\uparrow$}\\ 
\cmidrule(l){4-6} \cmidrule(l){7-9} \cmidrule(l){10-12} \cmidrule(l){13-15} \cmidrule(l){16-18} \cmidrule(l){19-21}
$\mathcal{S}$ & $\mathcal{I}$ & T&   2.0 &  5.0 &  10.0 &   2.0 &  5.0 &  10.0
&    2.0 &  5.0 &  10.0 &    2.0 &  5.0 &  10.0 &   2.0 &  5.0 &  10.0
&    2.0 &  5.0 &  10.0 & \\
\midrule
& &\checkmark  &37.63&56.80&61.79&55.24&60.12&64.77&44.77&58.41&63.24&25.40&47.03&69.62&20.70&40.93&53.26&22.81 & 43.77 & 60.35&50.32\\
&\checkmark&\checkmark &40.09&55.83&65.97&57.41&63.28&66.21&47.21&59.32&66.09&36.95&56.38&73.72&33.39&58.64&64.30&35.08 & 57.49 & 68.69&61.22\\
\checkmark& &\checkmark &59.80&65.34&72.79&62.19&73.83&79.02&60.97&69.33&75.78&32.14&50.86&68.44&32.62&55.04&62.47&32.38 & 52.87 & 65.32&59.90\\
\checkmark&\checkmark&   &56.04&69.91&74.52&62.88&71.64&74.56&59.26&70.76&74.54&31.65&57.21&63.90&32.10&59.33&62.42&31.87&58.25&63.15&60.79\\
\midrule
\checkmark&\checkmark&\checkmark  &\textbf{65.63} & \textbf{72.31} & \textbf{77.08}& \textbf{66.42} & \textbf{75.08 }& \textbf{82.13}&\textbf{66.02} &\textbf{73.67} & \textbf{79.52} & 
\textbf{42.37} & \textbf{65.30}& \textbf{72.18} & \textbf{40.40} & \textbf{66.50} & \textbf{73.23} & \textbf{41.36} & \textbf{65.89} & \textbf{72.70} &  \textbf{67.88}
 \\
\bottomrule 
\label{tab_ablation}
\end{tabular} 
\end{threeparttable}
\end{table*}

\subsection{Evaluation metrics}
In our experiments, we use three metrics pixel-precision (P-P), pixel-recall (P-R) and pixel-F1-score (P-F) for pixel-level evaluation, three metrics intersection-precision (I-P), intersection-recall (I-R) and intersection-F1-score (I-F) for intersection point evaluation and one metric average path length similarity (APLS) \cite{van2018spacenet} for topology correctness evaluation. 

To calculate the pixel-level metric scores, we rasterize the ground-truth graph and the predicted graph as binary images $B^*$ and $\hat{B}$, respectively. For a pixel in $B^*$, if there exists a pixel in $\hat{B}$ and the Euclidean distance between them is smaller than $\delta$, then this pixel is treated as correctly retrieved. Similarly, if a pixel in $\hat{B}$ can find a pixel in $B^*$ within $\delta$ distance, then we say this pixel is correctly detected. In this way, P-P, P-R and P-F can be obtained by the following equations:
\begin{equation}
\begin{aligned}
    \text{P-P} &= \frac{|\{
        p | \lVert p,q\rVert<\delta,\exists q\in B^*,\forall p\in \hat{B}
        \}|}{|\hat{B}|},\\
        \text{P-R} &= \frac{|\{
        p | \lVert p,q\rVert<\delta,\exists q\in \hat{B},\forall p\in B^*
        \}|}{|B^*|},\\
        \text{P-F} &=\frac{2\text{P-P}\cdot \text{P-R}}{\text{P-P}+\text{P-R}},
        \end{aligned}
\end{equation}
where $\lVert \cdot \rVert$ calculates the Euclidean distance and $|\cdot|$ is the cardinality of a set. Threshold $\delta$ measures the level of error tolerance, and we show the metric scores with $\delta$ as 2, 5 and 10 pixels for more comprehensive evaluation.

I-P, I-R and I-F are calculated in a similar way as the aforementioned pixels-level metrics. The only difference is that instead of rasterizing the whole graph into binary images $B^*$ and $\hat{B}$, these three metrics only care about intersection points (i.e., rasterized binary images only contain intersection points). These three metrics evaluate the ability of approaches to detect road intersections. We also show these metric scores in our experimental results with different $\delta$.

APLS measures the similarity of the ground-truth graph $G^*$ and the predicted graph $\hat{G}$. It samples multiple vertex pairs on both graphs and compares the difference between the shortest distance of vertex pairs. APLS is calculated by the following equation:
\begin{equation}
    \text{APLS} = 1 -\frac{1}{N_s}\sum^{N_s} \text{min}\left(1,\frac{|L(a,b)-L(a',b')|}{L(a,b)}\right),
\end{equation}
where $(a,b)$ demonstrates a vertex pair sampled from $G^*$ and $(a',b')$ is the corresponding vertex pair sampled from $\hat{G}$. The number of sampled vertex pairs is denoted by $N_s$, and $L(\cdot,\cdot)$ calculates the length of the shortest path between two vertices. Higher APLS indicates better graph similarity and topology correctness.

\subsection{Comparative results}
 
RNGDet is compared with five baseline approaches, including two segmentation-based baselines and three graph-based baselines. The quantitative comparison results are shown in Tab. \ref{tab_comparative_patch}. In Tab. \ref{tab_comparative_patch}, besides baseline approaches, we also evaluate RNGDet with different backbones (i.e., ResNet-34, ResNet-50 and ResNet-101) for more fair and comprehensive comparison. Qualitative visualizations are provided in Fig. \ref{fig_qualitative}.

For segmentation results, we first predict the segmentation map, and then binarize it by thresholding. Finally, we run skeletonization algorithms to extract the graph of predicted segmentation maps. From the comparison results, we find that these approaches have good pixel-level results since they directly optimize pixel-level segmentation. However, since they cannot fully utilize spatial and geometric information, they have poor performance on topology correctness. Thus segmentation-based approaches have relatively inferior intersection-level and topology-level scores. Therefore, segmentation-based approaches are not sufficient for our road network graph detection task.

Graph-based approaches directly optimize the graph, thus they present much better results from the topology perspective. RoadTracer has a fixed step length, which makes it unable to handle some scenarios very well, especially when the agent is near road intersections. VecRoad is more powerful due to the use of Res2Net \cite{gao2019res2net} backbone network and the flexible step size. However, VecRoad predicts vertices by finding local peaks on heatmaps, so when vertices are very close to each other, it may not be able to correctly detect them. Therefore, it may fail to correctly detect some precise graph structures. Sat2Graph presents good results on straight road detection, but it cannot handle curve roads very well, which degrades its final performance (please refer to the fourth row of Fig. \ref{fig_qualitative} as an example). Our proposed RNGDet can directly output the coordinates of vertices, therefore RNGDet can handle more complicated situations and has more powerful performance. RNGDet with ResNet-101 backbone has the best performance during evaluation, gaining 2-5\% improvement on almost all metrics scores compared with past state-of-the-art approaches. RNGDet with ResNet-50 backbone has a little bit inferior performance, but it still outperforms VecRoad except APLS. RNGDet with ResNet-34 backbone presents unsatisfactory results, which may be caused by the shallow backbone network and FPN segmentation heads. 

\subsection{Ablation studies}

In this section, we study the significance of some components of our network design, including the road segment segmentation $\mathcal{S}$, road intersection segmentation $\mathcal{I}$ and the transformer structure of RNGDet. All ablation studies are conducted on RNGDet with ResNet-101 backbone. The quantitative results of our ablation studies are shown in Tab. \ref{tab_ablation}.

First, we completely remove the segmentation heads from RNGDet. From the evaluation results, we notice that the performance of RNGDet drops a lot because the segmentation heads can provide strong supervision to assist the CNN backbone to extract features of the input. Then, we evaluate RNGDet by only removing the road segment segmentation $\mathcal{S}$. All metrics scores degrade, especially pixel-level metrics scores. After this, we remove the road intersection segmentation $\mathcal{I}$ only, and the intersection-level metrics scores and APLS are severely harmed. Because $\mathcal{I}$ is critical to make RNGDet aware of road intersections, removing $\mathcal{I}$ will lead to incorrect connections near road intersections and make the topology correctness much worse. In this way, the necessity of the segmentation maps including $\mathcal{S}$ and $\mathcal{I}$ is verified.

Finally, to examine the importance of the transformer structure in RNGDet, we replace the transformer with a modified Mask-RCNN \cite{he2017mask}. In our task, similar to VecRoad \cite{tan2020vecroad}, Mask-RCNN first predicts the heatmap of vertices at the next time step, and then extracts the coordinate of vertices by handcrafted
or heuristic post-processing. Thus, RNGDet with Mask-RCNN cannot be optimized in an end-to-end way either, which affects its final performance. From the experimental results, RNGDet with Mask-RCNN presents inferior results compared with the original RNGDet. Therefore, the importance of the transformer structure is proved.

\subsection{Number of vertex queries}
Under normal circumstances, the number of queries $|Q|$ should be obviously larger than the maximum number of vertices at the next time step. Usually, the road intersections have at most 5 roads incident with each other. Thus, we try to train RNGDet with 5, 10 and 20 queries and observe the obtained performance. We use RNGDet with ResNet-101 for the experiments. During the experiments, we use metrics P-F ($\delta=5$), I-F ($\delta=5$) and APLS to evaluate models.

Based on the results shown in Tab. \ref{tab_comparative_queries}, RNGDet with 10 input queries presents the best performance, thus the number of queries is set to 10.
 
\begin{table}[h] 
\setlength{\abovecaptionskip}{0pt} 
\setlength{\belowcaptionskip}{0pt} 
\renewcommand\arraystretch{1.0} 
\renewcommand\tabcolsep{16.5pt} 
\centering 
\begin{threeparttable}
\caption{The quantitative results of RNGDet with different numbers of queries ($|Q|$).} 
\begin{tabular}{@{} c c c c c c c c c c c c c c c@{}}
\toprule
\multirow{4}{*}{}& P-F $\uparrow$ & I-F $\uparrow$ & APLS $\uparrow$ \\ 
\midrule
 RNGDet with 5 queries & 73.10 & 61.74 & 62.78\\
 RNGDet with 10 queries & 73.67 &  65.89 & 67.88\\
 RNGDet with 20 queries & 73.90 & 64.27 & 65.31\\
\bottomrule 
\label{tab_comparative_queries}
\end{tabular} 
\end{threeparttable}
\end{table}

\subsection{Failure cases}
\begin{figure}[t]
 \begin{subfigure}[t]{0.158\textwidth}
        \begin{subfigure}[t]{\textwidth}
            \includegraphics[width=\textwidth]{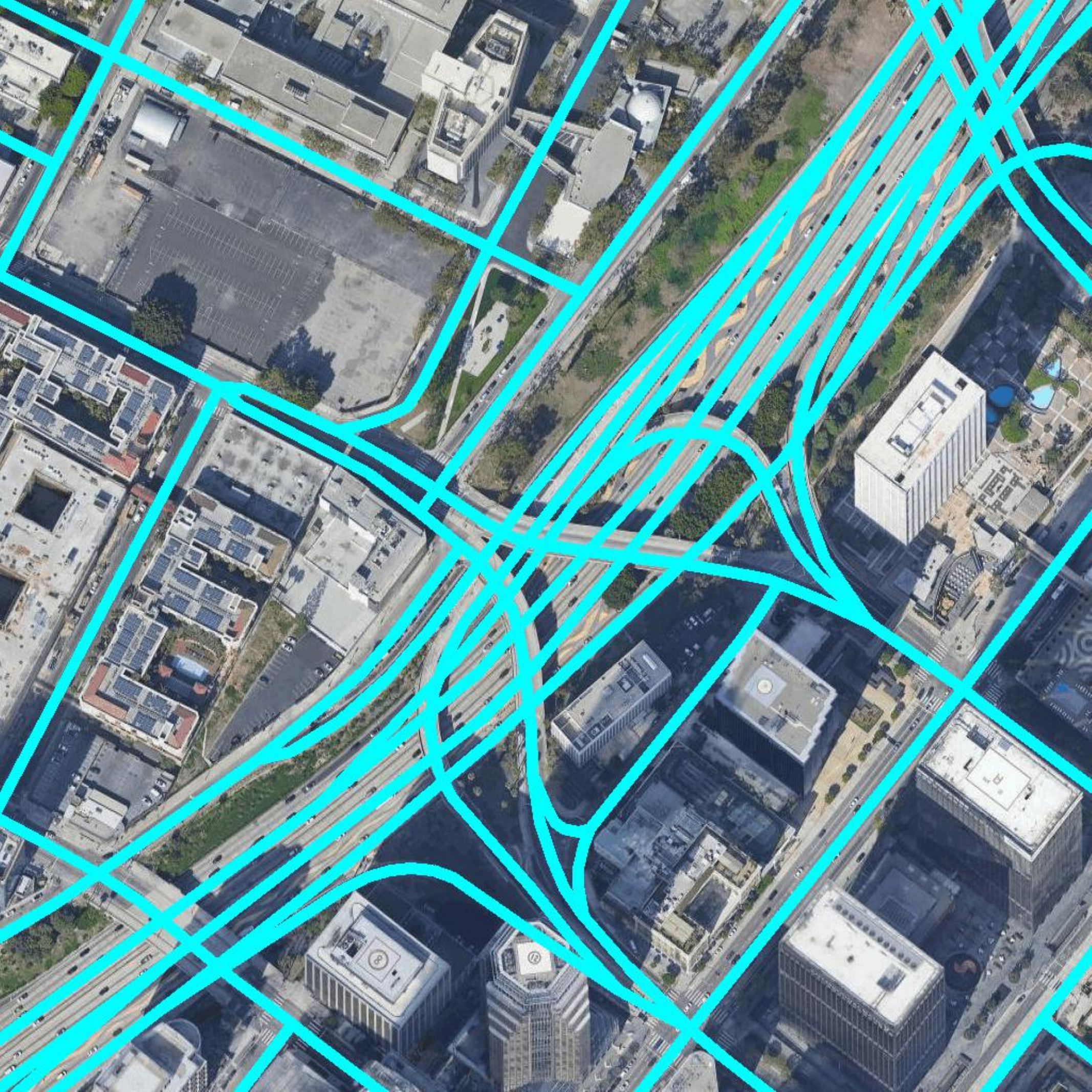}
        \end{subfigure}\vspace{.6ex}
        \begin{subfigure}[t]{\textwidth}
            \includegraphics[width=\textwidth]{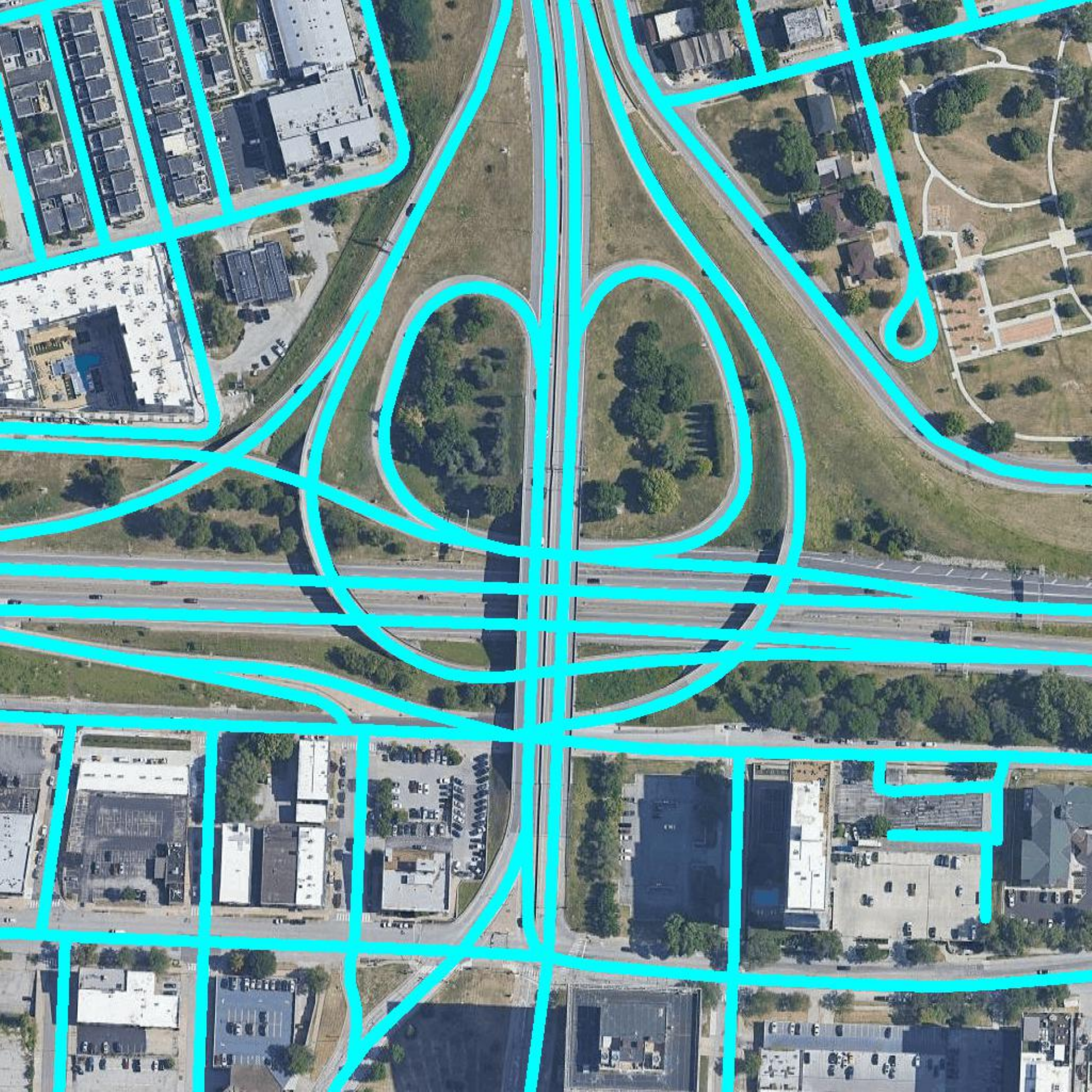}
        \end{subfigure}\vspace{.6ex}
        \begin{subfigure}[t]{\textwidth}
            \includegraphics[width=\textwidth]{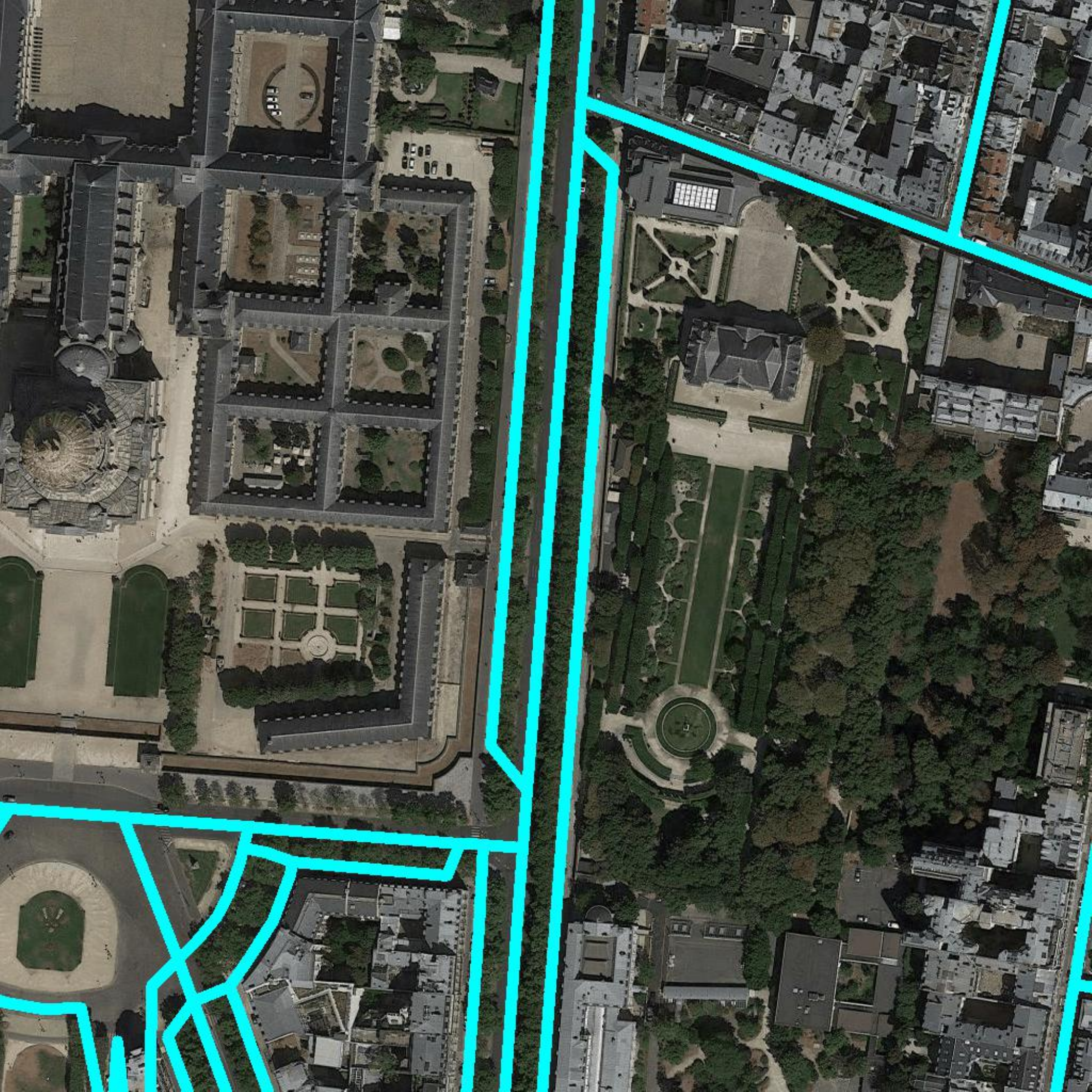}
        \end{subfigure}
        \caption{Ground truth}
    \end{subfigure}
    \begin{subfigure}[t]{0.158\textwidth}
        \begin{subfigure}[t]{\textwidth}
            \includegraphics[width=\textwidth]{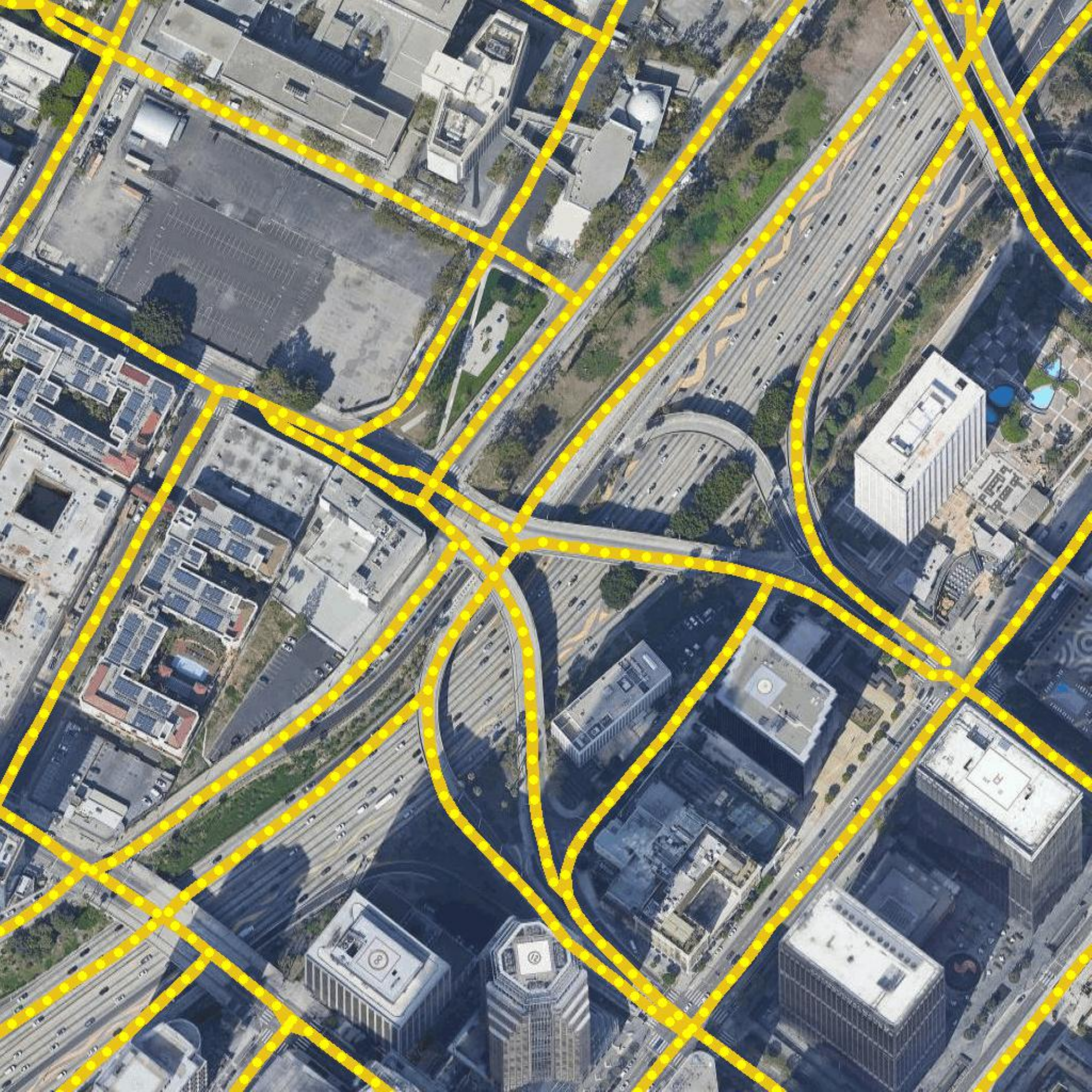}
        \end{subfigure}\vspace{.6ex}
        \begin{subfigure}[t]{\textwidth}
            \includegraphics[width=\textwidth]{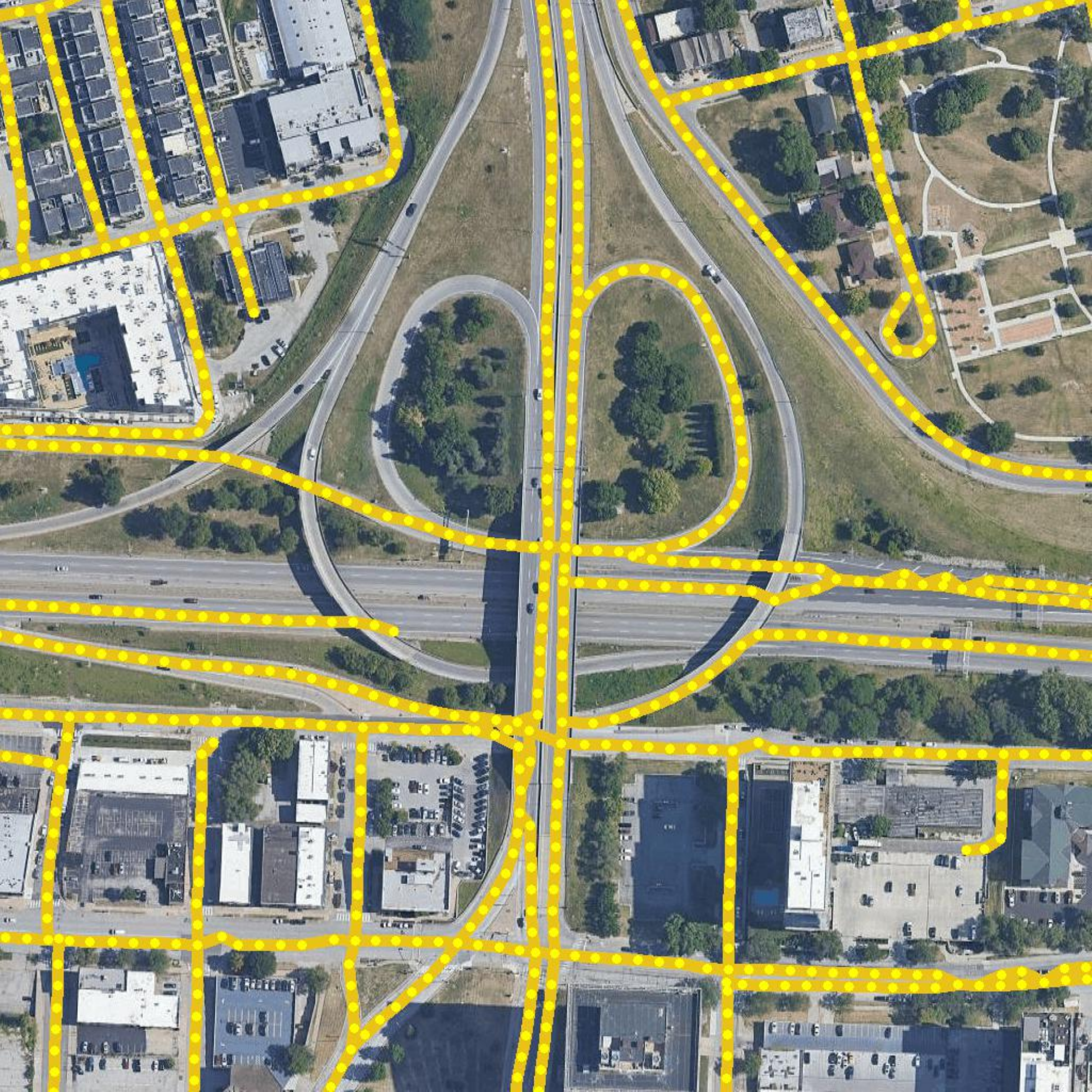}
        \end{subfigure}\vspace{.6ex}
        \begin{subfigure}[t]{\textwidth}
            \includegraphics[width=\textwidth]{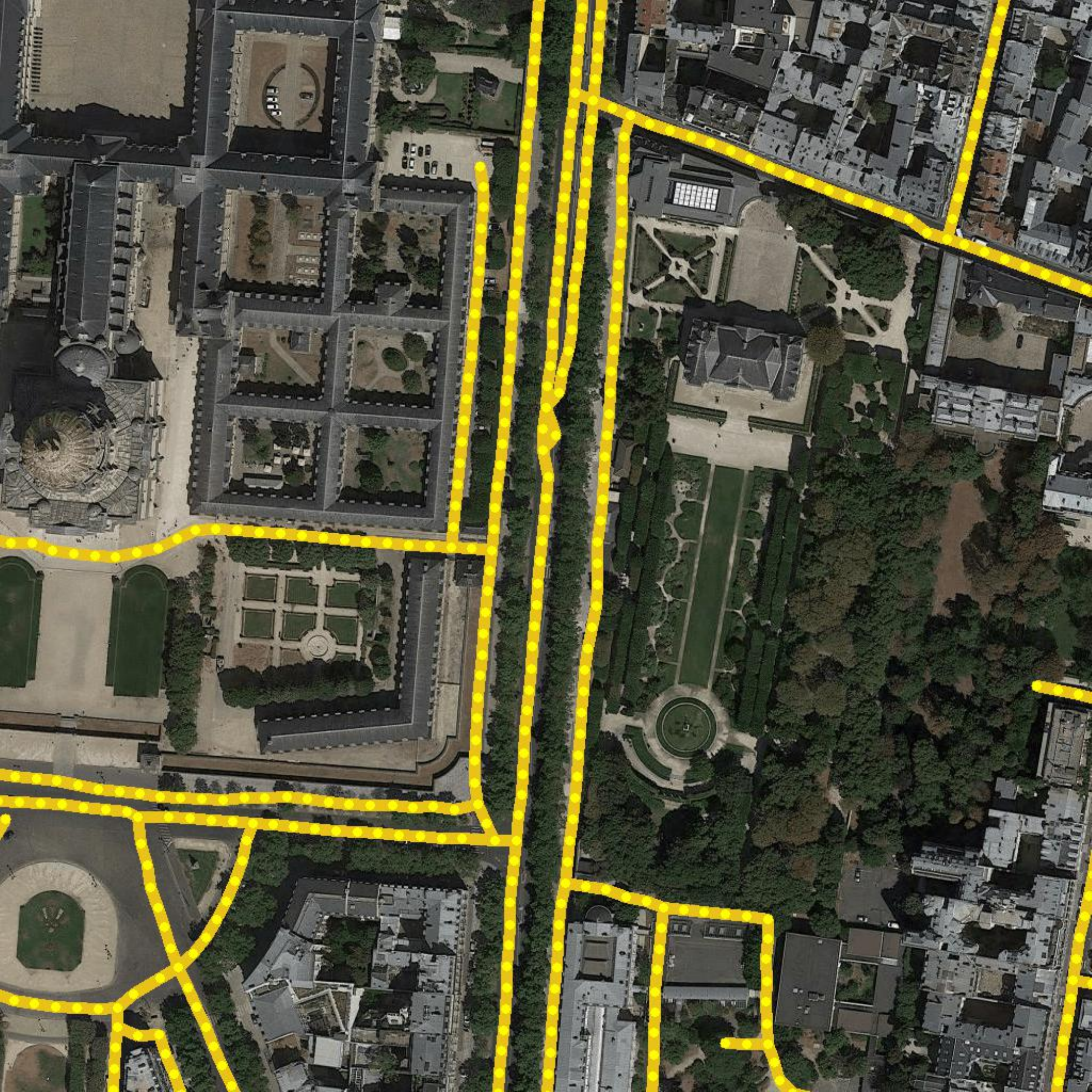}
        \end{subfigure}
        \caption{VecRoad}
    \end{subfigure}
    \begin{subfigure}[t]{0.158\textwidth}
        \begin{subfigure}[t]{\textwidth}
            \includegraphics[width=\textwidth]{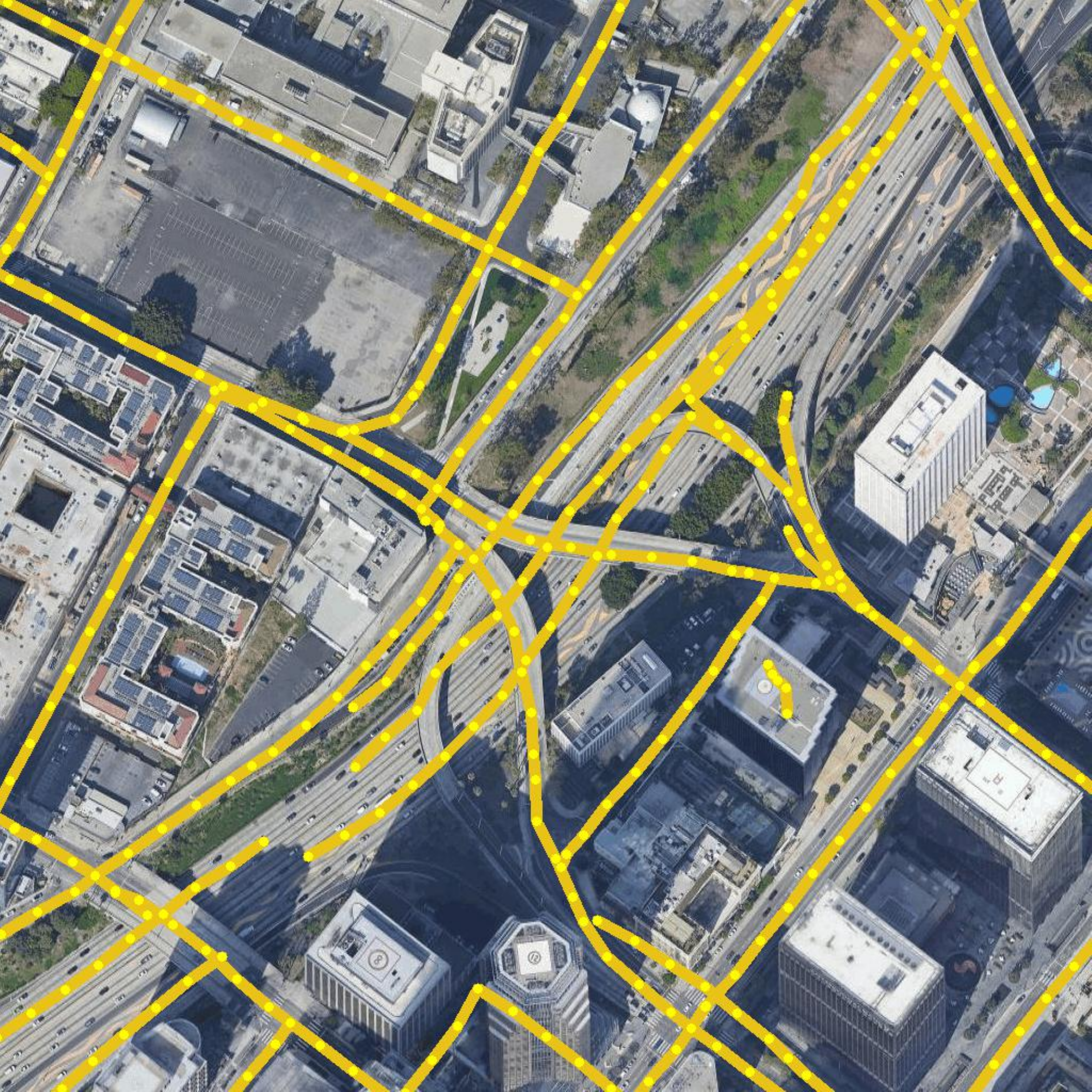}
        \end{subfigure}\vspace{.6ex}
        \begin{subfigure}[t]{\textwidth}
            \includegraphics[width=\textwidth]{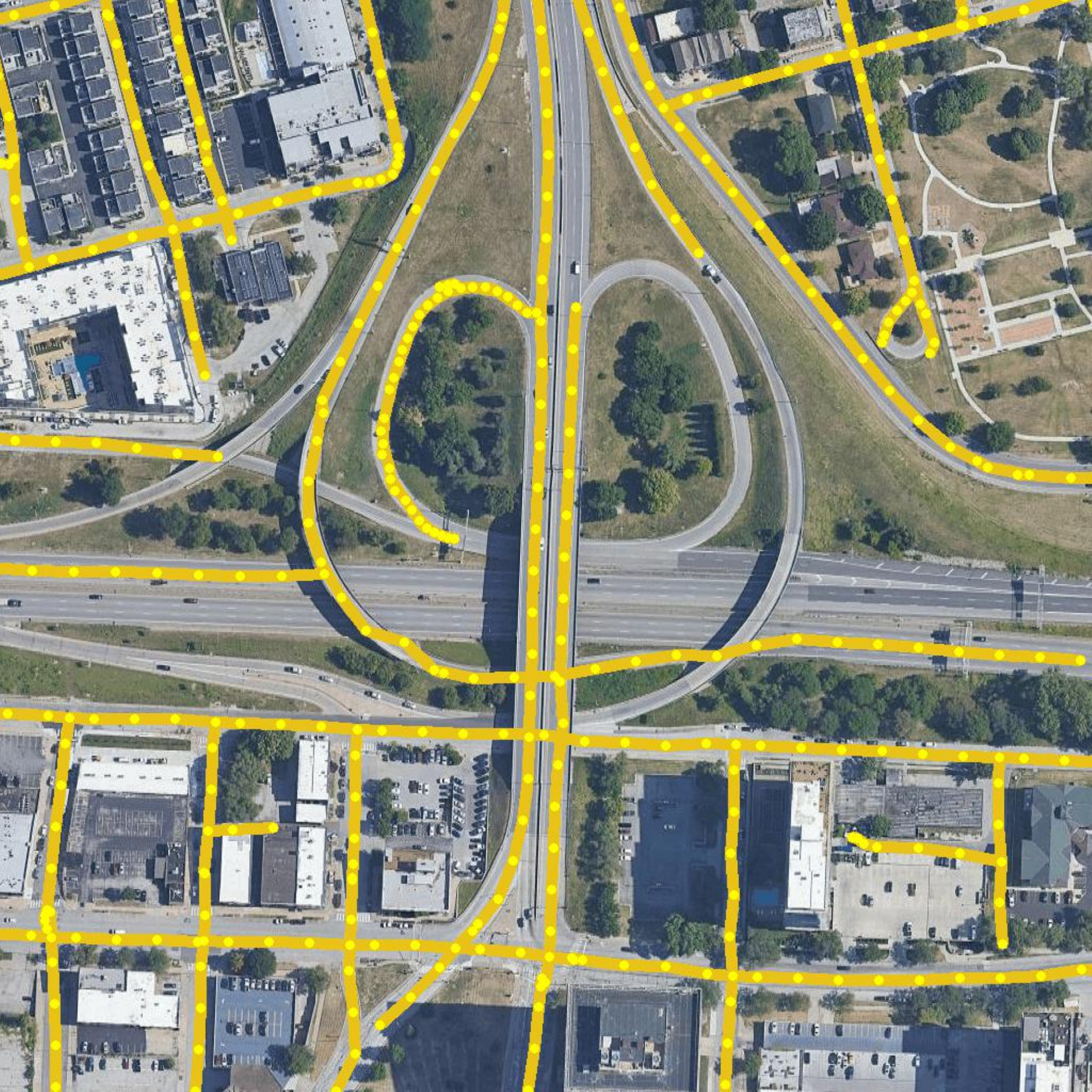}
        \end{subfigure}\vspace{.6ex}
        \begin{subfigure}[t]{\textwidth}
            \includegraphics[width=\textwidth]{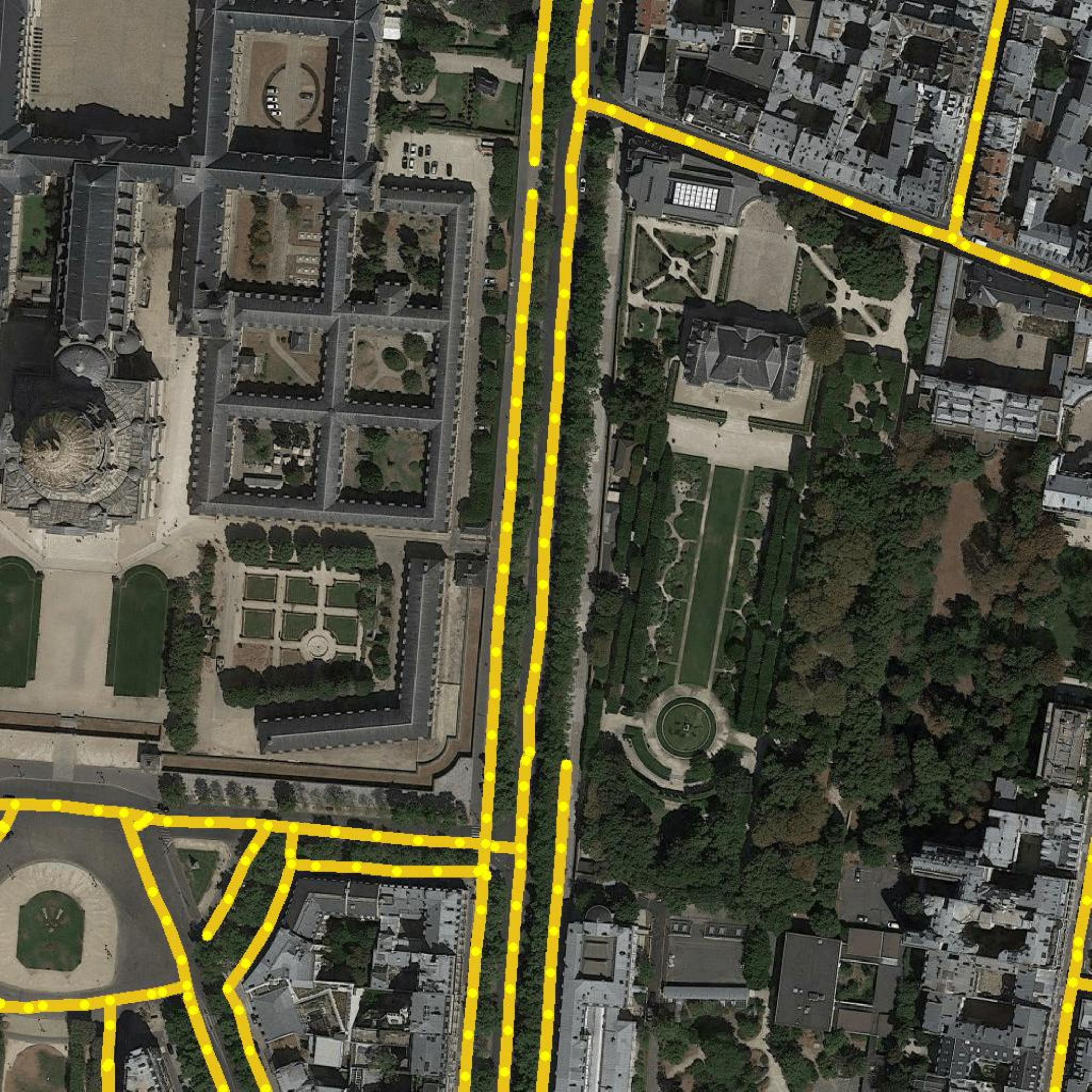}
        \end{subfigure}
        \caption{RNGDet}
    \end{subfigure}
    \caption{Qualitative demonstrations of the failure cases. (a) The ground-truth (cyan lines); (b) The result of VecRoad (orange lines for edges and yellow points for vertices); (c) The result of RNGDet (orange lines for edges and yellow points for vertices). The first and second rows show complicated road intersections and overpasses with occlusion. The last row is an example containing severe tree occlusion. RNGDet at this stage cannot handle these cases very well yet. This problem could be relieved in the future by using more powerful backbone networks or training strategies for RNGDet. The figure is best viewed in color. Please zoom in for details.}
    \label{fig_failure_case}
\end{figure}

Although RNGDet presents superiority against past approaches, it still cannot handle some very complicated cases, such as occluded overpasses, very well yet. Moreover, RNGDet also suffers from the drifting issue of imitation learning (i.e., when the agent is away from the right track, it may not be able to get back to the correct state) similar to other graph-based approaches, even though it can relieve this problem to some extent. Some example visualizations of failure cases of RNGDet are shown in Fig. \ref{fig_failure_case}. These cases could be handled in the future with more powerful backbone networks or training strategies.

\section{Conclusions and Future work}
We proposed here RNGDet, a novel iterative approach to automatically detect the road network graph from aerial images. Taken as input an aerial image, RNGDet could directly output the road network graph with vertices and edges. First, RNGDet predicted a set of candidate initial vertices, and then iteratively generated the road network graph vertex-by-vertex starting from each candidate initial vertex. Due to the use of transformer and deep queries, RNGDet could handle complicated intersection points with an arbitrary number of incident road segments. RNGDet was evaluated on a publicly available dataset and presented the state-of-the-art performance in terms of all evaluation metrics, including pixel-level metrics, intersection-level metrics and topology-level metric APLS. The experimental results demonstrated the superiority of our work. In the future, we would like to further improve RNGDet by using more powerful backbone networks and training strategies. Besides, we will adapt RNGDet to other graph detection tasks, such as road laneline detection and road lane centerline detection.

\bibliographystyle{IEEEtran}
\bibliography{mybib}

\begin{IEEEbiography}[{\includegraphics[width=1in,height=1.25in,clip,keepaspectratio]{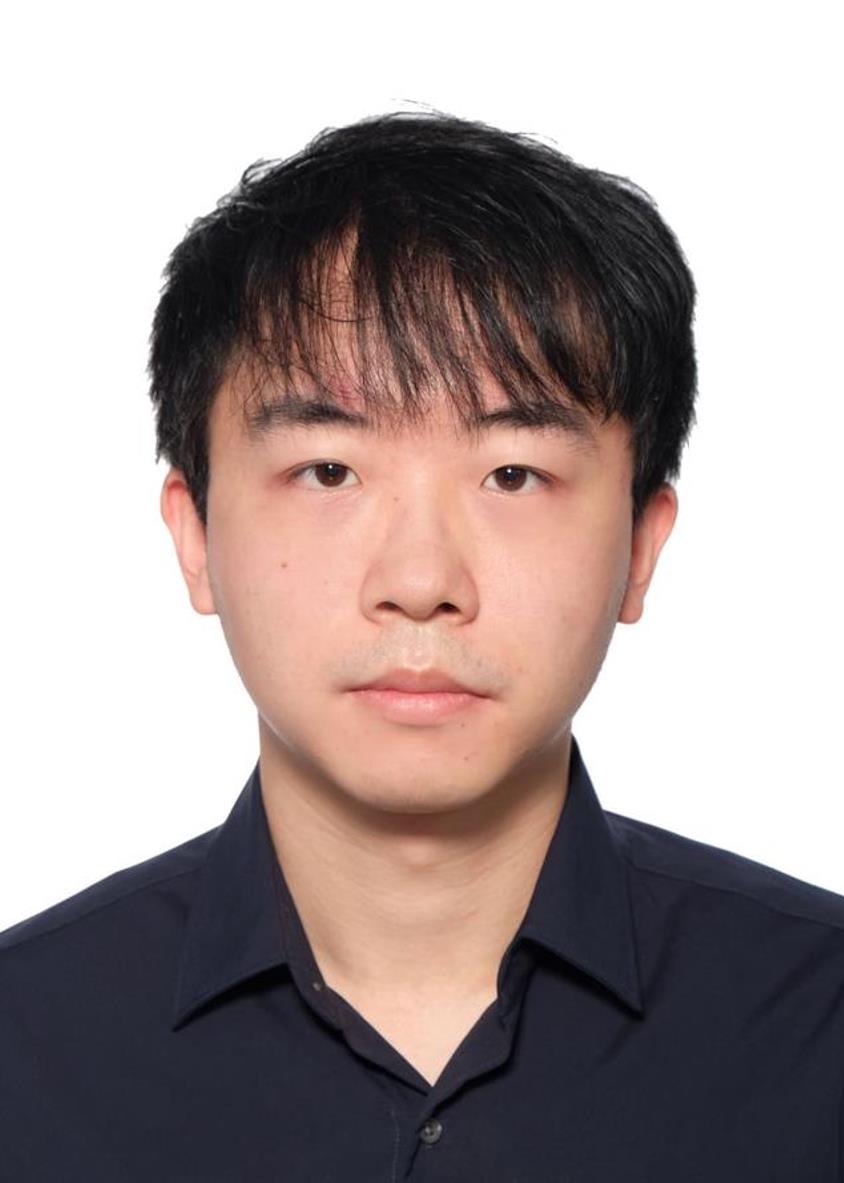}}]{Zhenhua Xu} (Student Member 2022) received the bachelor's degree from Harbin Institute of Technology, Harbin, China, in 2018. He is now a PhD candidate supervised by Prof. Ming Liu and Prof. Huamin Qu at the Department of Computer Science and Engineering, The Hong Kong University of Science and Technology, HKSAR, China. His current research interests include HD map automatic annotation, line-shaped object detection, imitation learning, autonomous driving, \textit{etc}. 
\end{IEEEbiography}

\begin{IEEEbiography}[{\includegraphics[width=1in,height=1.25in,clip,keepaspectratio]{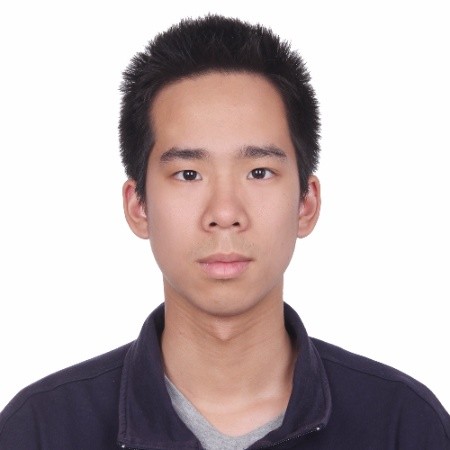}}]{Yuxuan Liu} (Student Member 2022) Yuxuan Liu received his Bachelor's degree from Zhejiang University, Zhejiang, China in 2019, majoring in Mechatronic. He is now a Ph.D candidate at the Department of Electronic and Computer Engineering, The Hong Kong University of Science and Technology, Hong Kong, China. His current research interests include autonomous driving, deep learning, robotics, visual 3D object detection, visual depth prediction, etc.
\end{IEEEbiography}

\begin{IEEEbiography}[{\includegraphics[width=1in,height=1.25in,clip,keepaspectratio]{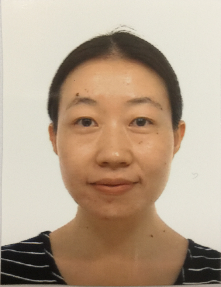}}]{Lu Gan} (Student Member 2022) received both her bachelor and master degreee from Nanjing University of Aeronautics and Astronautics. After that, she worked in Nanyang Technological University as a research associate. Now she is a PhD student in Hong Kong University od Science and Technology (GZ). Her research interests include autonomous driving, deep learning and uncertainty-aware motion prediction and planning. 

\end{IEEEbiography}

\begin{IEEEbiography}[{\includegraphics[width=1in,height=1.25in,clip,keepaspectratio]{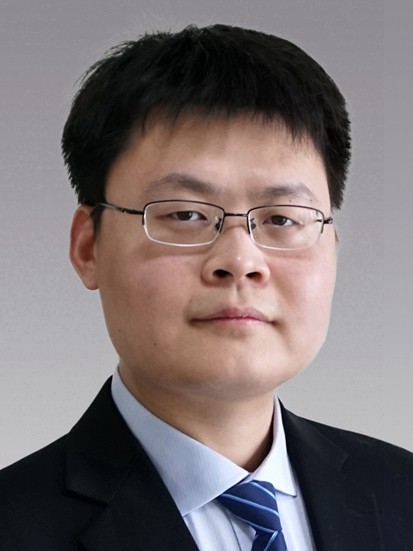}}]{Yuxiang Sun} (Member 2022) received the bachelor’s degree from the Hefei University of Technology, Hefei, China, in 2009, the master’s degree from the University of Science and Technology of China, Hefei, in 2012, and the Ph.D. degree from The Chinese University of Hong Kong, Hong Kong, in 2017.
 
He is now a research associate at the Department of Electronic and Computer Engineering, The Hong Kong University of Science and Technology, Hong Kong, China. His current research interests include autonomous driving, deep learning, robotics and autonomous systems, semantic scene understanding, etc. 

He is a recipient of the Best Paper in Robotics Award at IEEE-ROBIO 2019, and the Best Student Paper Finalist Award at IEEE-ROBIO 2015. 
\end{IEEEbiography}

\begin{IEEEbiography}[{\includegraphics[width=1in,height=1.25in,clip,keepaspectratio]{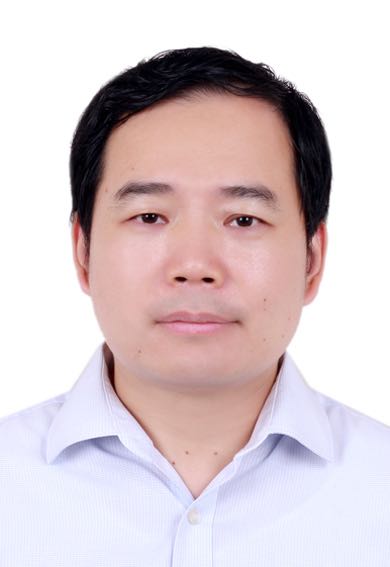}}]{Xinyu Wu} (Member 2022) is now a professor at Shenzhen
Institutes of Advanced Technology, and director of the Center
for Intelligent Bionic. He received his B.E. and M.E. degrees
from the Department of Automation, University of Science and
Technology of China in 2001 and 2004, respectively. His Ph.D.
degree was awarded from the Chinese University of Hong Kong
in 2008. He has published over 180 papers and two monographs. His research interests include computer vision, robotics,
and intelligent systems.
\end{IEEEbiography}
\begin{IEEEbiography}[{\includegraphics[width=1in,height=1.25in,clip,keepaspectratio]{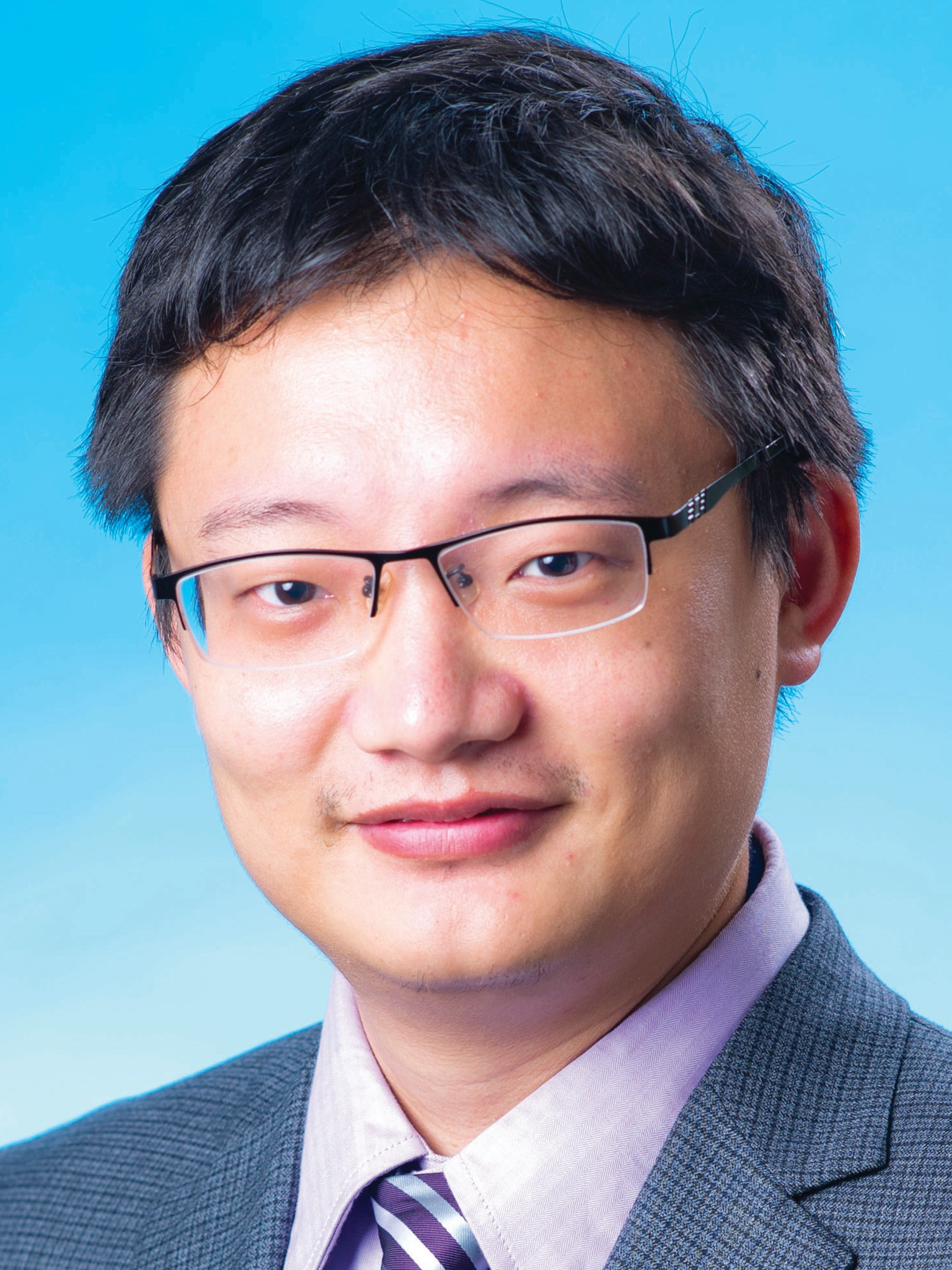}}]{Ming Liu} (Senior Member 2022) received the B.A. degree at Tongji University in 2005. He stayed one year in Erlangen-Nünberg University and Fraunhofer Institute IISB, Germany, as visiting scholar. He graduated as a PhD student from ETH Zürich in 2013. 

He is currently an Assoicate Professor at the Department of Electronic and Computer Engineering, The Hong Kong University of Science and Technology, Hong Kong. He has been involved in several NSF projects, and National 863-Hi-Tech-Plan projects in China. He is PI of 20+ projects including projects funded by RGC, NSFC, ITC, SZSTI, etc. He was the general chair of ICVS-2017, the program chair of IEEE-RCAR 2016, and the program chair of International Robotic Alliance Conference 2017. 

His current research interests include dynamic environment modeling, 3D mapping, machine learning and visual control, etc.
\end{IEEEbiography}

 \begin{IEEEbiography}[{\includegraphics[width=1in,height=1.25in,clip,keepaspectratio]{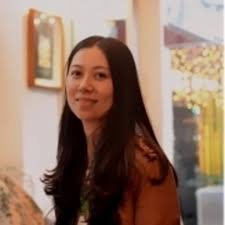}}]{Lujia Wang} (Member 2022) received the Ph.D. degree from the Department of Electronic Engineering, The Chinese University of Hong Kong, Hong Kong, in 2015. She was a Research Fellow with the School of Electrical Electronic Engineering, Nanyang Technological University, Singapore, from 2015 to 2016. She was an associate professor with the Shenzhen Institutes of Advanced Technology, Chinese Academy of Sciences, Shenzhen, Guangdong, from 2016-2021. Her current research interests include Cloud Robotics, Lifelong Federated Robotic Learning, Resource/Task Allocation for Robotic Systems, and Applications on Autonomous Driving.
\end{IEEEbiography}

\end{document}